# BRIDGE: Benchmarking Large Language Models for Understanding Real-world Clinical Practice Text

Jiageng Wu[1,*], Bowen Gu[1,*], Ren Zhou[2], Kevin Xie[3], Doug Snyder[4,5], Yixing Jiang[6], Valentina Carducci[4], Richard Wyss[1], Rishi J Desai[1], Emily Alsentzer[6], Leo Anthony Celi[5,7,8], Adam Rodman[9], Sebastian Schneeweiss[1], Jonathan H. Chen[10,11,12], Santiago Romero-Brufau[4,5], Kueiyu Joshua Lin[1,#], and Jie Yang[1,13,14,15,#]

[1] Division of Pharmacoepidemiology and Pharmacoeconomics, Department of Medicine, Brigham and Women's Hospital, Harvard Medical School, Boston, MA, USA

[2] Siebel School of Computing and Data Science, The Grainger College of Engineering, University of Illinois Urbana-Champaign, Urbana, IL, USA

[3] Department of Electrical Engineering and Computer Science, Massachusetts Institute of Technology, Cambridge, MA, USA

[4] Department of Otorhinolaryngology – Head & Neck Surgery, Mayo Clinic, Rochester, MN, USA

[5] Department of Biostatistics, Harvard T.H. Chan School of Public Health, Harvard University, Boston, MA, USA

[6] Department of Biomedical Data Science, Stanford University, Palo Alto, CA, USA

[7] Laboratory for Computational Physiology, Massachusetts Institute of Technology, Cambridge, MA, USA

[8] Division of Pulmonary, Critical Care and Sleep Medicine, Beth Israel Deaconess Medical Center, Boston, MA, USA

[9] Division of General Internal Medicine, Department of Medicine, Beth Israel Deaconess Medical Center, Boston, MA, USA

[10] Stanford Division of Computational Medicine, Stanford University, Stanford, CA, USA

[11] Division of Hospital Medicine, Stanford University, Stanford, CA, USA

[12] Clinical Excellence Research Center, Stanford University, Stanford, CA, USA

[13] Kempner Institute for the Study of Natural and Artificial Intelligence, Harvard University, MA, USA

[14] Broad Institute of MIT and Harvard, Cambridge, MA, USA

[15] Harvard Data Science Initiative, Harvard University, Cambridge, MA, USA

[*] **These authors contribute equally**

[#] **These authors jointly supervised this work**

**Correspondence:**

Jie Yang, PhD (jyang66@bwh.harvard.edu) and Kueiyu Joshua Lin, MD, ScD (jklin@bwh.harvard.edu), Division of Pharmacoepidemiology and Pharmacoeconomics, Brigham and Women's Hospital & Harvard Medical School, 75 Francis St, Boston MA 02115, USA

# Abstract

Large language models (LLMs) are evolving rapidly and hold great promise for medical applications, yet benchmarking on real-world clinical data such as electronic health records (EHRs) remains limited. Most existing benchmarks rely on medical exam-style questions or PubMed-derived text, failing to capture the complexity of clinical practice, while others target specific application scenarios with limited generalizability. Here we present BRIDGE, a comprehensive multilingual benchmark comprising 87 tasks sourced from 59 real-world clinical data sources across nine languages. It covers eight task types spanning the patient care continuum, including triage, information extraction, diagnosis, prognosis, and billing coding, and involves 14 clinical specialties. We systematically evaluated 95 LLMs (including DeepSeek-R1, GPT-4o, Gemini, and Qwen3) under multiple inference strategies. Results reveal substantial performance variation across model sizes, languages, natural language processing tasks, and clinical specialties. Open-source LLMs can match proprietary models, while medically fine-tuned models built on older backbones often lag behind updated general-purpose LLMs. BRIDGE and its continuously updated leaderboard provide foundational resources and important references for evaluating and developing LLMs for real-world clinical text understanding.

# Main

Recent advances in large language models (LLMs) have demonstrated a transformative potential in improving healthcare delivery and clinical research.[1] By combining extensive pretraining on vast corpora with supervised instruction tuning across diverse tasks,[2,3] LLMs exhibit exceptional capabilities in textual understanding, generation, and reasoning,[4–6] and show promise in medical applications. The prompt-based instruction provides an intuitive and easy-to-use approach to interact with LLMs on diverse tasks. For clinicians, LLMs can support the drafting of clinical documentation[7,8] and assist in clinical decision-making[9], enhancing efficiency and reducing workload burdens.[10,11] LLMs also offer the promise to benefit patients by providing simple interpretations of complex medical information[12] and personalized preventive advice,[13] promoting patient engagement, treatment adherence, and overall disease management.[14,15] Consequently, LLMs hold significant promise for improving the quality, cost and accessibility of healthcare services worldwide.[16] However, concerns persist regarding the reliability and clinical validity of LLM-generated outputs,[17] particularly given the high diversity of clinical tasks, specialties, and languages.[18] Moreover, LLMs are rapidly evolving, with diverse models emerging weekly, including proprietary and open-source, medical-specific and general-purpose, and varying in size. The general-purpose nature of LLM usage complicates institutional model selection and comparison. While clinical trials are necessary for the most high-risk cases, they are slow, expensive, and cannot possibly investigate every single use case. Clinical benchmarks, which provide automated, timely, and systematic evaluations of model performance, remain essential for clinicians, patients, health systems, and regulators, providing both understanding of the usability and trustworthiness of LLMs across diverse clinical scenarios.[19]

Existing benchmarks do not fully reflect the performance of LLMs in real-world clinical data, such as electronic health records (EHRs), which are critical as they document clinical status, history, and procedures that directly inform decision-making in clinical practice.[20] The commonly-used benchmarks focus on medical questions sourced from medical examinations or derived from academic literature, exemplified by the United States Medical Licensing Examination (USMLE)[21,22] or PubMed-based datasets.[23,24] These benchmarks oversimplify LLM evaluation in clinical practice by relying on multiple-choice tasks and normalized text, failing to reflect the complexity of real-world data such as noisy notes, abbreviations, acronyms, and non-standard expressions.[25] As a result, LLMs like GPT-4[26] and Med-PaLM2[27] achieve expert-level scores on the USMLE but perform unsatisfactorily in real-world clinical applications.[28,29] The recently released HealthBench[30] simulates consultation scenarios without using clinical text, and therefore cannot adequately reflect LLMs' capabilities in clinical text understanding. Although some studies have evaluated LLMs in specific clinical workflow,[31,32] they typically focus on a limited scope (e.g., a single specialty, task, language, or a small set of models),[33,34] making it difficult to generalize their findings or track LLMs' progress. Other benchmarks, such as MedHELM[35] and MedS-Bench[36], combine exam- and literature-based questions with limited amounts of EHR data, resulting in insufficient coverage of the complexity and diversity observed in genuine clinical workflows. In addition, the limited research on benchmarking medical tasks in non-English languages impedes the broader applicability of LLMs in global healthcare[35,37] and raises concerns about the potential bias arising from underrepresented languages and regions.[38–41] Such under-representative evaluations not only undermine the reliability and generalizability of LLMs but also amplify fairness and bias concerns, particularly across languages, regions, and healthcare systems.[42] When training and evaluation data are concentrated in high-resource languages or specific healthcare contexts, model outputs risk reinforcing existing inequities,[43] such as inferior performance in underrepresented populations or culturally specific clinical expressions.[44]

The rapid evolution of LLMs underscores a high demand for comprehensive and continuously updated medical leaderboards.[45] With advanced technology and models emerging every few weeks, the landscape of LLMs is dynamic and ever-changing. Notably, the recent introduction of medical LLMs such as Med-PaLM1/2[27,46], MeLLaMA,[47] and MedFound[48] also highlights the growing focus on improving performance and clinical relevance in healthcare. Leaderboards, which provide objective displays of model performance across a wide variety of tasks, are essential for providing fair comparisons of LLM capabilities and tracking performance variations, thereby offering valuable guidance for subsequent model development and clinical implementation. Such leaderboards have already been widely implemented in non-medical fields[49] but have not yet been well adopted in clinical domains. As LLMs are progressively integrated and deployed into clinical practice, robust benchmarking is vital for proactively identifying and mitigating potential risks and biases before they impact patient care.[50] Therefore, establishing a unified, realistic, and multilingual clinical benchmark is crucial for bridging the gap between the theoretical capabilities of LLMs and their practical implementation in specific use cases and care settings.[51,52]

To address the above challenges, we developed a large-scale benchmark that evaluates LLM performance on multilingual, real-world clinical text across diverse tasks. Building upon our systematic review of global clinical text resources,[53] this study proposes BRIDGE, a multilingual LLM benchmark that comprises 87 tasks from 59 real-world clinical text datasets, spanning nine languages. Reference standards for benchmark evaluation are sourced from the original data releases, including various forms of manual review and labels derived from structured data linked to the source EHRs.[53] BRIDGE is among the largest and most comprehensive multilingual benchmarks that exclusively focus on real-world clinical data, and one of the largest evaluations of LLMs in medicine to date. (A detailed comparison with existing benchmarks is provided in Supplementary Table S1.) We evaluated 95 LLMs, including DeepSeek-R1,[54] OpenAI GPT-4o,[55] Google Gemini,[56,57] Llama 4,[58] and Qwen3,[59] under three different commonly employed inference strategies (Zero-shot, Few-shot,[60] and Chain-of-Thought[61]) through HIPAA-compliant environments. By integrating a systematic task taxonomy, we established a leaderboard that not only provides a holistic perspective on LLM performance but also investigates their capabilities across various clinical settings, including inference strategies, languages, task types, and clinical specialties. This study provides critical insights and resources for integrating LLM into clinical practice, bridging the gap between LLM development and clinical applications.

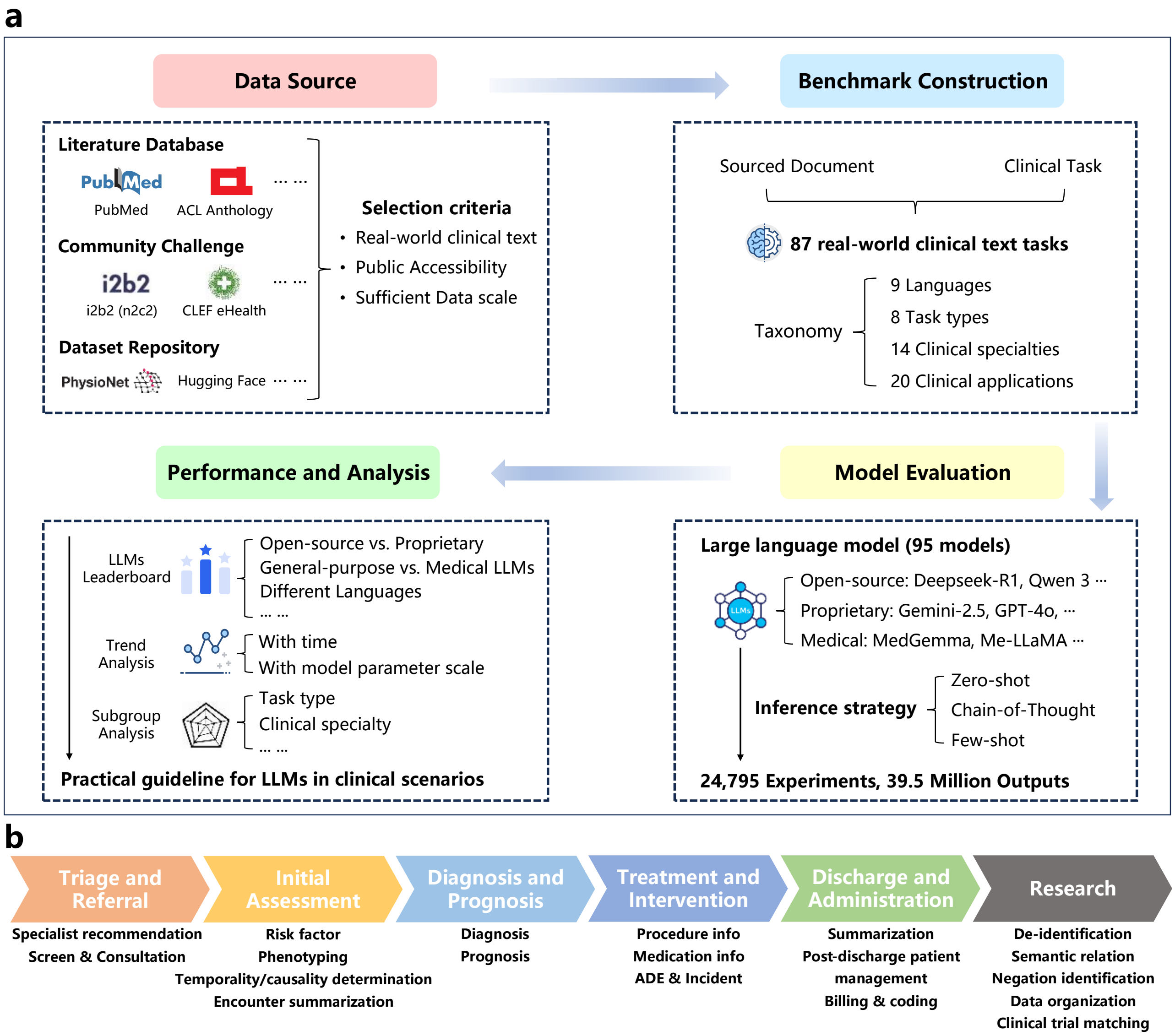

**Figure 1**. **Overview of benchmarking large language models in clinical text understanding.** (a) Workflow of benchmark construction, model evaluation, and performance analysis; (b) Clinical applications supported by the benchmark across different stages of patient care.

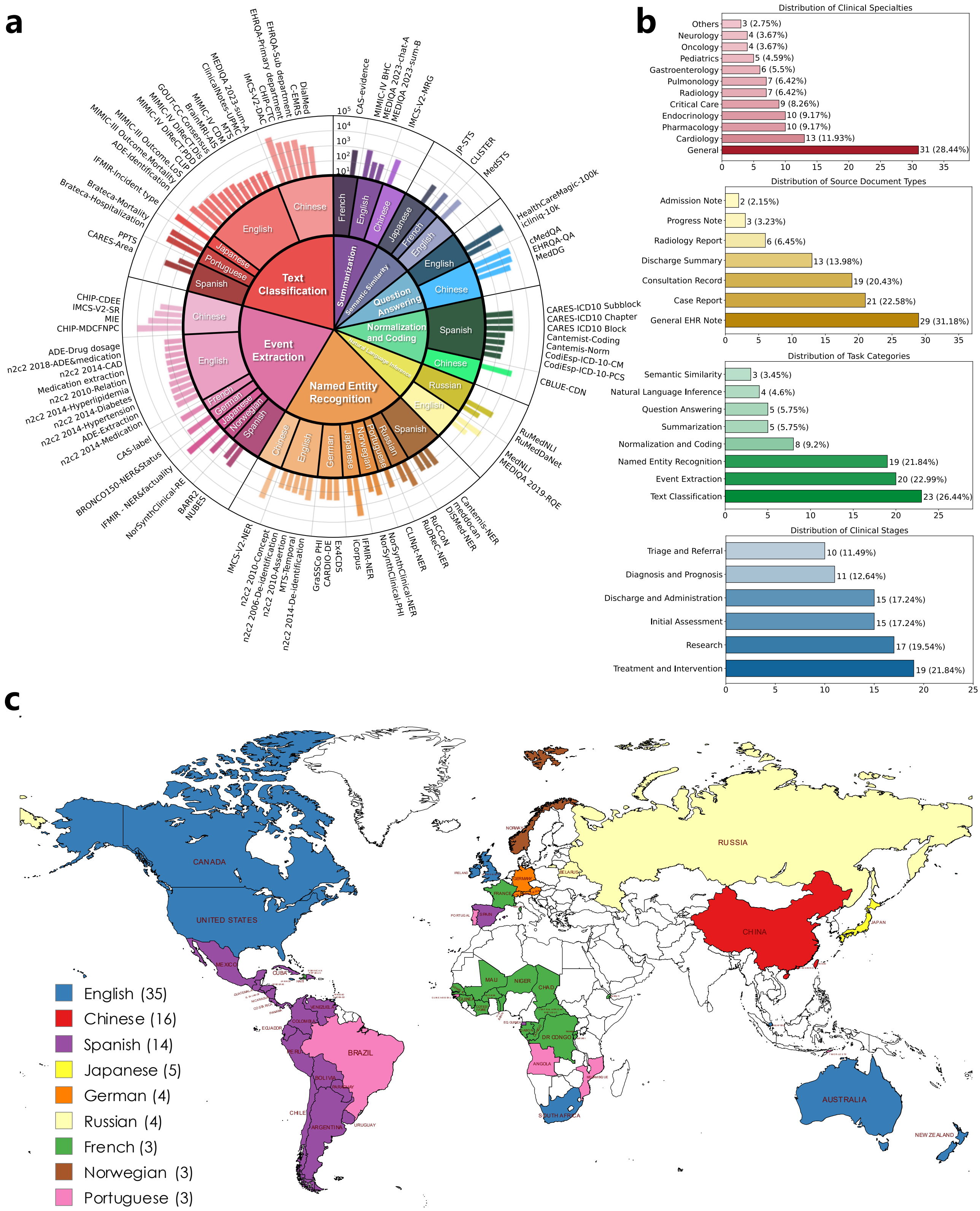

**Figure 2. Overview of Benchmark Characteristics and Task Distribution.** (a) Distribution of task types and associated languages, (b) Statistics on the distribution of clinical specialties, source document types, and task categories., and (c) Geographic distribution of countries where our benchmark covers the official languages.

# Results

## Benchmark Overview

The overall workflow of this study is illustrated in Figure 1. In total, our benchmark encompasses 87 tasks from 59 clinical text datasets. These tasks span nine languages and 1,418,042 samples, with 138,472 samples reserved for testing. Among them, 68 tasks (78.2%) are sourced from real-world EHR notes or clinical case reports, and 19 tasks (21.8%) are derived from real-world online patient-doctor consultation records. Figures 2a and 2b visualize the distribution of these tasks, covering 8 types, such as named entity recognition (e.g., phenotyping), classification (e.g., disease prediction), question answering, EHR summarization, and others. Figure 2c highlights this benchmark's coverage of nine languages, distributed as follows: English (35 tasks, 40.2%), Chinese (16, 18.4%), Spanish (14, 16.1%), Japanese (5, 5.8%), German (4, 4.6%), Russian (4, 4.6%), French (3, 3.5%), Norwegian (3, 3.5%), and Portuguese (3, 3.5%). Detailed information about all tasks can be found in the Section Methods.

As shown in Figure 3a, we evaluated 95 LLMs, covering advanced proprietary, open-source models (0.6B to 671B parameters), and medically specialized models. All LLMs were assessed under three distinct inference strategies: Zero-shot, Chain-of-Thought (CoT), and Few-shot, totally 24,795 ($95 \times 87 \times 3$) experiments. Comprehensive descriptions of the models, alongside their technical specifications, are available in Supplementary Table S2. Additionally, we investigated potential data contamination of the benchmark and found that most tasks did not appear to have been included in the training corpora of these models. Detailed analyses are in Supplementary Section 4 and Figure S2. The benchmark and leaderboard are publicly available and are increasingly and regularly updated.

## Overall Performance

We assessed LLMs using an overall score, defined as the average of their primary-metric scores across all tasks, to reflect their comparative performance across the entire benchmark, with the score ranging from 0 to 100 (Supplementary Table S3 for details). Figure 3b demonstrates the zero-shot performance of LLMs, while Figure 4 highlights the leading models under each inference strategy. Under the zero-shot setting, the three best-performing LLMs were Gemini-2.5-Flash[57] (44.8 [44.1-45.6], 95%CI), DeepSeek-R1[54] (44.2 [43.5-45.0]), and GPT-4o[55] (44.2 [43.4-45.0]). With CoT prompting, the overall

performance did not improve in general, with Gemini-2.5-flash (43.3 [42.5-44.1]) maintaining the top position, followed by Qwen3-Next-80B-Thinking (42.9 [42.1-43.7]) and DeepSeek-R1 (42.1 [41.3-42.9]). In contrast, few-shot prompting led to substantial performance gains, with Gemini-Series leading: Gemini-1.5-Pro (55.5 [54.7-56.3]), Gemini-2.5-Flash (53.4 [52.6-54.2]), and Gemini-2.0-Flash (53.3 [52.5-54.2]), followed by GPT-4o (52.6 [51.8-53.4]) and MedGemma-27b-it (52.0 [51.1-52.8]). Within the medical LLMs, these top-performing models showed distinct strengths across different inference strategies: HuatuoGPT-o1[62] achieved the highest score in the zero-shot setting (41.0 [40.2-41.8]), Baichuan-M2-32B[63] led under CoT (38.3 [37.4–39.2]), while MedGemma-27B-it[64] excelled in few-shot (52.0 [51.1–52.8]). Overall, few-shot learning emerged as the most effective inference strategy for these clinical text tasks. Although LLM performance varied across different tasks, our leaderboard reveals the substantial ability gap of current LLMs for comprehensive clinical text understanding across diverse tasks, with the highest overall score on the whole BRIDGE only 55.5 under few-shot setting, indicating considerable space for further improvements.

**LLMs Comparison**

Figure 3b reveals a generally upward trajectory in overall performance under zero-shot setting across the evolving landscape of LLMs, illustrating both the rapid development and substantial potential of LLMs. While proprietary models, represented by Google Gemini series and OpenAI GPT-4o, maintain a performance lead, open-source LLMs have been rapidly advancing and narrowing the gap. The newly released DeepSeek-R1 (671B) and Qwen3-Next-80B-Thinking outperform several proprietary LLMs, signaling a shift toward increasingly competitive open-source performance. Other widely adopted open-source LLMs also achieved comparable performance, including Mistral-Large-Instruct (score of 42.3),[65] Qwen2.5-72B-Instruct (41.6),[66] Gemma-3-27B-it (39.9),[67] and Llama-3.3-70B-Instruct (39.9)[68], as well as the derived variant Athene-V2-Chat[69] (41.7). Moreover, the progress across model generations highlights the rapid advance of LLM developments: the Qwen3 series consistently outperformed its Qwen2 predecessors, and Llama-3.3-70B-Instruct modestly surpassed Llama-3.1-70B-Instruct (39.9 [39.1–40.7] > 39.1 [38.3–39.9], $p = 0.16$). In contrast, the latest Llama-4-Scout-17B-16E-Instruct (109B)[58] showed a marked decline despite having more parameters (35.1 [34.4–35.9] vs. 39.9 [39.1–40.7], $p<10^{-4}$).

Domain-specialized LLMs demonstrated promising yet inconsistent trends. For instance, MedGemma-27B-it outperformed the Gemma-3-27B-it (40.8 [40.0–41.6] > 39.9 [39.1–40.7], $p$=0.12). Nevertheless, most medical variants did not surpass their general-purpose counterparts: MeLLaMA-70B-chat (32.3) and Llama-3-70B-UltraMedical (33.4) scored lower than the related Llama-3.1-70B-Instruct (391) ($p<10^{-4}$). Some even lagged behind their foundation models. For instance, HuatuoGPT-o1-72B (41.0) trailed Qwen2.5-72B-Instruct (41.6), and Llama-3.1-8B-UltraMedical (20.2) fell well below Llama-3.1-8B-Instruct (29.0) ($p<10^{-4}$).

Model scale also emerged as a key determinant of performance. Figure 3b illustrates the performance gains associated with increasing model size, with larger models generally outperforming smaller ones. The comparisons between DeepSeek-R1 and its variants with different model sizes (Figure 3c) demonstrate a consistent improvement in performance as model parameter size increases, aligning with the trend observed in the Llama, Qwen, and MeLLaMA model families. Together, these results highlight the effectiveness of scaling laws in enhancing LLM performance for clinical applications.[2] Models within the 70-80B range represent the most common category of open-source LLMs and typically achieve robust performance, led by Qwen3-Next-80B-A3B-Thinking (43.9), Athene-V2-Chat (72B) (41.7), and Qwen2.5-72B (41.6). Among middle-sized LLMs (~30B parameters), Qwen3-30B-A3B-2507-Thinking (41.8), Qwen3-32B-Thinking (41.0), and MedGemma-27B-it (40.8) stand out.

Furthermore, recently developed reasoning LLMs exhibit distinct model behavior with notable performance, which enables explicit thinking processes during inference.[6,54] As shown in Figure 3d, the reasoning LLM's performance on English text-classification tasks increases with longer CoT outputs, suggesting that in-depth thinking and reasoning may contribute to improved task accuracy. Representative reasoning models, such as DeepSeek-R1, gpt-oss-120B, and Qwen3-Next-80B-A3B-Thinking, provide substantially longer reasoning chains and outperform earlier non-thinking models. A similar positive association was also validated in other languages; detailed results are provided in Supplementary Section 3.6 and Supplementary Figure S1. In parallel, emerging mixture-of-experts (MoE) architectures markedly improve computational efficiency by activating only a subset of experts during inference. For example, DeepSeek-R1 (671 B) activates approximately 37B parameters per forward pass, whereas Qwen3-Next-80B-A3B-Thinking and Qwen3-30B-A3B-Thinking-2507 each activate about 3B parameters, achieving considerable efficiency gains without sacrificing accuracy.

**Inference Strategy Performance**

Figure 4a compares the performance of representative LLMs under three inference strategies: zero-shot, CoT, and five-shot (full comparison in Supplementary Table S3). Across the benchmark, few-shot prompting consistently yielded the highest accuracy, with 91 of 95 models (95.8%) achieving better results than in the zero-shot setting. Remarkably, 64 models (66.7%) showed performance gains exceeding 20%, underscoring the broad effectiveness of few-shot learning even when examples were randomly selected. Few-shot enhancements benefited both top-tier and lower-ranked LLMs. Among the leading models under zero-shot, DeepSeek-R1 improved from 44.2 to 51.4 (+7.2, +16.3%), while Gemini-1.5-Pro rose from 43.8 to 55.5 (+11.7, +26.7%), indicating that few-shot further augments even the most capable models. Similarly, models initially underperforming in zero-shot mode also exhibited significant improvements with few-shot prompting, such as the smaller LLMs (e.g., Llama-3.2-1B-Instruct from 12.7 to 24.4, +92.1%).

Notably, several medical LLMs benefited the most from the few-shot strategy. Llama3-OpenBioLLM-8B achieved the largest improvement, growing from 14.2 to 33.1 (+18.9, 133.1%), followed closely by Meditron-70B (15.7 to 32.1, +16.4, +104.5%). Other medical models, such as MedReason-8B, MMed-Llama-3,[36] HuatuoGPT-o1-8B, and Llama-3.1-8B-UltraMedical, also exhibited improvements ranging from 53.5% to 70.9%, indicating that task exemplars may be particularly valuable for domain-tuned models trained on limited instruction diversity. In contrast, explicitly enforcing step-by-step reasoning through CoT did not yield the expected performance gains for most LLMs, with only 11 models showing slight improvement.

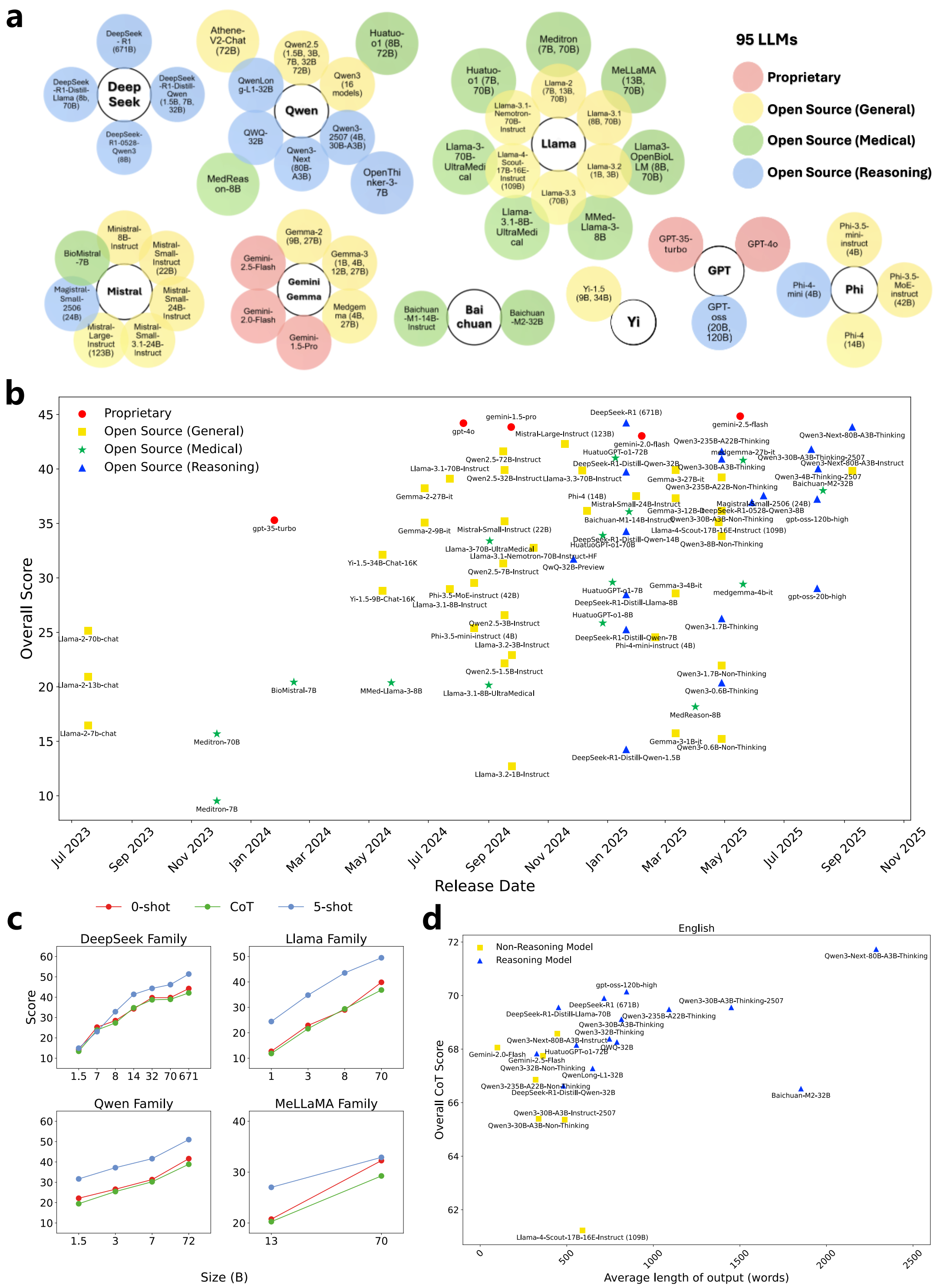

**Figure 3. Overview of evaluated LLMs and their performance.** (a) Categorization and information of evaluated LLMs. (b) Benchmark performance (zero-shot) of LLMs with their release dates. (c) Comparative performance analysis of LLMs of varying sizes within the same model family. (d) Model (≥30B and released after 2025/1/1) performance on English text classification tasks versus output length under CoT prompting (Other language see Supplementary Figure S1).

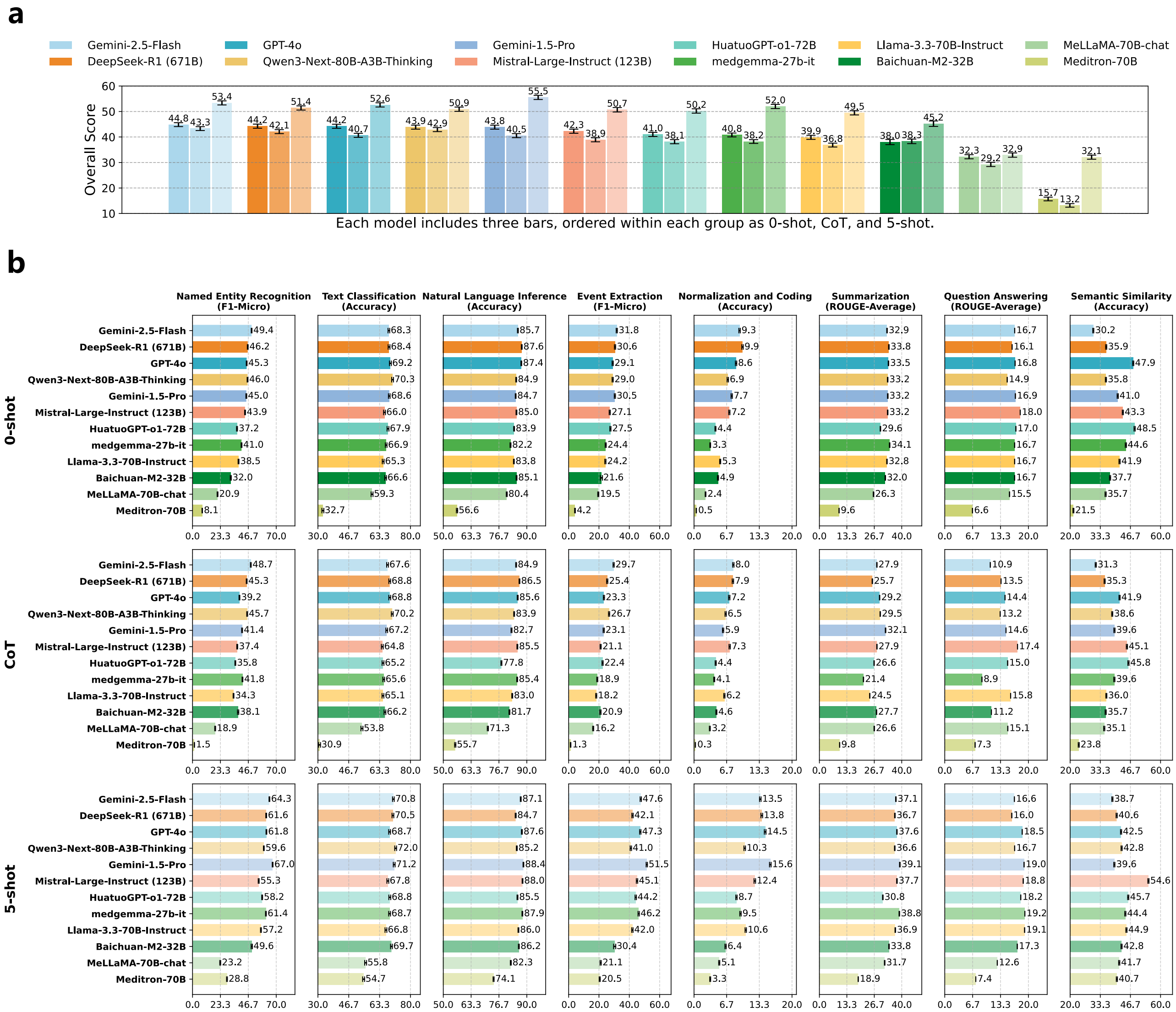

**Figure 4. Comparative performance of 12 leading and commonly used LLMs under different inference strategies.** (a) Overall score of LLMs across all tasks under three inference strategies: zero-shot, CoT, and 5-shot prompting. For each model, three bars are shown in the order of zero-shot, CoT, and 5-shot. (b) Performance of LLMs across different task categories under the three inference strategies. Each bar shows the mean performance of an LLM on the full benchmark (a) or within a task category (b), with 95% confidence intervals estimated via bootstrap resampling (1,000 iterations). Error bars are centered on the mean, with lower and upper bounds indicating the 95% CI.

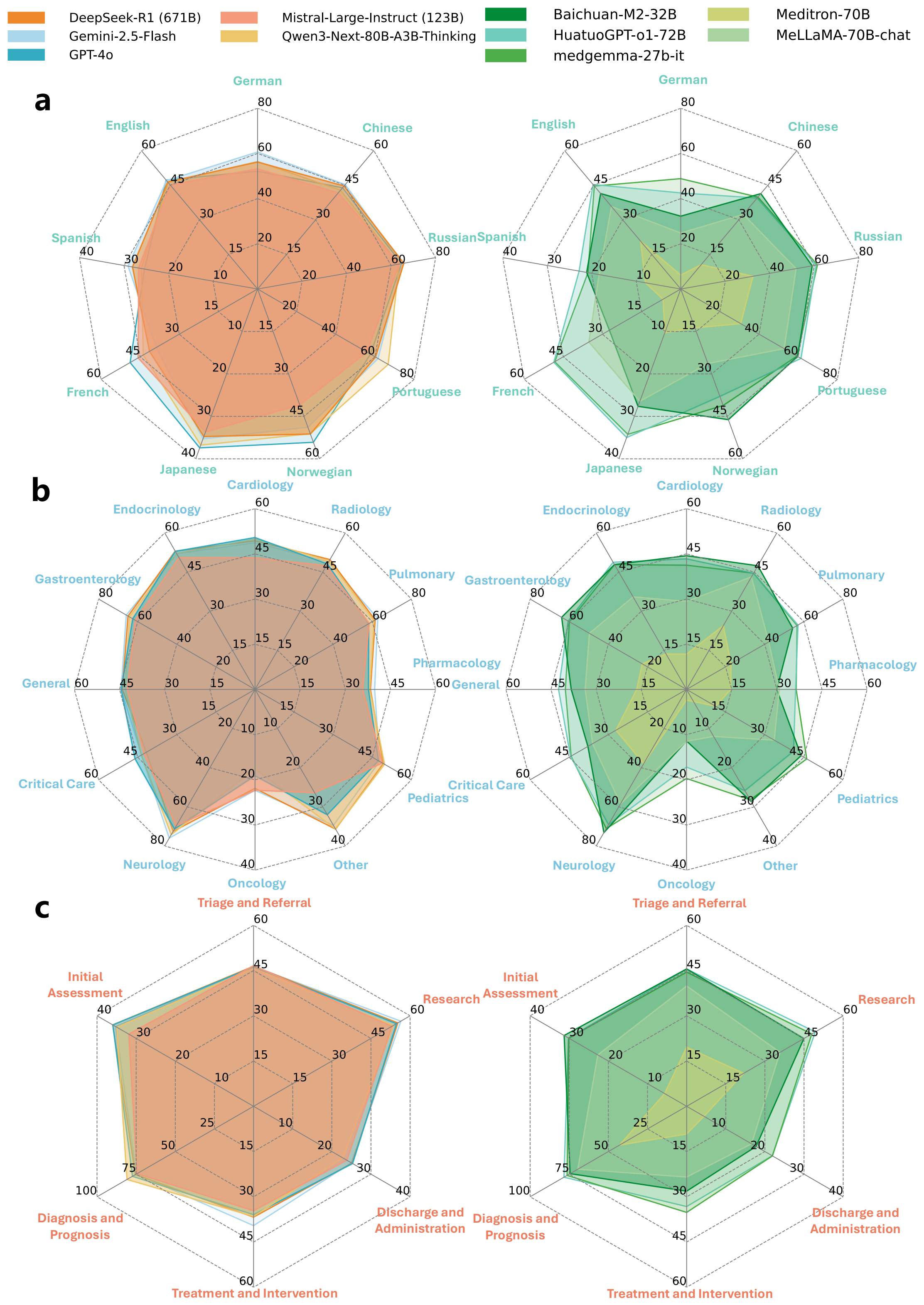

**Figure 5. Zero-shot performance of the five leading and commonly used LLMs in both the general and medical domains across different BRIDGE subgroups.** (a) languages, (b) clinical specialties, and (c) clinical stages. Each data point represents the mean model performance across tasks within the corresponding subgroup.

## Performance analysis for different task types

This benchmark encompasses a wide spectrum of clinical text tasks, and Figure 4b provides an overview of model abilities across different task types for both general and medical LLMs (Supplementary Table S4 for full details). Under zero-shot setting, gpt-oss-120B[70] achieved the highest accuracy in text classification (70.9), while Gemini-2.5-Flash excelled in information extraction, leading in both named entity recognition (NER, 49.4) and event extraction (31.8). DeepSeek-R1 attained top scores in natural language inference (NLI, 87.6) and normalization and coding (9.9), whereas HuatuoGPT-o1-72B showed the strongest performance in semantic similarity (48.5), Gemini-2.0-Flash performed best in question answering (QA, 18.4), and Gemma-2-27B-it led in summarization (34.7). In contrast, aside from HuatuoGPT-o1-72B and MedGemma-27B-it, most medically specialized LLMs struggled to generalize across diverse task types, with average ranks consistently below 20 out of 95 LLMs.

Figure 4b further illustrates the performance variations across different inference strategies. LLMs typically performed best on text classification and NLI, both of which offer well-defined, discrete label sets and thus present fewer ambiguities in the output. By contrast, information extraction tasks, such as NER and event extraction, benefited substantially from few-shot prompting, suggesting that more complex and context-dependent clinical tasks require examples to clarify detailed definitions and criteria. Meanwhile, normalization and coding tasks, which demand alignment with standardized medical coding systems (e.g., ICD-10), remained particularly challenging. Even with few-shot prompting, accuracy in these tasks remained relatively low (around 15%), reflecting the lack of embedded medical coding knowledge within most LLMs.[28] Similarly, text generation tasks, including QA and summarization, yield modest performance levels (around 20%), indicating that LLMs face significant challenges in clinical text generation.

## Performance analysis for different languages

Figure 5a summarizes the performance of LLMs across different languages, with details provided in Supplementary Table S5. Since task composition varies by language, model performance was averaged across tasks within each language to indicate its language-specific ability. Overall, many advanced models exhibited robust cross-linguistic adaptability, maintaining consistent performance across diverse languages. For instance, Gemini-2.5-Flash ranked first in Chinese, Spanish, and German, while Gemini-1.5-Pro performed best in English, and GPT-4o led in Japanese and Norwegian. Then, Qwen3-Next-80B-A3B-Thinking achieved the highest score in Portuguese, Qwen2.5-72B-Instruct led in French, and DeepSeek-R1 excelled in Russian. Based on the average score across languages, Qwen3-Next-80B-A3B-Thinking achieved the highest overall average (48.2). Notably, models from the Qwen family demonstrated exceptional multilingual competence, as exemplified by 12 of the top-20 models belonging to the Qwen2/3 series and their derived variants. This underscores the strong multilingual potential of well-optimized foundation models. However, for the parallel task of MedNLI (English)[71] and RuMedNLI (Russian)[72], where RuMedNLI is a full translation of MedNLI, all LLMs performed better on MedNLI, indicating the consistent performance degradation in non-English clinical contexts.

Among specialized medical LLMs, HuatuoGPT-o1-72B exhibited the broadest cross-linguistic versatility, achieving top performance in five languages (English, Spanish, French, Japanese, and Portuguese). MedGemma-27B-it showed competitive results in German and Russian, while Baichuan-M2-32B performed best in Chinese and German. However, English-centric medical LLMs (e.g., MeLLaMA and BioMistral) perform comparatively lower when applied to other languages. These results emphasize the necessity for more diverse multilingual corpora and language-specific tuning to ensure effective global deployment of LLM-based solutions.

**Performance analysis for different clinical specialties and clinical stages**

We further examined model performance across various clinical specialties (Figure 5b), reflecting the varied medical background from which datasets originated or the specific clinical challenges they addressed (Supplementary Table S6). As Figure 2b shows, this benchmark comprises 14 specialties, with “General” denoting datasets that span more than five domains or unspecified contexts, and “Others” including nephrology, dermatology, and psychology (one dataset each). Across specialties, Gemini-2.5-Flash consistently delivered top performance in gastroenterology (65.8), neurology (76.0), pulmonology (62.8), pharmacology (39.7), and general datasets (44.5). DeepSeek-R1 achieved the highest scores in radiology (49.8) and the “Others” category (35.7). Gemini-1.5-Pro led in critical care (46.6), pediatrics (50.4), and oncology (22.5), while GPT-4o excelled in cardiology, and Qwen3-30B-A3B-Thinking-2507 performed best in endocrinology. Despite domain-focused pretraining and supervised fine-tuning, the specialized medical LLMs did not show superiority over their general-purpose counterparts. Among these, MedGemma-27B-it demonstrated the broadest versatility, ranking first in 3 clinical specialties and achieving the highest average score across specialties (44.7), placing 14th among all 95 models, notably surpassing several larger general models such as Qwen3-Next-80B-A3B-Instruct and DeepSeek-R1-Distill-Llama-70B. These findings suggest that well-targeted domain adaptation can confer meaningful advantages even without further scaling.

Figure 5c further presents LLM performance across different clinical stages (Supplementary Table S7). The Gemini series demonstrated strong results: Gemini-1.5-Pro achieved the best performance in Initial assessment (37.1), Gemini-2.0-Flash excelled in Triage and Referral (47.5), and Gemini-2.5-Flash outperformed others in both Treatment and Intervention and Research-related tasks. GPT-4o and gpt-oss-120B led in Discharge and Administration, while among medical models, MedGemma-27B-it, HuatuoGPT-o1-72B, and Baichuan-M2-32B each delivered top performance in two clinical stages.

Overall, LLMs demonstrated stronger performance in the Diagnosis and Prognosis stage, likely due to the prevalence of well-structured classification tasks. In contrast, other stages often involve more complex tasks such as information extraction (e.g., phenotyping, temporal or causal relations) and text generation (e.g., summarization), which may pose greater challenges.

## Discussion

This study represented the largest benchmark to date for LLMs evaluations on multilingual, real-world clinical text, encompassing 87 tasks in nine languages. We developed a systematic framework and leaderboard for categorizing tasks and defining corresponding evaluation methods, enabling a thorough assessment of 95 LLMs with 24,795 experiments and 39.5 million inferences in total. With a broad task scope and a continuously updated leaderboard, clinicians can employ this benchmark to determine the candidate LLMs that best fit specific clinical or research contexts and deployment environments, while AI developers in healthcare can leverage it as a robust reference for further model fine-tuning and system integration. For patients, the benchmark serves as a preliminary assessment of the reliability of LLM outputs, thereby promoting better transparency and confidence in AI-assisted healthcare services.

Unlike medical exams or literature, the information in our benchmark is drawn from actual patient care, including EHR and doctor-patient interactions.[73,74] Differing from the simplified and standardized multiple-choice questions,[20] BRIDGE includes tasks specific to the administration and provision of health care, better reflecting the multifaceted capabilities of LLMs. Although certain LLMs achieved scores above 80 on standardized exams[27,34], for example, Deepseek-R1 reached a score of 92 on the USMLE dataset,[34] it attained an overall score of only 44.2 out of 100 on our benchmark. While this was the highest among the models evaluated, there remains substantial room for improvement. This stark discrepancy is sobering and highlights limitations of the current LLMs in clinical applications and the necessity of more clinically oriented evaluation before integrating LLMs into clinical practice.[75] Furthermore, many existing models and benchmarks remain English-centric, which restricts both the generalizability and equitable deployment of LLMs in global healthcare.[37] Our evaluation also reveals substantial performance variation across languages. Such variability may reflect differences in models' multilingual training exposure and regional data distributions, as well as whether and how models have been specifically optimized or aligned for non-English use. This underrepresented situation not only constrains accessibility but also introduces systematic biases, leading to performance degradation and even contextually inappropriate or unsafe clinical interpretations when applied to non-English settings.[39,42,76] This highlights the need for more community efforts in multilingual clinical data resources and evaluation to support more inclusive deployment.[77] Meanwhile, BRIDGE captures a broad spectrum of real-world clinical contexts, from early-stage triage and referral to diagnosis, treatment recommendation, and billing coding, thereby avoiding evaluations that overemphasize any single dimension.[36] Notably, clinical workflow-critical tasks such as normalization and coding have been relatively underexplored and remain challenging for current LLMs.[28] These tasks demand not only robust interpretation of noisy clinical narratives but also precise understanding and adherence to formal coding schemes.[78,79] Additionally, LLM performance varies across clinical specialties, such as relatively weaker performance in oncology-related tasks. This heterogeneity underscores the need for continued domain- and application-oriented advances to support reliable use in practice, including better integration of specialty knowledge and supporting resources. For instance, retrieval-augmented generation (RAG)[80] can improve LLM performance and reliability by incorporating external domain knowledge with in-context inference, thereby helping mitigate misinformation in clinical applications.[81] Collectively, by encompassing nine languages and diverse applications, BRIDGE facilitates more inclusive and clinically relevant evaluation of LLMs across diverse real-world healthcare systems.[82]

Given the rapid and transformative progress of LLMs,[83] especially fueled by open-source initiatives such as Qwen, LLaMA, and DeepSeek,[66,68] our evaluation of 95 LLMs provides timely and valuable insights on integrating LLMs into clinical applications. The results reveal the continuous progress of open-source models, exemplified by Deepseek-R1 surpassing several evaluated proprietary systems (e.g., GPT-4o, Gemini-2.0-Flash).[84] These open-source foundation models also expedited the development of medical LLMs, with mid-sized domain models such as MedGemma-27B and Baichuan-M2-32B showing encouraging performance, suggesting the potential of targeted domain adaptation. Nevertheless, most medical-specific variants generally underperform their general-purpose counterparts. This gap partially stems from the outdated base models these medical LLMs were built on, and the limited clinical relevance of their training data,[85] which is mainly drawn from biomedical literature (e.g., PubMed) rather than EHR data.[86,87] Although some models (e.g., BioMistral-7B and MMed-Llama 3) have undergone domain-specific pretraining and demonstrate strong potential, they often lack broad multi-task instruction fine-tuning or disproportionately rely on medical question datasets. The lack of task diversity may lead to overfitting and reduce the LLM's generalizability across tasks.[88] This limitation is reflected in their weaker performance under zero-shot settings and the substantial gains from few-shot prompts, which are compensated for by injecting task-specific context.[89] Moreover, models' ability to accurately interpret task instructions and generate valid responses (as shown in Supplementary Table S8) varies considerably; several smaller and medical-specific LLMs exhibit lower valid-response rates, which further undermine model performance. Additionally, models such as HuatuoGPT-o1-72B, which incorporated reinforcement learning on top of an instruction-tuned base (i.e., Qwen2.5-72B-Instruct), demonstrated notable performance gains with robust versatility to broad tasks.[62] Despite the multilingual capabilities inherited from extensive web-data pretraining,[37,90] most medical LLMs remain primarily optimized for English, leaving genuinely multilingual clinical foundation models relatively underexplored.[91] Additionally, open-source solutions facilitate on-site deployments within hospitals, enabling more localized data and model governance, thereby reducing potential privacy and security risks.[92] Our experiments further confirm that scaling laws[2] persist within clinical tasks: the larger models significantly and consistently outperform the small ones, which emphasizes the enduring trade-off between performance and computational cost.

Inference strategy plays a pivotal role in the practical deployment of LLMs, particularly since most clinical applications will likely not involve LLM fine-tuning but rather rely on task-specific prompting.[93] Our findings indicate that few-shot prompting proves highly effective for clinical text tasks, significantly enhancing task-specific comprehension and contextualization with only five randomly selected examples. To enhance interpretability, CoT explicitly instructs LLMs to generate step-by-step reasoning before arriving at a final answer.[61] Contrary to observations in other domains,[94] CoT did not yield consistent performance gains in our benchmark and mostly impaired results.[22] This discrepancy appears to stem from the lack of sufficiently grounded medical knowledge to support accurate multi-step reasoning[95] and the heightened risk of hallucinations in such a knowledge-intensive setting.[96] Meanwhile, the newly developed reasoning LLMs (e.g., DeepSeek-R1 and Qwen3-Series), which employ reinforcement learning to strengthen reasoning capabilities, achieved superior results on our benchmark.[54,97] These models demonstrated a promising direction for developing interpretable LLMs that more closely mirror human decision-making processes. Beyond improved reasoning, the integration of external domain knowledge, such as via retrieval-augmented generation,[80] can further enhance LLM performance and reliability, leveraging their in-context learning abilities while mitigating the risk of misinformation in clinical applications.[80]

Given the tremendous potential of LLMs in healthcare scenarios as well as the considerable risks to patient safety, transparent and robust benchmarking will be essential to adopt LLMs into clinical practice.[98,99] Future benchmarks should encourage closer collaboration among clinicians, patients, and LLMs to better simulate real-world interactions, enhance overall clinical effectiveness, and provide a validated human baseline.[100] Critically, such evaluations can also extend beyond cross-sectional text snapshots to longitudinal, multimodal patient-journey settings where diagnosis and therapy decisions are made under evolving and noisy information.[101] Meanwhile, adopting standardized reporting frameworks (e.g., TRIPOD-LLM[102]) are essential for enhancing transparency, reproducibility, and comparability in future medical applications, thereby harmonizing evaluation and broad clinical practices. Furthermore, LLMs have been observed to exhibit overoptimism in their inferences,[103] which is critical to address in medical applications. Our benchmark provides a foundation for evaluating the trustworthiness of such predictions. Given the complexity of clinical practice, the scores in our benchmark do not fully equate to LLM performance on specific clinical applications, which require further rigorous assessment. However, BRIDGE provides timely and comprehensive comparisons across diverse tasks and models, serving as a valuable starting point for model selection and filtering, guiding decisions before committing to resource-intensive evaluations or further development of selected base models.

This study has several limitations. First, given the large scale of the dataset, we used reference labels obtained from the original data sources rather than re-annotating by human clinicians. Second, as concerns regarding privacy and safety restricted the availability of accessible clinical text datasets, certain tasks were developed on overlapping data sources, restricting corpus diversity. Third, this work primarily evaluated LLMs under common inference settings without incorporating specific instruction optimization. We will further examine more strategies (e.g., instruction tuning, fine-tuning, and RAG) to enhance model adaptability to clinical workflows in future studies. Fourth, the overall score was calculated by averaging the primary metrics across task types, which may introduce inconsistencies due to varying metrics. However, we provide fine-grained comparisons on the leaderboard across task types under the same metric to ensure fair comparisons. Finally, while we investigated 95 advanced models, several newly released LLMs (e.g., OpenAI o3, Gemini-Pro-2.5-Pro, and Med-PaLM2) were not evaluated due to the constraints of model access and resource limitations within HIPAA-compliant healthcare systems. Nonetheless, this leaderboard is continuously updated to maintain its relevance and currency, offering a dynamic resource for tracking and advancing LLM performance in clinical text understanding.

In conclusion, this study established a comprehensive benchmark and systematically evaluated LLMs on real-world clinical text understanding. By centering on real-world EHR-based tasks and capturing the complexity of clinical text, our findings highlight the gap between current LLM capabilities and the demands of clinical practice, while also revealing substantial performance variability across models, languages, and clinical scenarios. These insights provide critical guidance for optimizing LLMs in healthcare and inform future efforts to align model development with the practical needs of clinical applications.

# Methods

## Clinical Text Dataset Collection

To comprehensively evaluate the LLMs performance on real-world clinical text data, we systematically identified and curated a diverse collection of clinical text datasets representative of authentic clinical scenarios.

This process was initially guided by our prior systematic review of clinical text datasets,[53] which conducted a global survey of publicly available resources. Building upon this foundation, we expanded our search scope and employed a standardized protocol to ensure that the included datasets fully satisfied the benchmarking criteria of clinical relevance, diversity, and suitability.

Specifically, we targeted three primary sources:

1. **Literature Databases**: Widely recognized biomedical literature databases, including PubMed and MEDLINE, and computational linguistic repositories, notably the ACL Anthology, a leading digital archive of Natural Language Processing (NLP)-focused journal articles and conference proceedings.
2. **Community Challenges**: Commonly used and actively maintained clinical NLP challenges and benchmarks, such as the National NLP Clinical Challenges[104] (n2c2, formerly i2b2) and CLEF eHealth.[105]
3. **Dataset Repositories**: Biomedical dataset platforms (PhysioNet[106]) and NLP-focused dataset hubs (Hugging Face[107]), which store extensive collections of biomedical datasets and are frequently updated with new resources.

Detailed search strategies, including specific Medical Subject Headings (MeSH) terms and keywords, can be found in our prior review.[53]

## Criteria for Dataset Selection

Datasets identified through these sources underwent screening based on the following predefined inclusion and exclusion criteria:

1. **Real-World Clinical Text Data**: Eligible datasets were required to consist of authentic clinical texts derived directly from real-world medical settings, such as electronic health records (EHRs), clinical case reports, or healthcare consultations. Non-clinical sources (e.g., textbooks, social media) or datasets relying primarily on non-textual (e.g., genomic or protein sequences) or multimodal inputs were excluded.

2. **Public Accessibility and Availability**: Datasets included in this investigation are publicly available or accessible through standardized request procedures to ensure reproducibility and transparency.
3. **Sufficient Data Scale**: Only datasets containing at least 200 samples were included to ensure reliable evaluations and robust statistical analyses.

Finally, the curated datasets provided a diverse corpus representative of authentic clinical scenarios.

**Benchmark Construction**

Based on the included datasets, we constructed a set of clinical text tasks tailored for assessing LLMs. These tasks simulate diverse clinical scenarios characterized by complexity, contextual variability, and multi-source information requirements. Unlike traditional NLP methods, which typically rely on supervised training with task-specific model architectures, LLMs perform tasks by interpreting textual prompts without dedicated training. Therefore, precise task design and standardization are critical for fair and objective evaluations. Detailed information about all tasks can be found in Supplementary Section 4 and Section 5, and the metadata of tasks are in the Supplementary Table S9.

To ensure task suitability and consistency, we transformed and standardized datasets through the following structured process:

**Task Definition and Categorization**

Task objectives and evaluation criteria were primarily derived from original dataset descriptions or primary publications (hereafter collectively referred to as dataset source). Tasks were categorized into different types:

1. Text classification: Determine or predict categorical labels (e.g., diagnosis, risk stratification) based on the provided clinical texts.
2. Semantic similarity: Assessing the similarity of two sentences or clinical notes.
3. Natural Language Inference (NLI): Evaluating the logical relationships (e.g., entailment, contradiction, neutrality) between paired texts.

4. Normalization and coding: Map the whole clinical note or the extracted entities to standardized clinical code systems (e.g., ICD, SNOMED)
5. Named Entity Recognition (NER): Identify the medical entities and label them with appropriate types (e.g., symptom, disease, examination).
6. Event extraction: Identify the medical entities and capture additional attributes or relations beyond simple entity types (e.g., temporal status, severity).
7. Question-Answering (QA): Generating accurate responses to healthcare inquiries.
8. Summarization: Condense clinical notes into concise summaries by retaining essential information, with extraction or generation methods.

**Input Text Preparation and Standardization**

Relevant textual information for each task was systematically extracted from the original datasets and integrated using standardized templates. For instance, the required text fields were distilled from the whole EHR database and then condensed into structured inputs with templates (e.g., "Chief Complaint: ..., Examination: ..."). Additionally, for tasks introducing structured metadata (e.g., demographic information and examination results), we transformed these structured features into explicitly labeled textual forms, integrating them seamlessly with clinical notes. The template for input and output can be found in Supplementary Section 4.1.

**Output Standardization and Formatting**

Given the heterogeneity of original task outputs, including classification logits, BIO-style entity tags, or structured annotations, all outputs were instructed to be standardized into clear, structured textual responses for automatic result processing and analysis on evaluation. Specifically:

1. Tasks for Text classification, Semantic similarity, NLI, and Normalization and Coding (document level): Outputs were standardized into explicit textual labels indicating predicted categories.
2. Tasks for NER, Event extraction, and Normalization and Coding (entity level): Outputs were formatted into structured textual annotations clearly indicating subject spans and other required attributes (e.g., “Entity: ..., Type: ..., Status: ...”).

3. Tasks for QA and Summarization: Outputs were directly formulated as concise, structured free-text outputs.

Outputs that failed to follow the required format were regarded as invalid responses during the evaluation phase. This uniform output format enables automated extraction and quantitative evaluation of LLM-generated results, ensuring efficient and objective performance assessment.

**Task Instruction (Prompt) Definition**

Due to the complexity of clinical text tasks, which often rely on professional definitions and domain-specific terminology, we prioritized the use of task instructions from authoritative dataset sources, including original dataset papers, annotation guidelines, and supplementary data descriptions. To minimize variability and potential biases introduced by differing prompts, we adopted straightforward and concise instruction templates for all tasks.

Given the multilingual nature of our benchmark, we preferentially retained the task description aligned with their original language if available, preserving context-specific semantics critical for accurate interpretation. Meanwhile, the instructions for other tasks and the base templates (both input and output may involve) were uniformly provided in English, as the existing LLMs all support English and yield fine performance in English.

**Dataset Splitting**

Dataset partitions followed the official splits defined by dataset sources whenever available, facilitating direct comparability with prior studies. For datasets without predefined splits, we applied the following selection strategy: for datasets with over 2000 samples, 10% were randomly selected as the test set; for datasets with 1000 to 2000 samples, 20% were selected; and for datasets with fewer than 1000 samples, all samples were used for testing except for 20 cases reserved as a pool for selecting few-shot examples. For each dataset, five samples outside the test set were randomly selected as few-shot examples. All the benchmark experiments were conducted on the split test partitions.

**Task Taxonomy and Characteristics**

To systematically investigate the abilities of LLMs across different clinical scenarios, we extracted key features for each task and mapped them into standardized taxonomies. These include

Language, Sourced Clinical Document, Clinical Specialty, and Clinical Stage and Application, which can refer to Supplementary Section 4.2 Task Taxonomy and Characteristics.

**Model Implementation**

We included a diverse range of state-of-the-art LLMs, covering both proprietary and open-source LLMs. Detailed information for all models is provided in Supplementary Table S2. All experiments, including data selection and model inference, used a fixed random seed (42) across all tasks and models to ensure reproducibility. For decoding configuration, unless explicitly specified by the model's official documentation, all models were evaluated using a greedy decoding strategy (temperature = 0, top_p = None, top_k = None) to eliminate randomness and produce deterministic outputs. For the Qwen3 series, the decoding parameters recommended in its official guideline were adopted.

The model inference was conducted using the following computational setups:

1. Open-source models: all open-sourced models (except DeepSeek-R1[671B]) were deployed locally on Mass General Brigham (MGB) institutional server with 8*NVIDIA H100 GPUs. The inference process was accelerated using the vLLM framework[108] to optimize efficiency. DeepSeek-R1 was deployed on Microsoft Azure managed by Stanford University.
2. Proprietary models: Due to privacy and security considerations, proprietary models were evaluated via institutional cloud infrastructure within the compliance of HIPAA:
    a) OpenAI models (GPT-35-Turbo-0125 and GPT-4o-0806): Deployed on Microsoft Azure server at Mass General Brigham.
    b) Google models (Gemini series): Deployed on Google Cloud Platform at Mayo Clinic.

All records of inference requests and responses were securely stored on institutional servers at MGB in accordance with MGB data governance policies. All experiments were conducted in full compliance with the HIPAA regulations.

**Inference Strategy**

In this study, we systematically evaluated three distinct inference strategies:

1. Zero-shot: Only the task instructions and input data were provided. The LLM was prompted to directly produce the target outputs without any support.

2. Chain-of-Thought (CoT): Task instructions explicitly directed the LLM to generate a step-by-step explanation of its reasoning process before providing the final output, which can significantly improve the model's interpretability.
3. Few-shot: Five reserved independent samples serve as examples, which leverage the LLM's capability of in-context learning to guide the model to conduct tasks. For models supporting conversational interactions, examples were presented sequentially to simulate realistic user-system dialogues; otherwise, input-output pairs were directly appended to the instruction.

Details about the prompt for different inference strategies can be found in Supplementary Section 4.1.

**Evaluation Framework**

To ensure consistent, objective, and automated evaluations, we established standardized evaluation schemes covering reference standard, result extraction, task-specific metrics, and statistical analysis.

**Reference Standard**

For each task in our benchmark, the reference standard is sourced from the labels released with the original source datasets. These labels were generated through different mechanisms, including expert manual annotation and derivation from structured EHR systems, and underwent the quality-control procedures defined by dataset creators. To maintain consistency with prior work and preserve the integrity of each dataset, we adopted these original labels without additional modification.

**Result Extraction**

We develop an automated script for each task separately to extract results from the standardized LLM outputs described previously. For outputs failing to meet the required formatting standards, we regarded them as invalid responses. We calculated the valid rate for each experiment setting and presented the results in Supplementary Table S8. For tasks under the types of text classification, semantic similarity, NLI, and normalization/coding (with explicitly defined labels), invalid model outputs were replaced with randomly assigned labels from the valid label set. For the remaining task types, invalid outputs were retained as empty responses.

**Evaluation Metrics**

Representative metrics were carefully selected for each task category, with a designated primary metric facilitating overall benchmarking comparisons:

1. Text classification, Semantic similarity, NLI, and Normalization and Coding (document level): We evaluate these tasks with Accuracy (primary metric), micro F1-score, and macro F1-score. Accuracy directly reflects overall classification performance by measuring the proportion of correct predictions. The F1 scores provide complementary insights by considering precision and recall across classes. The micro F1-score emphasizes performance in common classes, while the macro F1-score equally weights all classes, highlighting performance in less frequent categories.
2. NER, Event extraction, and Normalization and Coding (entity level): We evaluate these tasks with subject-level F1-score and event-level F1-score (both calculated by micro-scoring). The subject-level F1-score only evaluates the model's ability to identify the correct subjects without considering their attributes, providing preliminary performance insights. Event-level F1-score, the primary metric, comprehensively evaluates model accuracy by measuring extraction precision and recall across entities and their attributes.
3. QA and Summarization: We evaluate these tasks with BLEU-4, ROUGE-average (primary metric), and BERTScore. ROUGE-average[109] is the average score of ROUGE-1, ROUGE-2, and ROUGE-L, thus capturing the recall of unigrams, bigrams, and longest common subsequences between the candidate and reference texts. BLEU-4[110] combines the precision of 1-, 2-, 3-, and 4-gram matches between generated outputs and references. BERTScore[111] evaluates semantic alignment between generated and reference texts by leveraging contextual embeddings from BERT. All these metrics generally show a consistent trend across tasks, while the ROUGE-average exhibits more distinctions among models in our experiments. Therefore, we adopt ROUGE-average as the primary metric.

These metrics were computed using standard libraries to ensure reproducibility: *Scikit-learn*[112] for classification and extraction tasks, nltk[113] for BLEU-4, *rouge_scorer*[114] for ROUGE-average, and *bert_score*[111] for BERTScore.

**Performance Calculation**

To enable quantitative comparisons, we compute an overall score for each LLM by averaging its primary-metric values across all tasks. This aggregate measure reflects a model's relative performance on the benchmark as a whole. For subgroup analyses, such as task type, language, clinical specialty, and clinical stages, the same averaging procedure is applied to the subset of tasks that meet the specified criteria, yielding a focused performance estimate within that domain.

**Statistical Analysis**

Model performance was evaluated with non-parametric bootstrapping (1,000 resamples with replacement). For each model, we computed the bootstrapped mean and its 95 % confidence interval (CI), yielding robust estimates of central tendency and sampling variability. Pairwise model comparisons were assessed with a two-sided significance test. All analyses were performed in Python 3.10.15 (NumPy v1.26.4, SciPy v1.14.1).

**Data Contamination Analysis**

The advanced LLMs typically undergo extensive training on vast data, raising the possibility of unintended data exposure of benchmark. To assess this potential data contamination, we employed a text completion-based approach to detect possible leakage of benchmark data into the evaluated models' training corpora.[115] Specifically, we tokenized each test sample using model-specific tokenizers, truncating sequences at predetermined positions (tokens 10, 15, 20, 25, and 30) and prompting the LLM to predict the subsequent five tokens. The accuracy of predicted tokens compared to actual tokens (5-gram accuracy) was measured. A test sample was classified as potentially leaked if predictions exactly matched actual tokens at three or more truncation positions. To balance sufficient contextual information while avoiding overly simplistic long-context completions, truncation points began from token position 10, incrementing by intervals of 5 tokens.

## Acknowledgments

We thank Xiaocong Liu, Wanxin Li, Qingcheng Zeng, Zichang Su, and Xiaoyue Wang for the initial dataset processing. This study was partially funded by PCORI ME-2022C1-25646, Goldberg Scholarship and Brigham Research Institute, National Institute on Aging RF1AG090405, and National Library of Medicine R01LM014667. L.A.C. is funded by the National Institute of Health through DS-I Africa U54 TW012043-01 and Bridge2AI OT2OD032701, the National Science Foundation through ITEST #2148451, a grant of the Boston-Korea Innovative Research Project (RS-2024-00403047) and a grant of the Korea Health Technology R&D Project (RS-2024-00439677) through the Korea Health Industry Development Institute (KHIDI) as funded by the Ministry of Health & Welfare, Republic of Korea. S.S. is funded in part by FDA research contracts (HHSF223201710186C and HHSF223201710146C), the FDA Sentinel Innovation Center (75F40119F19002), the NIH (NHLBI R01-HL141505, NIAMS R01-AR080194), the Burroughs Wellcome Fund, and PCORI. J.H.C. was supported by NIH/National Institute of Allergy and Infectious Diseases (1R01AI17812101), NIH-NCATS-Clinical & Translational Science Award (UM1TR004921), Stanford Bio-X Interdisciplinary Initiatives Seed Grants Program (IIP) [R12] [JHC], NIH/Center for Undiagnosed Diseases at Stanford (U01 NS134358), Stanford RAISE Health Seed Grant 2024, Josiah Macy Jr. Foundation (AI in Medical Education), Stanford CARE AI Scholar Fellowship. The funders had no role in study design, data collection and analysis, decision to publish, or preparation of the manuscript.

## Author contribution

J.Y. designed this study. J.W. and B.G. constructed the benchmark and conducted the experiments. J.W., B.G., and J.Y. analyzed the results and drafted the initial manuscript. R.Z. and K.X. contributed to the benchmark construction, J.W., B.G., and R.Z. contributed to the data contamination experiments, and K.X. and J.W. contributed to building the leaderboard. D.S., V.C., S.R-B, and Y.J. contributed the inferences of Gemini-series and Deepseek-R1. R.W., R.J.D., E.A., L.A.C., A.R., S.S., J.H.C., S.R-B., K.J.L., and J.Y. contributed to manuscript design and refinement. J.Y. and K.J.L. supervised this study. All authors revised, read, and approved the manuscript.

## Competing interests

K.J.L. has received research grants from Takeda, AbbVie, and UCB for projects unrelated to this study. E.A. reports consultant fees from Fourier Health. S.S. is participating in investigator-initiated grants to the Brigham and Women's Hospital from Boehringer Ingelheim, Takeda, and UCB unrelated to the topic of this study. He is an advisor to Aetion Inc., a software manufacturer. S.S. is an advisor to Temedica GmbH, a patient-oriented data generation company and his interests were declared, reviewed, and approved by the Brigham and Women's Hospital in accordance with their institutional compliance policies. J.H.C. reports cofounding Reaction Explorer, that develops and licenses organic chemistry education software, and receive medical expert witness fees from Sutton Pierce, Younker Hyde MacFarlane, Sykes McAllister, Elite Expert, consulting fees from ISHI Health, and honoraria or travel expenses for invited presentations by insitro, General Reinsurance Corporation, Cozeva, and other industry conferences, academic institutions, and health systems. The remaining authors declare no competing interests.

## Data availability

All fully open-access datasets in BRIDGE are shared at https://huggingface.co/datasets/YLab-Open/BRIDGE-Open under their respective data use agreements. All dataset sources and corresponding links for additional data access are listed in Supplementary Section 4. The leaderboard is available at https://huggingface.co/spaces/YLab-Open/BRIDGE-Medical-Leaderboard, where we also provide contact information for submitting new models for evaluation without requiring direct data access. Additional metadata supporting this study are available from the corresponding author upon reasonable request.

## Code availability

The corresponding benchmark evaluation code can be found at https://github.com/YLab-Open/BRIDGE.

# Supplementary Information

## Contents

## 1 Comparison of existing evaluations and BRIDGE

**BRIDGE** focuses on clinical text derived from real-world clinical practice, addressing the complexity and diversity often absent in existing medical benchmarks. Built upon a systematic collection of multilingual and publicly accessible clinical text datasets, BRIDGE not only standardizes diverse tasks to make them compatible with large language model (LLM) evaluation but also introduces newly curated tasks, thereby filling critical gaps in assessing LLMs on real-world clinical text. It establishes the largest and most comprehensive task suite to date, encompassing 87 clinical text understanding tasks that span the full continuum of patient care, organized into eight major task categories and covering nine languages.

By leveraging a systematic task taxonomy and an open, continuously updated leaderboard, BRIDGE enables multidimensional evaluation of LLM capabilities in healthcare, across languages, clinical specialties, and inference strategies. It further provides the largest systematic assessment to date, evaluating 95 large language models, including proprietary, open-source, and medically fine-tuned variants, under zero-shot, few-shot, and chain-of-thought prompting conditions, totaling 24,795 ($95\times87\times3$) controlled experiments.

Together, this scale, diversity, and transparency enable fair, side-by-side comparisons and comprehensive analyses that prior benchmarks could not capture. A detailed comparison with existing datasets and benchmarks is presented in Supplementary Table S1.

## 2 Model information

We systematically evaluated 95 state-of-the-art large language models (LLMs) in the BRIDGE benchmark, encompassing both proprietary and open-source models. In addition, the benchmark includes *medical-specialized* LLMs that have undergone domain-specific adaptation for healthcare applications, as well as *reasoning* LLMs designed to explicitly generate intermediate reasoning traces before producing final outputs, such as the DeepSeek and Qwen3-Thinking series.

Details of all included models are summarized in Table S2. The models are listed in alphabetical order and organized by model accessibility type (Proprietary / Open-Source), model family, release date, and parameter size. Derived variants are presented at the end of their respective model families for clarity.

Recent *reasoning LLMs* introduce specialized configuration settings that control different model behaviors, resulting in considerable performance variation across modes. For example, early versions of *Qwen3* support both *thinking* and *non-thinking* modes, which are explicitly indicated in the table. Similarly, *GPT-oss* provides multiple reasoning configurations, for which we report results using the "high" reasoning mode by default. For all other models, unless otherwise specified, evaluations were conducted using their default configuration settings.

**Table S1.** Comparison of existing evaluations and BRIDGE

| Date | Study | Data source | Focus on real-world data* | Number of Language | Number of Tasks | Number of EHR Tasks | Number of Data Sources | Number of Evaluated Model | Task Category† | Inference Strategy | Online Leaderboard |
|---|---|---|---|---|---|---|---|---|---|---|---|
| 1-May-25 | BRIDGE (ours) | Electronic health record, Clinical case report, Medical consultation | Yes | 9 | 87 | 68 | 59 | 95 (Proprietary, Open-source, Medical, Reasoning) | 8: Text classification, Semantic similarity, Normalization and coding, Natural language inference (NLI), Named entity recognition (NER), Event extraction, Question-answering (QA), and Text summarization | Zero-shot, CoT, Few-shot | Yes |
| 30-Nov-22 | Agrawal *et al.* [1] | Electronic health record, Biomedical literature | No | 1: English | 5 | 4 | 3 | 4 | 2: NER, Event extraction | Zero-shot, Few-shot | No |
| 16-Feb-23 | Lehman *et al.* [2] | Electronic health record | Yes | 1: English | 3 | 3 | 3 | 7 | 3: Text classification, NLI, QA, | Few-shot, Fine-tune | No |
| 12-Jul-23 | MultiMedQA (Med-PaLM) [3] | Medical exam, Biomedical literature, Simulated medical consultation | No | 1: English | 7 | 0 | 7 | 6 | 2: Multi-choice question, QA | Zero-shot, CoT, Few-shot | No |
| 8-Sep-23 | BLURB [4] | Biomedical literature | No | 1: English | 13 | 0 | 12 | 4 | 5: Text classification, Semantic similarity, NER, Event extraction, Multi-choice question | Zero-shot, Fine-tune | No |
| 24-Mar-24 | MedAlign [5] | Electronic health record, Simulated medical consultation | Yes | 1: English | 1 | 1 | 1 | 7 | 1: QA | Zero-shot | No |
| 22-Jan-24 | Chen *et al.* [6] | Biomedical literature | No | 1: English | 2 | 0 | 2 | 4 | 2: Text classification, NLI | Zero-shot, CoT, Few-shot | No |
| 8-Mar-24 | Liévin *et al.* [7] | Medical exam, Biomedical literature | No | 1: English | 3 | 0 | 3 | 9 | 1: Multi-choice question | Zero-shot, CoT, Few-shot | No |
| 19-Apr-24 | Soroush *et al.* [8] | Electronic health record | Yes | 1: English | 3 | 3 | 1 | 4 | 1: Normalization and coding | Zero-shot | No |
| 27-Sep-24 | Qiu *et al.* [9] | Medical exam, Biomedical literature | No | 6 | 5 | 0 | 5 | 13 | 1: Multi-choice question | Zero-shot, CoT, Fine-tune | No |
| 27-Jan-25 | MedS-Bench [10] | Medical exam, Biomedical literature, Medical consultation, Electronic health record | No | 6 | 52 | 8 | 28 | 9 | 7: Multi-choice question, Text classification, NLI, NER, Event extraction, QA, Text summarization, | Few-shot | Yes |
| 6-Apr-25 | Chen *et al.* [11] | Medical exam, Biomedical literature | No | 1: English | 12 | 0 | 12 | 6 | 5: Text classification, NER, Event extraction, QA, and Text summarization | Zero-shot, Few-shot, Fine-tune | No |
| 23-Apr-25 | Tordjman *et al.* [12] | Electronic health record, Medical exam, Biomedical literature | No | 1: English | 4 | 3 | 4 | 3 | 4: Text classification, QA, Text summarization, Multi-choice question | Zero-shot | No |
| 23-Apr-25 | Sandmann *et al.* [13] | Electronic health record | Yes | 1: English | 1 | 1 | 1 | 7 | 2: Text classification, QA | Zero-shot | No |
| 13-May-25 | HealthBench [14] | Simulated medical consultation | No | 1: English | 1 | 0 | 1 | 16 | 1: QA | Zero-shot | No |
| 2-Jun-25 | MedHELM [15] | Electronic health record, Medical exam, Biomedical literature | No | 1: English | 35 | 12 | 34 | 9 | 6: Text classification, Normalization and coding, NER, Event extraction, QA, and Text summarization | Zero-shot | Yes |

** Focus on real-world clinical data: all tasks are derived from real-world clinical text, including EHRs and patient–doctor interactions.*
*† Task category: all tasks are standardized and aligned with our BRIDGE setting to facilitate comparison (See Section Methods).*

**Table S2.** Model Information of Evaluated Large Language Models

| Model Name | Release Date | Model Link | Creator | Size (B)* | Domain | Reasoning Model | Model License |
|---|---|---|---|---|---|---|---|
| Athene-V2-Chat | 11/12/2024 | link | Nexusflow | 72 | General | No | Nexusflow Research License |
| Baichuan-M1-14B-Instruct | 1/23/2025 | link | baichuan-inc | 14 | Medical | No | Baichuan-M1-14B |
| Baichuan-M2-32B | 8/10/2025 | link | baichuan-inc | 32 | Medical | Yes | Apache 2.0 |
| BioMistral-7B | 2/14/2024 | link | Nantes Université | 7 | Medical | No | Apache 2.0 |
| DeepSeek-R1 | 1/20/2025 | link | deepseek-ai | 671 | General | Yes | MIT |
| DeepSeek-R1-0528-Qwen3-8B | 5/29/2025 | link | deepseek-ai | 8 | General | Yes | MIT |
| DeepSeek-R1-Distill-Llama-70B | 1/20/2025 | link | deepseek-ai | 70 | General | Yes | MIT |
| DeepSeek-R1-Distill-Llama-8B | 1/20/2025 | link | deepseek-ai | 8 | General | Yes | MIT |
| DeepSeek-R1-Distill-Qwen-1.5B | 1/20/2025 | link | deepseek-ai | 1.5 | General | Yes | MIT |
| DeepSeek-R1-Distill-Qwen-14B | 1/20/2025 | link | deepseek-ai | 14 | General | Yes | MIT |
| DeepSeek-R1-Distill-Qwen-32B | 1/20/2025 | link | deepseek-ai | 32 | General | Yes | MIT |
| DeepSeek-R1-Distill-Qwen-7B | 1/20/2025 | link | deepseek-ai | 7 | General | Yes | MIT |
| HuatuoGPT-o1-70B | 12/27/2024 | link | FreedomIntelligence | 70 | Medical | Yes | Apache 2.0 |
| HuatuoGPT-o1-72B | 1/9/2025 | link | FreedomIntelligence | 72 | Medical | Yes | Apache 2.0 |
| HuatuoGPT-o1-7B | 1/6/2025 | link | FreedomIntelligence | 7 | Medical | Yes | Apache 2.0 |
| HuatuoGPT-o1-8B | 12/27/2024 | link | FreedomIntelligence | 8 | Medical | Yes | Apache 2.0 |
| Llama-2-13b-chat | 7/18/2023 | link | Meta-llama | 13 | General | No | llama2 |
| Llama-2-70b-chat | 7/18/2023 | link | Meta-llama | 70 | General | No | llama2 |
| Llama-2-7b-chat | 7/18/2023 | link | Meta-llama | 7 | General | No | llama2 |
| Llama-3-70B-UltraMedical | 9/2/2024 | link | Tsinghua C3I Lab | 70 | Medical | No | Llama-3 |
| Llama-3.1-70B-Instruct | 7/23/2024 | link | Meta-llama | 70 | General | No | Llama-3.1 |
| Llama-3.1-8B-Instruct | 7/23/2024 | link | Meta-llama | 8 | General | No | Llama-3.1 |
| Llama-3.1-8B-UltraMedical | 9/1/2024 | link | Tsinghua C3I Lab | 8 | Medical | No | Llama-3 |
| Llama-3.1-Nemotron-70B-Instruct-HF | 10/17/2024 | link | Nvidia | 70 | General | No | Llama-3.1 |
| Llama-3.2-1B-Instruct | 9/25/2024 | link | Meta-llama | 1 | General | No | Llama-3.1 |
| Llama-3.2-3B-Instruct | 9/25/2024 | link | Meta-llama | 3 | General | No | Llama-3.1 |
| Llama-3.3-70B-Instruct | 12/6/2024 | link | Meta-llama | 70 | General | No | Llama-3.3 |
| Llama-4-Scout-17B-16E-Instruct | 4/25/2025 | link | Meta-llama | 109 | General | No | Llama-4 |
| Llama3-OpenBioLLM-70B | 4/24/2024 | link | SAAMA | 70 | Medical | No | Llama-3 |
| Llama3-OpenBioLLM-8B | 4/20/2024 | link | SAAMA | 8 | Medical | No | Llama-3 |
| MMed-Llama-3-8B | 5/24/2024 | link | SJTU and Shanghai AI Lab | 8 | Medical | No | Llama-3 |
| Magistral-Small-2506 | 6/10/2025 | link | mistralai | 24 | General | Yes | Apache 2.0 |
| MeLLaMA-13B-chat | 6/5/2024 | link | University of Florida | 13 | Medical | No | PhysioNet Credentialed Health Data License 1.5.0 |
| MeLLaMA-70B-chat | 6/5/2024 | link | University of Florida | 70 | Medical | No | PhysioNet Credentialed Health Data License 1.5.0 |
| MedReason-8B | 4/1/2025 | link | UCSC-VLAA | 8 | Medical | Yes | Apache 2.0 |
| Ministral-8B-Instruct-2410 | 10/21/2024 | link | mistralai | 8 | General | No | MRL |
| Mistral-Large-Instruct-2411 | 11/18/2024 | link | mistralai | 123 | General | No | MRL |
| Mistral-Small-24B-Instruct-2501 | 1/30/2025 | link | mistralai | 24 | General | No | Apache 2.0 |
| Mistral-Small-3.1-24B-Instruct-2503 | 3/17/2025 | link | mistralai | 24 | General | No | Apache 2.0 |
| Mistral-Small-Instruct-2409 | 9/17/2024 | link | mistralai | 22 | General | No | MRL |
| OpenThinker3-7B | 5/27/2025 | link | open-thoughts | 7 | General | Yes | Apache 2.0 |
| Phi-3.5-MoE-instruct | 8/17/2024 | link | Microsoft | 42 | General | No | MIT |
| Phi-3.5-mini-instruct | 8/17/2024 | link | Microsoft | 4 | General | No | MIT |
| Phi-4 | 12/11/2024 | link | Microsoft | 14 | General | No | MIT |
| Phi-4-mini-instruct | 2/19/2025 | link | Microsoft | 4 | General | No | MIT |
| Phi-4-mini-reasoning | 4/29/2025 | link | Microsoft | 4 | General | Yes | MIT |
| QWQ-32B | 3/5/2025 | link | Qwen | 32 | General | Yes | Apache 2.0 |
| QwQ-32B-Preview | 11/27/2024 | link | Qwen | 32 | General | Yes | Apache 2.0 |
| Qwen2.5-1.5B-Instruct | 9/17/2024 | link | Qwen | 1.5 | General | No | Apache 2.0 |
| Qwen2.5-32B-Instruct | 9/17/2024 | link | Qwen | 32 | General | No | Apache 2.0 |
| Qwen2.5-3B-Instruct | 9/17/2024 | link | Qwen | 3 | General | No | Apache 2.0 |
| Qwen2.5-72B-Instruct | 9/16/2024 | link | Qwen | 72 | General | No | Qwen |
| Qwen2.5-7B-Instruct | 9/16/2024 | link | Qwen | 7 | General | No | Apache 2.0 |
| Qwen3-0.6B-Non-Thinking | 4/28/2025 | link | Qwen | 0.6 | General | No | Apache 2.0 |
| Qwen3-0.6B-Thinking | 4/28/2025 | link | Qwen | 0.6 | General | Yes | Apache 2.0 |
| Qwen3-1.7B-Non-Thinking | 4/28/2025 | link | Qwen | 1.7 | General | No | Apache 2.0 |
| Qwen3-1.7B-Thinking | 4/28/2025 | link | Qwen | 1.7 | General | Yes | Apache 2.0 |
| Qwen3-14B-Non-Thinking | 4/28/2025 | link | Qwen | 14 | General | No | Apache 2.0 |
| Qwen3-14B-Thinking | 4/28/2025 | link | Qwen | 14 | General | Yes | Apache 2.0 |
| Qwen3-235B-A22B-Non-Thinking | 4/28/2025 | link | Qwen | 235 | General | No | Apache 2.0 |
| Qwen3-235B-A22B-Thinking | 4/28/2025 | link | Qwen | 235 | General | Yes | Apache 2.0 |
| Qwen3-30B-A3B-Instruct-2507 | 7/28/2025 | link | Qwen | 30 | General | No | Apache 2.0 |
| Qwen3-30B-A3B-Non-Thinking | 4/28/2025 | link | Qwen | 30 | General | No | Apache 2.0 |
| Qwen3-30B-A3B-Thinking | 4/28/2025 | link | Qwen | 30 | General | Yes | Apache 2.0 |
| Qwen3-30B-A3B-Thinking-2507 | 7/29/2025 | link | Qwen | 30 | General | Yes | Apache 2.0 |
| Qwen3-32B-Non-Thinking | 4/28/2025 | link | Qwen | 32 | General | No | Apache 2.0 |
| Qwen3-32B-Thinking | 4/28/2025 | link | Qwen | 32 | General | Yes | Apache 2.0 |
| Qwen3-4B-Instruct-2507 | 8/5/2025 | link | Qwen | 4 | General | No | Apache 2.0 |
| Qwen3-4B-Non-Thinking | 4/28/2025 | link | Qwen | 4 | General | No | Apache 2.0 |
| Qwen3-4B-Thinking | 4/28/2025 | link | Qwen | 4 | General | Yes | Apache 2.0 |
| Qwen3-4B-Thinking-2507 | 8/5/2025 | link | Qwen | 4 | General | Yes | Apache 2.0 |
| Qwen3-8B-Non-Thinking | 4/28/2025 | link | Qwen | 8 | General | No | Apache 2.0 |
| Qwen3-8B-Thinking | 4/28/2025 | link | Qwen | 8 | General | Yes | Apache 2.0 |
| Qwen3-Next-80B-A3B-Instruct | 9/9/2025 | link | Qwen | 80 | General | No | Apache 2.0 |
| Qwen3-Next-80B-A3B-Thinking | 9/9/2025 | link | Qwen | 80 | General | Yes | Apache 2.0 |
| QwenLong-L1-32B | 5/23/2025 | link | Tongyi-Zhiwen | 32 | General | Yes | Apache 2.0 |
| Yi-1.5-34B-Chat-16K | 5/15/2024 | link | 01-ai | 34 | General | No | Apache 2.0 |
| Yi-1.5-9B-Chat-16K | 5/15/2024 | link | 01-ai | 9 | General | No | Apache 2.0 |
| gemini-1.5-pro-002 | 9/24/2024 | link | Google | Unknown | General | No | Proprietary |
| gemini-2.0-flash-001 | 2/5/2025 | link | Google | Unknown | General | No | Proprietary |
| gemini-2.5-flash | 5/17/2025 | link | Google | Unknown | General | Yes | Proprietary |
| gemma-2-27b-it | 6/27/2024 | link | Google | 27 | General | No | Gemma |
| gemma-2-9b-it | 6/27/2024 | link | Google | 9 | General | No | Gemma |
| gemma-3-12b-it | 3/12/2025 | link | Google | 12 | General | No | Gemma |
| gemma-3-1b-it | 3/12/2025 | link | Google | 1 | General | No | Gemma |
| gemma-3-27b-it | 3/12/2025 | link | Google | 27 | General | No | Gemma |
| gemma-3-4b-it | 3/12/2025 | link | Google | 4 | General | No | Gemma |
| gpt-35-turbo-0125 | 1/25/2024 | link | OpenAI | Unknown | General | No | Proprietary |
| gpt-4o-0806 | 8/6/2024 | link | OpenAI | Unknown | General | No | Proprietary |
| gpt-oss-120b | 8/4/2025 | link | OpenAI | 120 | General | Yes | Apache 2.0 |
| gpt-oss-20b | 8/4/2025 | link | OpenAI | 20 | General | Yes | Apache 2.0 |
| medgemma-27b-it | 5/20/2025 | link | Google | 27 | Medical | No | Health AI Developer Foundations terms of use |
| medgemma-4b-it | 5/20/2025 | link | Google | 4 | Medical | No | Health AI Developer Foundations terms of use |
| meditron-70b | 11/27/2023 | link | EPFL | 70 | Medical | No | Apache 2.0 |
| meditron-7b | 11/27/2023 | link | EPFL | 7 | Medical | No | Apache 2.0 |

** Size (B) refers to the number of model parameters in billions.*

## 3 Performance and Analysis

### 3.1 Overall Performance of BRIDGE

For each LLM, an overall performance is computed as the average score of its primary metric across all tasks, providing a holistic view of its relative performance on the BRIDGE benchmark. Table S3 presents the overall performance of LLMs under different inference strategies: zero-shot, chain-of-thought (CoT), and five-shot prompting. The primary metrics are defined as follows for each task type: (1) *Accuracy*: Text Classification, Semantic Similarity, Natural Language Inference (NLI), and Normalization and Coding (document level); (2) *Event-level F1-score*: Named Entity Recognition (NER), Event Extraction, and Normalization and Coding (entity level). (3) *ROUGE-average*: Question-Answering (QA) and Summarization tasks.

For Table S3, the columns labeled "Δ *Score few-shot*" and "Δ *Score (%) few-shot*" represent the absolute and relative improvements in performance when using few-shot prompting compared to zero-shot prompting. Similarly, "Δ *Score CoT*" and "Δ *Score (%) CoT*" show the changes under CoT prompting. Values enclosed in square brackets represent the 95% confidence intervals (CIs), computed via bootstrapping with 1,000 resamples.

## **Table S3.** Overall Performance of LLMs Across Different Inference Strategies.

| Model | Model Type | Model Domain | Score Zero-shot | Score CoT | Score 5-shot | Δ Score Few-shot | Δ Score (%) Few-shot | Δ Score CoT | Δ Score (%) CoT |
|---|---|---|---|---|---|---|---|---|---|
| gemini-2.5-flash | Proprietary | General | **44.8 [44.1, 45.6]** | **43.3 [42.5, 44.1]** | 53.4 [52.6, 54.2] | 8.6 | 19.2 | −1.5 | −3.3 |
| gpt-4o | Proprietary | General | 44.2 [43.4, 45.0] | 40.7 [39.9, 41.4] | 52.6 [51.8, 53.4] | 8.4 | 19.0 | −3.5 | −7.9 |
| DeepSeek-R1 | Open-Source | General | 44.2 [43.5, 45.0] | 42.1 [41.3, 42.9] | 51.4 [50.6, 52.2] | 7.2 | 16.3 | −2.1 | −4.8 |
| Qwen3-Next-80B-A3B-Thinking | Open-Source | General | 43.9 [43.1, 44.6] | 42.9 [42.1, 43.7] | 50.9 [50.1, 51.7] | 7.0 | 15.9 | −1.0 | −2.3 |
| gemini-1.5-pro | Proprietary | General | 43.8 [43.1, 44.6] | 40.5 [39.7, 41.3] | **55.5 [54.7, 56.3]** | 11.7 | 26.7 | −3.3 | −7.5 |
| gemini-2.0-flash | Proprietary | General | 43.0 [42.2, 43.8] | 42.0 [41.2, 42.8] | 53.3 [52.5, 54.2] | 10.3 | 24.0 | −1.0 | −2.3 |
| Mistral-Large-Instruct-2411 | Open-Source | General | 42.3 [41.5, 43.1] | 38.9 [38.1, 39.7] | 50.7 [49.8, 51.5] | 8.4 | 19.9 | −3.4 | −8.0 |
| Qwen3-30B-A3B-Thinking-2507 | Open-Source | General | 41.8 [41.0, 42.6] | 41.4 [40.6, 42.2] | 50.3 [49.4, 51.1] | 8.5 | 20.3 | −0.4 | −1.0 |
| Athene-V2-Chat | Open-Source | General | 41.7 [40.9, 42.5] | 39.3 [38.6, 40.1] | 50.7 [49.8, 51.5] | 9.0 | 21.6 | −2.4 | −5.8 |
| Qwen3-235B-A22B-Thinking | Open-Source | General | 41.6 [40.8, 42.5] | 40.1 [39.4, 40.9] | 48.1 [47.3, 49.0] | 6.5 | 15.6 | −1.5 | −3.6 |
| Qwen2.5-72B-Instruct | Open-Source | General | 41.6 [40.8, 42.4] | 38.9 [38.1, 39.7] | 51.0 [50.2, 51.8] | 9.4 | 22.6 | −2.7 | −6.5 |
| Qwen3-32B-Thinking | Open-Source | General | 41.0 [40.3, 41.8] | 34.5 [33.7, 35.3] | 47.4 [46.6, 48.2] | 6.4 | 15.6 | −6.5 | −15.9 |
| HuatuoGPT-o1-72B | Open-Source | Medical | 41.0 [40.2, 41.8] | 38.1 [37.4, 38.9] | 50.2 [49.4, 51.1] | 9.2 | 22.4 | −2.9 | −7.1 |
| Qwen3-30B-A3B-Thinking | Open-Source | General | 40.9 [40.1, 41.8] | 39.4 [38.6, 40.1] | 47.0 [46.2, 47.8] | 6.1 | 14.9 | −1.5 | −3.7 |
| medgemma-27b-it | Open-Source | Medical | 40.8 [40.0, 41.6] | 38.2 [37.5, 38.9] | 52.0 [51.1, 52.8] | 11.2 | 27.5 | −2.6 | −6.4 |
| Qwen3-14B-Thinking | Open-Source | General | 40.2 [39.4, 41.0] | 37.1 [36.3, 37.9] | 46.8 [45.8, 47.8] | 6.6 | 16.4 | −3.1 | −7.7 |
| Qwen3-8B-Thinking | Open-Source | General | 40.0 [39.2, 40.8] | 37.1 [36.2, 37.9] | 45.2 [44.4, 46.0] | 5.2 | 13.0 | −2.9 | −7.2 |
| Qwen3-4B-Thinking-2507 | Open-Source | General | 40.0 [39.2, 40.8] | 38.8 [38.0, 39.7] | 47.0 [46.1, 47.8] | 7.0 | 17.5 | −1.2 | −3.0 |
| gemma-3-27b-it | Open-Source | General | 39.9 [39.1, 40.7] | 37.5 [36.8, 38.3] | 49.5 [48.6, 50.3] | 9.6 | 24.1 | −2.4 | −6.0 |
| Qwen2.5-32B-Instruct | Open-Source | General | 39.9 [39.1, 40.7] | 37.7 [36.9, 38.5] | 49.5 [48.7, 50.4] | 9.6 | 24.1 | −2.2 | −5.5 |
| Llama-3.3-70B-Instruct | Open-Source | General | 39.9 [39.1, 40.7] | 36.8 [36.1, 37.6] | 49.5 [48.7, 50.3] | 9.6 | 24.1 | −3.1 | −7.8 |
| Qwen3-Next-80B-A3B-Instruct | Open-Source | General | 39.8 [39.0, 40.6] | 40.5 [39.7, 41.3] | 50.6 [49.7, 51.4] | 10.8 | 27.1 | 0.7 | 1.8 |
| DeepSeek-R1-Distill-Llama-70B | Open-Source | General | 39.8 [39.0, 40.6] | 38.9 [38.2, 39.7] | 46.2 [45.3, 47.0] | 6.4 | 16.1 | −0.9 | −2.3 |
| Mistral-Small-3.1-24B-Instruct-2503 | Open-Source | General | 39.7 [38.9, 40.6] | 36.2 [35.4, 37.0] | 40.9 [40.2, 41.7] | 1.2 | 3.0 | −3.5 | −8.8 |
| DeepSeek-R1-Distill-Qwen-32B | Open-Source | General | 39.7 [39.0, 40.5] | 38.7 [37.9, 39.5] | 44.3 [43.5, 45.2] | 4.6 | 11.6 | −1.0 | −2.5 |
| QWQ-32B | Open-Source | General | 39.4 [38.6, 40.2] | 37.0 [36.2, 37.8] | 48.2 [47.4, 49.0] | 8.8 | 22.3 | −2.4 | −6.1 |
| Qwen3-32B-Non-Thinking | Open-Source | General | 39.3 [38.5, 40.1] | 32.8 [32.0, 33.6] | 48.9 [48.1, 49.8] | 9.6 | 24.4 | −6.5 | −16.5 |
| Qwen3-235B-A22B-Non-Thinking | Open-Source | General | 39.2 [38.4, 40.0] | 38.0 [37.2, 38.8] | 48.7 [47.9, 49.6] | 9.5 | 24.2 | −1.2 | −3.1 |
| QwenLong-L1-32B | Open-Source | General | 39.2 [38.4, 40.1] | 36.7 [35.9, 37.5] | 43.4 [42.6, 44.3] | 4.2 | 10.7 | −2.5 | −6.4 |
| Llama-3.1-70B-Instruct | Open-Source | General | 39.1 [38.3, 39.9] | 35.1 [34.3, 35.9] | 50.5 [49.7, 51.4] | 11.4 | 29.2 | −4.0 | −10.2 |
| Qwen3-4B-Thinking | Open-Source | General | 38.5 [37.7, 39.3] | 37.0 [36.2, 37.8] | 42.7 [41.8, 43.6] | 4.2 | 10.9 | −1.5 | −3.9 |
| gemma-2-27b-it | Open-Source | General | 38.2 [37.4, 39.0] | 34.2 [33.4, 35.0] | 45.8 [44.9, 46.7] | 7.6 | 19.9 | −4.0 | −10.5 |
| Baichuan-M2-32B | Open-Source | Medical | 38.0 [37.0, 39.0] | 38.3 [37.4, 39.2] | 45.2 [44.1, 46.3] | 7.2 | 18.9 | 0.3 | 0.8 |
| Magistral-Small-2506 | Open-Source | General | 37.6 [36.8, 38.4] | 34.2 [33.3, 35.0] | 49.1 [48.3, 50.0] | 11.5 | 30.6 | −3.4 | −9.0 |
| Mistral-Small-24B-Instruct-2501 | Open-Source | General | 37.5 [36.7, 38.4] | 31.6 [30.8, 32.4] | 39.7 [38.9, 40.4] | 2.2 | 5.9 | −5.9 | −15.7 |
| gemma-3-12b-it | Open-Source | General | 37.3 [36.6, 38.1] | 35.4 [34.6, 36.1] | 47.7 [46.9, 48.6] | 10.4 | 27.9 | −1.9 | −5.1 |
| gpt-oss-120b-high | Open-Source | General | 37.2 [36.4, 38.1] | 32.1 [31.4, 32.9] | 39.0 [38.2, 39.9] | 1.8 | 4.8 | −5.1 | −13.7 |
| Qwen3-30B-A3B-Instruct-2507 | Open-Source | General | 36.9 [36.1, 37.6] | 37.5 [36.7, 38.3] | 48.6 [47.8, 49.5] | 11.7 | 31.7 | 0.6 | 1.6 |
| DeepSeek-R1-0528-Qwen3-8B | Open-Source | General | 36.9 [36.1, 37.8] | 36.3 [35.5, 37.1] | 42.9 [42.1, 43.8] | 6.0 | 16.3 | −0.6 | −1.6 |
| Qwen3-14B-Non-Thinking | Open-Source | General | 36.8 [36.0, 37.6] | 34.6 [33.8, 35.4] | 46.6 [45.8, 47.5] | 9.8 | 26.6 | −2.2 | −6.0 |
| Qwen3-30B-A3B-Non-Thinking | Open-Source | General | 36.2 [35.4, 36.9] | 36.9 [36.1, 37.7] | 47.4 [46.5, 48.2] | 11.2 | 30.9 | 0.7 | 1.9 |
| Phi-4 | Open-Source | General | 36.1 [35.4, 36.9] | 32.6 [31.8, 33.4] | 46.8 [45.9, 47.7] | 10.7 | 29.6 | −3.5 | −9.7 |
| Baichuan-M1-14B-Instruct | Open-Source | Medical | 36.1 [35.3, 36.9] | 34.4 [33.6, 35.2] | 48.3 [47.4, 49.2] | 12.2 | 33.8 | −1.7 | −4.7 |
| gpt-35-turbo | Proprietary | General | 35.3 [34.5, 36.1] | 31.6 [30.8, 32.4] | 43.6 [42.7, 44.5] | 8.3 | 23.5 | −3.7 | −10.5 |
| Mistral-Small-Instruct-2409 | Open-Source | General | 35.2 [34.4, 36.0] | 31.2 [30.4, 32.0] | 36.0 [35.1, 36.9] | 0.8 | 2.3 | −4.0 | −11.4 |
| gemma-2-9b-it | Open-Source | General | 35.1 [34.3, 35.9] | 29.9 [29.2, 30.7] | 43.5 [42.7, 44.4] | 8.4 | 23.9 | −5.2 | −14.8 |
| Llama-4-Scout-17B-16E-Instruct | Open-Source | General | 35.1 [34.4, 35.9] | 29.4 [28.7, 30.1] | 40.6 [39.8, 41.4] | 5.5 | 15.7 | −5.7 | −16.2 |
| DeepSeek-R1-Distill-Qwen-14B | Open-Source | General | 34.3 [33.5, 35.1] | 34.8 [34.0, 35.6] | 41.4 [40.6, 42.2] | 7.1 | 20.7 | 0.5 | 1.5 |
| Qwen3-4B-Instruct-2507 | Open-Source | General | 34.1 [33.3, 34.9] | 36.2 [35.4, 37.0] | 44.8 [44.0, 45.7] | 10.7 | 31.4 | 2.1 | 6.2 |
| HuatuoGPT-o1-70B | Open-Source | Medical | 33.9 [33.2, 34.6] | 32.3 [31.6, 33.1] | 43.1 [42.3, 43.9] | 9.2 | 27.1 | −1.6 | −4.7 |
| Qwen3-8B-Non-Thinking | Open-Source | General | 33.9 [33.1, 34.7] | 33.3 [32.5, 34.1] | 45.5 [44.7, 46.4] | 11.6 | 34.2 | −0.6 | −1.8 |
| Llama-3-70B-UltraMedical | Open-Source | Medical | 33.4 [32.6, 34.2] | 29.4 [28.7, 30.2] | 44.4 [43.6, 45.3] | 11.0 | 32.9 | −4.0 | −12.0 |
| Qwen3-4B-Non-Thinking | Open-Source | General | 33.3 [32.5, 34.1] | 33.2 [32.4, 34.0] | 42.8 [41.9, 43.6] | 9.5 | 28.5 | −0.1 | −0.3 |
| Llama3-OpenBioLLM-70B | Open-Source | Medical | 33.0 [32.2, 33.8] | 28.8 [28.0, 29.5] | 44.5 [43.6, 45.4] | 11.5 | 34.8 | −4.2 | −12.7 |
| Llama-3.1-Nemotron-70B-Instruct-HF | Open-Source | General | 32.7 [32.0, 33.5] | 24.1 [23.5, 24.7] | 42.8 [42.0, 43.6] | 10.1 | 30.9 | −8.6 | −26.3 |
| MeLLaMA-70B-chat | Open-Source | Medical | 32.3 [31.5, 33.1] | 29.2 [28.4, 30.1] | 32.9 [32.1, 33.7] | 0.6 | 1.9 | −3.1 | −9.6 |
| Yi-1.5-34B-Chat-16K | Open-Source | General | 32.1 [31.4, 32.9] | 29.6 [28.8, 30.3] | 41.0 [40.1, 41.8] | 8.9 | 27.7 | −2.5 | −7.8 |
| QwQ-32B-Preview | Open-Source | General | 31.7 [31.0, 32.5] | 23.3 [22.5, 24.1] | 48.4 [47.6, 49.3] | 16.7 | 52.7 | −8.4 | −26.5 |
| Qwen2.5-7B-Instruct | Open-Source | General | 31.3 [30.5, 32.1] | 30.2 [29.5, 31.0] | 41.6 [40.7, 42.5] | 10.3 | 32.9 | −1.1 | −3.5 |
| Ministral-8B-Instruct-2410 | Open-Source | General | 30.4 [29.5, 31.2] | 25.9 [25.1, 26.7] | 39.6 [38.7, 40.6] | 9.2 | 30.3 | −4.5 | −14.8 |
| HuatuoGPT-o1-7B | Open-Source | Medical | 29.6 [28.8, 30.4] | 26.7 [26.0, 27.4] | 40.4 [39.5, 41.3] | 10.8 | 36.5 | −2.9 | −9.8 |
| Phi-3.5-MoE-instruct | Open-Source | General | 29.5 [28.8, 30.3] | 25.3 [24.5, 26.0] | 36.6 [35.7, 37.4] | 7.1 | 24.1 | −4.2 | −14.2 |
| medgemma-4b-it | Open-Source | Medical | 29.4 [28.3, 30.5] | 28.5 [27.6, 29.3] | 37.5 [36.4, 38.5] | 8.1 | 27.6 | −0.9 | −3.1 |
| Llama-3.1-8B-Instruct | Open-Source | General | 29.0 [28.2, 29.8] | 29.4 [28.6, 30.2] | 43.5 [42.7, 44.4] | 14.5 | 50.0 | 0.4 | 1.4 |
| gpt-oss-20b-high | Open-Source | General | 29.0 [28.3, 29.8] | 24.9 [24.1, 25.6] | 25.1 [24.3, 26.0] | −3.9 | −13.4 | −4.1 | −14.1 |
| Yi-1.5-9B-Chat-16K | Open-Source | General | 28.8 [28.0, 29.6] | 25.4 [24.6, 26.2] | 37.4 [36.5, 38.3] | 8.6 | 29.9 | −3.4 | −11.8 |
| gemma-3-4b-it | Open-Source | General | 28.6 [27.8, 29.4] | 28.2 [27.4, 29.0] | 38.8 [37.9, 39.7] | 10.2 | 35.7 | −0.4 | −1.4 |
| DeepSeek-R1-Distill-Llama-8B | Open-Source | General | 28.5 [27.7, 29.2] | 27.3 [26.6, 28.1] | 32.9 [32.0, 33.7] | 4.4 | 15.4 | −1.2 | −4.2 |
| Qwen2.5-3B-Instruct | Open-Source | General | 26.6 [25.8, 27.3] | 25.4 [24.7, 26.2] | 37.2 [36.3, 38.0] | 10.6 | 39.8 | −1.2 | −4.5 |
| Qwen3-1.7B-Thinking | Open-Source | General | 26.3 [25.5, 27.0] | 27.7 [27.0, 28.4] | 32.9 [32.0, 33.7] | 6.6 | 25.1 | 1.4 | 5.3 |
| HuatuoGPT-o1-8B | Open-Source | Medical | 25.9 [25.1, 26.6] | 22.5 [21.8, 23.2] | 40.4 [39.5, 41.3] | 14.5 | 56.0 | −3.4 | −13.1 |
| Phi-3.5-mini-instruct | Open-Source | General | 25.4 [24.7, 26.1] | 23.9 [23.2, 24.6] | 31.3 [30.5, 32.1] | 5.9 | 23.2 | −1.5 | −5.9 |
| DeepSeek-R1-Distill-Qwen-7B | Open-Source | General | 25.3 [24.5, 26.0] | 23.9 [23.1, 24.7] | 23.1 [22.3, 23.9] | −2.2 | −8.7 | −1.4 | −5.5 |
| Llama-2-70b-chat | Open-Source | General | 25.2 [24.4, 25.9] | 19.0 [18.3, 19.7] | 27.1 [26.3, 28.0] | 1.9 | 7.5 | −6.2 | −24.6 |
| Phi-4-mini-instruct | Open-Source | General | 24.5 [23.8, 25.3] | 22.5 [21.8, 23.2] | 29.9 [29.1, 30.8] | 5.4 | 22.0 | −2.0 | −8.2 |
| Llama-3.2-3B-Instruct | Open-Source | General | 22.9 [22.1, 23.7] | 21.6 [20.9, 22.3] | 34.8 [33.9, 35.7] | 11.9 | 52.0 | −1.3 | −5.7 |
| Qwen2.5-1.5B-Instruct | Open-Source | General | 22.2 [21.4, 22.9] | 19.5 [18.7, 20.2] | 31.7 [30.8, 32.6] | 9.5 | 42.8 | −2.7 | −12.2 |
| Qwen3-1.7B-Non-Thinking | Open-Source | General | 22.0 [21.2, 22.7] | 22.8 [22.0, 23.5] | 34.8 [33.9, 35.6] | 12.8 | 58.2 | 0.8 | 3.6 |
| Phi-4-mini-reasoning | Open-Source | General | 21.3 [20.6, 22.0] | 19.8 [19.1, 20.5] | 15.5 [14.8, 16.2] | −5.8 | −27.2 | −1.5 | −7.0 |
| Llama-2-13b-chat | Open-Source | General | 20.9 [20.2, 21.7] | 16.2 [15.5, 16.9] | 23.5 [22.7, 24.4] | 2.6 | 12.4 | −4.7 | −22.5 |
| MeLLaMA-13B-chat | Open-Source | Medical | 20.8 [20.0, 21.5] | 20.3 [19.5, 21.0] | 27.0 [26.2, 27.8] | 6.2 | 29.8 | −0.5 | −2.4 |
| MMed-Llama-3-8B | Open-Source | Medical | 20.4 [19.7, 21.1] | 16.2 [15.4, 16.9] | 34.1 [33.3, 35.0] | 13.7 | 67.2 | −4.2 | −20.6 |
| Qwen3-0.6B-Thinking | Open-Source | General | 20.4 [19.6, 21.1] | 18.9 [18.2, 19.7] | 24.9 [24.1, 25.7] | 4.5 | 22.1 | −1.5 | −7.4 |
| BioMistral-7B | Open-Source | Medical | 20.4 [19.7, 21.2] | 10.8 [10.2, 11.5] | 24.7 [23.8, 25.5] | 4.3 | 21.1 | −9.6 | −47.1 |
| Llama-3.1-8B-UltraMedical | Open-Source | Medical | 20.2 [19.5, 20.9] | 18.3 [17.7, 19.0] | 31.0 [30.1, 31.8] | 10.8 | 53.5 | −1.9 | −9.4 |
| OpenThinker3-7B | Open-Source | General | 19.8 [19.0, 20.5] | 18.4 [17.7, 19.1] | 13.0 [12.3, 13.7] | −6.8 | −34.3 | −1.4 | −7.1 |
| MedReason-8B | Open-Source | Medical | 18.2 [17.4, 18.9] | 18.3 [17.6, 19.0] | 31.1 [30.2, 32.0] | 12.9 | 70.9 | 0.1 | 0.5 |
| Llama-2-7b-chat | Open-Source | General | 16.5 [15.8, 17.2] | 13.7 [13.0, 14.3] | 22.3 [21.5, 23.1] | 5.8 | 35.2 | −2.8 | −17.0 |
| gemma-3-1b-it | Open-Source | General | 15.7 [15.0, 16.4] | 13.5 [12.8, 14.2] | 25.9 [25.1, 26.7] | 10.2 | 65.0 | −2.2 | −14.0 |
| meditron-70b | Open-Source | Medical | 15.7 [15.0, 16.4] | 13.2 [12.5, 13.9] | 32.1 [31.3, 32.9] | 16.4 | 104.5 | −2.5 | −15.9 |
| Qwen3-0.6B-Non-Thinking | Open-Source | General | 15.2 [14.5, 15.9] | 16.6 [15.9, 17.3] | 26.3 [25.5, 27.1] | 11.1 | 73.0 | 1.4 | 9.2 |
| DeepSeek-R1-Distill-Qwen-1.5B | Open-Source | General | 14.3 [13.6, 15.0] | 13.4 [12.7, 14.1] | 14.9 [14.2, 15.7] | 0.6 | 4.2 | −0.9 | −6.3 |
| Llama3-OpenBioLLM-8B | Open-Source | Medical | 14.2 [13.5, 14.9] | 13.3 [12.6, 14.0] | 33.1 [32.1, 34.1] | 18.9 | 133.1 | −0.9 | −6.3 |
| Llama-3.2-1B-Instruct | Open-Source | General | 12.7 [12.1, 13.4] | 11.9 [11.2, 12.5] | 24.4 [23.6, 25.3] | 11.7 | 92.1 | −0.8 | −6.3 |
| meditron-7b | Open-Source | Medical | 9.5 [8.9, 10.2] | 9.5 [8.9, 10.1] | 17.0 [16.3, 17.7] | 7.5 | 78.9 | 0.0 | 0.0 |

*Note: Models are ranked by the average score across all tasks under zero-shot inference ("Score Zero-shot") in descending order. The **bold blue** highlights the highest value for each score column. Δ represents the difference compared to zero-shot, with green indicating improvement and red indicating a decline.*

### 3.2 Subgroup Performance Across Different Task Types

Table S4 shows the performance of LLMs across different task types under zero-shot setting.

### 3.3 Subgroup Performance Across Different Languages

Table S5 shows the performance of LLMs across different language types under zero-shot setting.

### 3.4 Subgroup Performance Across Clinical Specialties

Table S6 shows the performance of LLMs across different clinical specialties under zero-shot prompting

### 3.5 Subgroup Performance Across Clinical Stages

Table S7 shows the performance of LLMs across different clinical specialties under zero-shot prompting

**Table S4.** Subgroup Performance of LLMs Across Different Task Types (Zero-shot)

| Model | Model Type | Model Domain | Score Avg | Score CLF | Score EE | Score NER | Score NLI | Score NOR | Score QA | Score SS | Score SUM |
|---|---|---|---|---|---|---|---|---|---|---|---|
| gpt-4o | Proprietary | General | **42.2** | 69.2 | 29.1 | 45.3 | 87.4 | 8.6 | 16.8 | 47.9 | 33.5 |
| DeepSeek-R1 | Open-Source | General | 41.1 | 68.4 | 30.6 | 46.2 | **87.6** | **9.9** | 16.1 | 35.9 | 33.8 |
| gemini-1.5-pro | Proprietary | General | 40.9 | 68.6 | 30.5 | 45.0 | 84.7 | 7.7 | 16.9 | 41.0 | 33.2 |
| Mistral-Large-Instruct-2411 | Open-Source | General | 40.5 | 66.0 | 27.1 | 43.9 | 85.0 | 7.2 | 18.0 | 43.3 | 33.2 |
| gemini-2.5-flash | Proprietary | General | 40.5 | 68.3 | **31.8** | **49.4** | 85.7 | 9.3 | 16.7 | 30.2 | 32.8 |
| gemini-2.0-flash | Proprietary | General | 40.3 | 69.5 | 26.2 | 44.8 | 84.7 | 6.8 | **18.4** | 38.5 | 33.6 |
| Athene-V2-Chat | Open-Source | General | 40.3 | 68.2 | 27.9 | 38.3 | 83.8 | 4.5 | 17.4 | 48.1 | 33.7 |
| Qwen2.5-72B-Instruct | Open-Source | General | 40.1 | 68.5 | 27.7 | 38.0 | 83.4 | 4.3 | 17.8 | 47.6 | 33.8 |
| Qwen3-Next-80B-A3B-Thinking | Open-Source | General | 40.1 | 70.3 | 29.0 | 46.0 | 84.9 | 6.9 | 14.9 | 35.8 | 33.2 |
| HuatuoGPT-o1-72B | Open-Source | Medical | 39.5 | 67.9 | 27.5 | 37.2 | 83.9 | 4.4 | 17.0 | **48.5** | 29.6 |
| medgemma-27b-it | Open-Source | Medical | 39.2 | 66.9 | 24.4 | 41.0 | 82.2 | 3.3 | 16.7 | 44.6 | 34.1 |
| Qwen3-235B-A22B-Thinking | Open-Source | General | 39.1 | 68.4 | 24.9 | 42.5 | 85.3 | 7.1 | 15.1 | 37.9 | 31.4 |
| Qwen3-30B-A3B-Thinking-2507 | Open-Source | General | 38.9 | 69.8 | 26.9 | 39.2 | 84.7 | 7.7 | 16.8 | 32.6 | 33.8 |
| Qwen3-32B-Thinking | Open-Source | General | 38.9 | 67.9 | 24.8 | 40.8 | 85.4 | 5.8 | 15.8 | 40.2 | 30.5 |
| Qwen2.5-32B-Instruct | Open-Source | General | 38.8 | 64.8 | 24.7 | 39.2 | 83.9 | 3.6 | 17.3 | 46.5 | 30.5 |
| Qwen3-32B-Non-Thinking | Open-Source | General | 38.7 | 65.5 | 23.0 | 36.2 | 86.5 | 5.0 | 17.8 | 44.6 | 30.8 |
| gemma-3-27b-it | Open-Source | General | 38.7 | 63.7 | 24.2 | 40.9 | 83.4 | 3.8 | 16.3 | 43.2 | 33.6 |
| Mistral-Small-3.1-24B-Instruct-2503 | Open-Source | General | 38.7 | 65.0 | 25.1 | 36.8 | 86.6 | 5.1 | 16.7 | 41.9 | 32.6 |
| Llama-3.3-70B-Instruct | Open-Source | General | 38.6 | 65.3 | 24.2 | 38.5 | 83.8 | 5.3 | 16.7 | 41.9 | 32.8 |
| Qwen3-235B-A22B-Non-Thinking | Open-Source | General | 38.5 | 66.8 | 21.3 | 36.6 | 83.8 | 4.2 | 16.5 | 46.9 | 31.9 |
| DeepSeek-R1-Distill-Qwen-32B | Open-Source | General | 38.5 | 65.8 | 22.3 | 39.3 | 84.0 | 5.3 | 16.9 | 41.5 | 32.6 |
| Qwen3-Next-80B-A3B-Instruct | Open-Source | General | 38.5 | 67.4 | 22.5 | 37.8 | 82.8 | 5.7 | 16.2 | 45.7 | 30.1 |
| DeepSeek-R1-Distill-Llama-70B | Open-Source | General | 38.4 | 67.2 | 22.1 | 38.0 | 85.0 | 6.1 | 16.4 | 39.7 | 32.2 |
| Llama-3.1-70B-Instruct | Open-Source | General | 38.2 | 65.9 | 21.0 | 37.0 | 81.3 | 6.6 | 17.9 | 43.3 | 32.9 |
| QWQ-32B | Open-Source | General | 38.2 | 66.9 | 20.6 | 38.1 | 87.3 | 5.6 | 15.4 | 39.8 | 32.1 |
| Qwen3-30B-A3B-Thinking | Open-Source | General | 38.2 | 68.4 | 26.1 | 39.3 | 83.8 | 6.6 | 16.3 | 37.7 | 27.3 |
| QwenLong-L1-32B | Open-Source | General | 38.2 | 67.1 | 21.7 | 36.2 | 84.9 | 5.1 | 15.4 | 42.6 | 32.7 |
| Qwen3-14B-Thinking | Open-Source | General | 38.2 | 67.9 | 22.8 | 39.4 | 83.3 | 5.2 | 15.5 | 40.7 | 30.8 |
| Qwen3-8B-Thinking | Open-Source | General | 37.8 | 67.4 | 23.9 | 38.6 | 81.3 | 4.9 | 15.8 | 39.9 | 30.6 |
| gemma-2-27b-it | Open-Source | General | 37.6 | 61.0 | 24.2 | 36.3 | 80.6 | 5.1 | 17.2 | 42.1 | **34.7** |
| Qwen3-4B-Thinking-2507 | Open-Source | General | 37.3 | 67.7 | 24.3 | 38.7 | 82.4 | 3.8 | 17.4 | 32.3 | 32.1 |
| Baichuan-M2-32B | Open-Source | Medical | 37.1 | 66.6 | 21.6 | 32.0 | 85.1 | 4.9 | 16.7 | 37.7 | 32.0 |
| Qwen3-14B-Non-Thinking | Open-Source | General | 36.9 | 63.8 | 20.2 | 31.3 | 83.5 | 3.3 | 17.2 | 44.4 | 31.1 |
| Mistral-Small-24B-Instruct-2501 | Open-Source | General | 36.9 | 58.7 | 25.1 | 35.6 | 84.9 | 5.0 | 16.0 | 37.3 | 32.6 |
| Qwen3-4B-Thinking | Open-Source | General | 36.6 | 65.7 | 23.0 | 35.6 | 82.7 | 4.5 | 16.0 | 36.5 | 29.2 |
| Qwen3-30B-A3B-Instruct-2507 | Open-Source | General | 36.6 | 63.7 | 20.1 | 32.5 | 85.9 | 3.4 | 15.6 | 41.3 | 30.3 |
| gemma-3-12b-it | Open-Source | General | 36.3 | 61.9 | 21.0 | 36.4 | 81.4 | 2.5 | 17.3 | 36.3 | 33.7 |
| Magistral-Small-2506 | Open-Source | General | 36.3 | 64.5 | 22.0 | 33.6 | 79.8 | 3.6 | 17.3 | 37.6 | 32.0 |
| Phi-4 | Open-Source | General | 36.2 | 62.5 | 19.1 | 31.2 | 85.1 | 5.2 | 15.6 | 41.6 | 29.2 |
| Qwen3-30B-A3B-Non-Thinking | Open-Source | General | 36.1 | 62.5 | 20.3 | 30.6 | 81.3 | 3.5 | 18.1 | 42.4 | 30.0 |
| DeepSeek-R1-0528-Qwen3-8B | Open-Source | General | 35.7 | 64.3 | 19.8 | 33.7 | 81.2 | 3.9 | 16.4 | 36.3 | 30.3 |
| Baichuan-M1-14B-Instruct | Open-Source | Medical | 35.6 | 64.5 | 19.3 | 29.0 | 80.0 | 5.8 | 16.2 | 38.1 | 31.7 |
| Llama-4-Scout-17B-16E-Instruct | Open-Source | General | 35.5 | 63.7 | 18.6 | 25.8 | 81.9 | 4.8 | 15.9 | 43.4 | 30.0 |
| gpt-35-turbo | Proprietary | General | 35.5 | 60.3 | 21.8 | 27.9 | 76.0 | 5.4 | 17.1 | 45.5 | 29.8 |
| Mistral-Small-Instruct-2409 | Open-Source | General | 35.4 | 62.4 | 18.3 | 28.3 | 75.8 | 5.4 | 17.4 | 43.8 | 31.7 |
| Qwen3-8B-Non-Thinking | Open-Source | General | 35.1 | 61.7 | 15.5 | 26.1 | 81.1 | 1.8 | 17.9 | 46.0 | 31.2 |
| gpt-oss-120b-high | Open-Source | General | 35.0 | **70.9** | 20.3 | 28.4 | 86.9 | 6.3 | 8.0 | 34.7 | 25.0 |
| gemma-2-9b-it | Open-Source | General | 34.9 | 58.2 | 19.8 | 32.7 | 76.1 | 2.8 | 16.4 | 38.6 | 34.3 |
| Qwen3-4B-Instruct-2507 | Open-Source | General | 34.8 | 61.4 | 17.3 | 27.0 | 81.9 | 1.3 | 13.6 | 45.8 | 29.6 |
| DeepSeek-R1-Distill-Qwen-14B | Open-Source | General | 34.5 | 64.2 | 11.9 | 29.3 | 81.6 | 4.0 | 16.6 | 36.8 | 31.4 |
| Qwen3-4B-Non-Thinking | Open-Source | General | 33.8 | 58.3 | 19.2 | 26.1 | 76.8 | 1.1 | 17.3 | 42.6 | 29.0 |
| Llama-3-70B-UltraMedical | Open-Source | Medical | 33.4 | 62.7 | 14.3 | 27.3 | 76.6 | 3.2 | 12.5 | 43.4 | 27.2 |
| QwQ-32B-Preview | Open-Source | General | 33.3 | 59.5 | 12.2 | 23.9 | 85.7 | 3.5 | 11.9 | 41.3 | 28.6 |
| Llama-3.1-Nemotron-70B-Instruct-HF | Open-Source | General | 33.2 | 65.6 | 9.9 | 25.4 | 82.0 | 1.8 | 10.5 | 44.8 | 25.8 |
| Yi-1.5-34B-Chat-16K | Open-Source | General | 32.9 | 62.1 | 14.7 | 20.9 | 73.8 | 2.0 | 15.8 | 42.1 | 32.0 |
| Llama3-OpenBioLLM-70B | Open-Source | Medical | 32.9 | 62.7 | 17.5 | 21.7 | 72.9 | 2.8 | 17.8 | 38.5 | 29.3 |
| Qwen2.5-7B-Instruct | Open-Source | General | 32.8 | 56.3 | 13.5 | 25.3 | 77.7 | 1.4 | 16.8 | 40.8 | 30.3 |
| MeLLaMA-70B-chat | Open-Source | Medical | 32.5 | 59.3 | 19.5 | 20.9 | 80.4 | 2.4 | 15.5 | 35.7 | 26.3 |
| Ministral-8B-Instruct-2410 | Open-Source | General | 31.7 | 52.5 | 16.2 | 23.4 | 75.9 | 1.8 | 17.4 | 35.1 | 31.1 |
| HuatuoGPT-o1-70B | Open-Source | Medical | 31.7 | 62.2 | 17.2 | 31.9 | 82.6 | 3.8 | 5.3 | 34.6 | 15.9 |
| HuatuoGPT-o1-7B | Open-Source | Medical | 31.1 | 55.6 | 9.6 | 24.5 | 77.6 | 1.9 | 13.4 | 43.0 | 23.7 |
| Phi-3.5-MoE-instruct | Open-Source | General | 30.9 | 55.5 | 13.1 | 21.4 | 77.3 | 3.0 | 10.1 | 45.2 | 21.6 |
| medgemma-4b-it | Open-Source | Medical | 30.7 | 55.4 | 13.4 | 19.7 | 74.7 | 1.5 | 15.9 | 37.1 | 27.9 |
| Llama-3.1-8B-Instruct | Open-Source | General | 30.5 | 53.8 | 9.9 | 22.8 | 72.3 | 2.8 | 16.5 | 34.1 | 31.4 |
| Yi-1.5-9B-Chat-16K | Open-Source | General | 30.4 | 52.9 | 13.5 | 19.9 | 73.0 | 1.3 | 16.0 | 37.1 | 29.7 |
| gemma-3-4b-it | Open-Source | General | 30.0 | 50.1 | 14.8 | 21.5 | 76.2 | 1.0 | 15.8 | 32.5 | 27.9 |
| gpt-oss-20b-high | Open-Source | General | 29.7 | 62.8 | 12.3 | 12.1 | 86.5 | 2.8 | 5.9 | 32.5 | 22.3 |
| DeepSeek-R1-Distill-Llama-8B | Open-Source | General | 29.3 | 56.1 | 8.6 | 21.3 | 73.2 | 1.8 | 14.6 | 32.0 | 26.7 |
| Qwen3-1.7B-Thinking | Open-Source | General | 28.4 | 50.3 | 8.6 | 17.8 | 75.9 | 2.1 | 15.8 | 31.9 | 24.7 |
| Qwen2.5-3B-Instruct | Open-Source | General | 28.2 | 48.5 | 11.1 | 19.9 | 73.3 | 0.7 | 16.8 | 31.7 | 23.9 |
| Phi-4-mini-instruct | Open-Source | General | 27.9 | 47.6 | 9.2 | 11.9 | 73.6 | 1.2 | 16.2 | 36.5 | 27.2 |
| DeepSeek-R1-Distill-Qwen-7B | Open-Source | General | 27.2 | 49.0 | 11.0 | 13.8 | 71.8 | 1.2 | 15.2 | 32.3 | 23.4 |
| Phi-3.5-mini-instruct | Open-Source | General | 27.1 | 51.3 | 7.8 | 16.1 | 79.0 | 1.1 | 15.5 | 31.6 | 14.4 |
| Llama-2-70b-chat | Open-Source | General | 26.6 | 52.0 | 8.0 | 14.0 | 72.2 | 0.8 | 12.1 | 28.6 | 25.1 |
| HuatuoGPT-o1-8B | Open-Source | Medical | 25.9 | 51.3 | 11.1 | 18.4 | 75.8 | 1.9 | 4.9 | 32.7 | 11.3 |
| Llama-3.2-3B-Instruct | Open-Source | General | 25.6 | 42.7 | 8.4 | 14.2 | 65.8 | 0.4 | 16.0 | 30.2 | 26.9 |
| Qwen2.5-1.5B-Instruct | Open-Source | General | 25.4 | 41.9 | 7.6 | 12.4 | 70.1 | 0.2 | 16.8 | 31.1 | 23.1 |
| Llama-3.1-8B-UltraMedical | Open-Source | Medical | 24.9 | 39.3 | 2.5 | 11.0 | 76.1 | 0.3 | 11.1 | 34.2 | 24.9 |
| Qwen3-1.7B-Non-Thinking | Open-Source | General | 24.1 | 39.4 | 7.4 | 17.2 | 59.9 | 0.7 | 15.5 | 28.5 | 24.3 |
| Phi-4-mini-reasoning | Open-Source | General | 23.9 | 44.4 | 4.2 | 11.9 | 73.1 | 1.6 | 11.5 | 29.2 | 15.0 |
| MeLLaMA-13B-chat | Open-Source | Medical | 23.4 | 43.5 | 7.4 | 6.6 | 67.9 | 0.6 | 15.6 | 25.3 | 20.1 |
| OpenThinker3-7B | Open-Source | General | 23.0 | 44.4 | 4.2 | 4.4 | 74.1 | 1.3 | 10.0 | 26.2 | 19.8 |
| Qwen3-0.6B-Thinking | Open-Source | General | 22.8 | 44.6 | 7.4 | 4.4 | 65.1 | 0.6 | 15.4 | 24.8 | 19.9 |
| MMed-Llama-3-8B | Open-Source | Medical | 22.7 | 41.9 | 7.4 | 9.8 | 70.6 | 0.8 | 5.3 | 34.0 | 11.5 |
| Llama-2-13b-chat | Open-Source | General | 22.4 | 43.9 | 6.2 | 10.7 | 63.2 | 0.3 | 11.0 | 24.2 | 19.5 |
| BioMistral-7B | Open-Source | Medical | 21.5 | 42.6 | 9.4 | 8.0 | 53.6 | 0.5 | 14.1 | 24.8 | 18.9 |
| MedReason-8B | Open-Source | Medical | 19.5 | 37.5 | 5.8 | 10.4 | 60.3 | 1.4 | 5.0 | 24.9 | 10.7 |
| gemma-3-1b-it | Open-Source | General | 18.7 | 31.3 | 4.7 | 5.6 | 51.1 | 0.0 | 15.7 | 21.6 | 19.9 |
| Qwen3-0.6B-Non-Thinking | Open-Source | General | 18.5 | 32.3 | 3.5 | 3.1 | 51.4 | 0.7 | 16.4 | 22.7 | 17.8 |
| Llama3-OpenBioLLM-8B | Open-Source | Medical | 18.2 | 33.9 | 0.4 | 0.7 | 63.6 | 0.1 | 8.3 | 27.3 | 11.0 |
| DeepSeek-R1-Distill-Qwen-1.5B | Open-Source | General | 18.0 | 31.5 | 2.1 | 1.5 | 54.4 | 0.4 | 13.9 | 24.1 | 16.4 |
| Llama-2-7b-chat | Open-Source | General | 17.9 | 36.0 | 5.8 | 4.9 | 48.8 | 0.4 | 10.1 | 19.6 | 18.0 |
| meditron-70b | Open-Source | Medical | 17.5 | 32.7 | 4.2 | 8.1 | 56.6 | 0.5 | 6.6 | 21.5 | 9.6 |
| Llama-3.2-1B-Instruct | Open-Source | General | 15.8 | 28.9 | 1.9 | 0.1 | 46.3 | 0.2 | 13.1 | 15.3 | 20.8 |
| meditron-7b | Open-Source | Medical | 12.7 | 22.4 | 0.1 | 0.1 | 42.4 | 0.0 | 6.6 | 20.6 | 8.9 |

*Note: The models are ranked by the average score across all categories ("Score Avg") in descending order, and the **bold blue** highlights the highest value for each score column. (The same applies to the following tables for subgroup analysis.)*

*Abbr for Task Type: CLF = Text Classification, EE = Event Extraction, NER = Named Entity Recognition, NLI = Natural Language Inference, NOR = Normalization and Coding, QA = Question Answering, SS = Semantic Similarity, SUM = Summarization.*

## **Table S5.** Subgroup Performance of LLMs Across Different Languages (Zero-shot)

| Model | Model Type | Model Domain | Score Avg | Score DE | Score EN | Score ES | Score FR | Score JA | Score NO | Score PT | Score RU | Score ZH |
|---|---|---|---|---|---|---|---|---|---|---|---|---|
| Qwen3-Next-80B-A3B-Thinking | Open-Source | General | **48.2** | 42.8 | 46.2 | 43.8 | 56.4 | 36.8 | 51.1 | **66.7** | 63.6 | 26.1 |
| gpt-4o | Proprietary | General | 47.9 | 43.9 | 47.2 | 48.8 | 52.2 | **37.4** | **54.1** | 57.7 | 64.7 | 25.4 |
| gemini-2.5-flash | Proprietary | General | 47.9 | **45.3** | 47.0 | 38.8 | **60.6** | 35.3 | 48.8 | 61.6 | 64.9 | **29.1** |
| DeepSeek-R1 | Open-Source | General | 47.6 | 44.6 | 46.4 | 41.4 | 56.2 | 34.7 | 51.2 | 60.2 | **65.7** | 28.1 |
| gemini-1.5-pro | Proprietary | General | 46.9 | 43.5 | **47.3** | 38.7 | 54.6 | 35.3 | 50.9 | 63.3 | 62.5 | 25.7 |
| gemini-2.0-flash | Proprietary | General | 46.6 | 42.6 | 45.8 | 43.6 | 51.4 | 36.0 | 49.6 | 60.7 | 65.0 | 25.3 |
| Qwen3-235B-A22B-Thinking | Open-Source | General | 45.6 | 41.0 | 44.0 | 43.3 | 51.5 | 34.7 | 48.2 | 60.9 | 62.9 | 24.2 |
| Mistral-Large-Instruct-2411 | Open-Source | General | 45.1 | 41.6 | 45.0 | 45.7 | 53.2 | 33.6 | 40.9 | 55.8 | 63.7 | 26.7 |
| Qwen3-30B-A3B-Thinking-2507 | Open-Source | General | 44.6 | 41.3 | 45.8 | 42.7 | 53.3 | 31.7 | 36.4 | 63.6 | 63.4 | 22.9 |
| Qwen3-32B-Thinking | Open-Source | General | 44.5 | 40.3 | 44.3 | 43.6 | 49.9 | 32.7 | 46.9 | 59.6 | 60.6 | 22.8 |
| Athene-V2-Chat | Open-Source | General | 44.3 | 41.6 | 45.5 | 49.8 | 43.0 | 33.6 | 40.1 | 61.2 | 60.3 | 23.8 |
| medgemma-27b-it | Open-Source | Medical | 44.3 | 39.8 | 44.6 | 48.3 | 48.8 | 34.2 | 41.1 | 59.4 | 61.4 | 20.8 |
| Qwen2.5-72B-Instruct | Open-Source | General | 44.1 | 41.7 | 45.5 | **50.0** | 42.1 | 34.5 | 38.2 | 61.3 | 59.5 | 23.8 |
| Qwen3-14B-Thinking | Open-Source | General | 44.0 | 39.1 | 43.7 | 43.6 | 49.8 | 32.9 | 43.8 | 62.9 | 59.5 | 20.5 |
| HuatuoGPT-o1-72B | Open-Source | Medical | 43.8 | 39.4 | 45.3 | 48.6 | 42.5 | 35.0 | 37.9 | 61.4 | 60.8 | 22.8 |
| Qwen3-8B-Thinking | Open-Source | General | 43.6 | 38.7 | 43.4 | 37.9 | 49.3 | 35.6 | 44.2 | 62.4 | 60.7 | 20.6 |
| Qwen3-30B-A3B-Thinking | Open-Source | General | 43.3 | 41.4 | 45.2 | 33.2 | 51.9 | 33.3 | 42.0 | 61.9 | 60.2 | 20.7 |
| Qwen3-235B-A22B-Non-Thinking | Open-Source | General | 43.0 | 37.6 | 42.4 | 48.1 | 40.5 | 32.1 | 41.0 | 62.0 | 61.2 | 21.8 |
| Qwen3-4B-Thinking-2507 | Open-Source | General | 42.9 | 41.3 | 43.5 | 33.1 | 49.4 | 33.4 | 43.1 | 62.6 | 60.1 | 19.8 |
| gemma-3-27b-it | Open-Source | General | 42.9 | 40.2 | 43.3 | 46.9 | 50.0 | 33.0 | 40.2 | 51.9 | 60.4 | 20.6 |
| Qwen3-32B-Non-Thinking | Open-Source | General | 42.8 | 37.5 | 42.8 | 48.8 | 43.9 | 31.6 | 39.7 | 58.8 | 60.2 | 21.5 |
| DeepSeek-R1-Distill-Qwen-32B | Open-Source | General | 42.8 | 39.4 | 43.1 | 40.5 | 48.1 | 31.7 | 41.0 | 57.9 | 62.0 | 21.6 |
| Qwen2.5-32B-Instruct | Open-Source | General | 42.4 | 38.8 | 44.3 | 42.2 | 46.3 | 32.3 | 38.2 | 56.1 | 61.7 | 21.3 |
| Llama-3.3-70B-Instruct | Open-Source | General | 42.4 | 37.6 | 44.6 | 45.6 | 47.2 | 31.0 | 36.4 | 58.7 | 58.8 | 21.8 |
| QWQ-32B | Open-Source | General | 42.2 | 42.5 | 41.3 | 34.6 | 41.5 | 29.4 | 47.0 | 58.3 | 62.2 | 22.7 |
| Qwen3-Next-80B-A3B-Instruct | Open-Source | General | 42.2 | 39.3 | 43.7 | 46.3 | 42.2 | 30.9 | 40.2 | 55.9 | 58.4 | 22.8 |
| DeepSeek-R1-Distill-Llama-70B | Open-Source | General | 41.8 | 38.7 | 44.5 | 33.0 | 45.1 | 29.0 | 43.9 | 59.7 | 60.2 | 22.0 |
| Llama-3.1-70B-Instruct | Open-Source | General | 41.8 | 36.2 | 43.2 | 39.9 | 50.0 | 32.1 | 34.6 | 60.6 | 57.4 | 22.5 |
| Mistral-Small-3.1-24B-Instruct-2503 | Open-Source | General | 41.8 | 37.3 | 44.2 | 38.7 | 42.8 | 33.1 | 42.2 | 53.7 | 60.5 | 23.6 |
| QwenLong-L1-32B | Open-Source | General | 41.6 | 41.5 | 42.3 | 39.2 | 38.3 | 28.7 | 42.7 | 57.2 | 62.3 | 21.8 |
| Qwen3-4B-Thinking | Open-Source | General | 41.2 | 37.2 | 42.5 | 36.1 | 45.5 | 31.4 | 39.2 | 59.4 | 59.0 | 20.6 |
| gemma-2-27b-it | Open-Source | General | 41.0 | 38.4 | 41.0 | 47.2 | 43.7 | 31.0 | 36.4 | 49.8 | 59.2 | 21.9 |
| Qwen3-14B-Non-Thinking | Open-Source | General | 40.8 | 34.4 | 40.5 | 48.5 | 38.5 | 29.5 | 34.5 | 64.0 | 59.6 | 18.1 |
| gemma-3-12b-it | Open-Source | General | 40.0 | 37.4 | 40.9 | 37.8 | 45.6 | 29.3 | 36.6 | 55.1 | 57.8 | 19.2 |
| Mistral-Small-24B-Instruct-2501 | Open-Source | General | 39.6 | 30.8 | 43.1 | 39.0 | 42.9 | 31.7 | 38.2 | 50.8 | 56.9 | 22.9 |
| Qwen3-30B-A3B-Instruct-2507 | Open-Source | General | 39.6 | 35.6 | 40.9 | 45.0 | 37.0 | 26.7 | 35.6 | 58.6 | 57.1 | 19.8 |
| Baichuan-M2-32B | Open-Source | Medical | 39.5 | 41.3 | 41.4 | 27.3 | 32.2 | 27.6 | 46.2 | 59.7 | 59.0 | 21.1 |
| gpt-oss-120b-high | Open-Source | General | 39.2 | 39.1 | 41.2 | 31.2 | 36.0 | 21.2 | 48.8 | 58.0 | 56.8 | 19.9 |
| Magistral-Small-2506 | Open-Source | General | 39.0 | 36.2 | 42.3 | 39.3 | 36.3 | 29.6 | 34.1 | 53.7 | 57.5 | 21.8 |
| DeepSeek-R1-0528-Qwen3-8B | Open-Source | General | 38.2 | 38.6 | 40.6 | 29.9 | 46.3 | 30.1 | 30.3 | 52.3 | 56.0 | 19.9 |
| Phi-4 | Open-Source | General | 37.6 | 35.5 | 40.1 | 36.7 | 38.4 | 24.2 | 31.4 | 53.9 | 56.8 | 21.7 |
| gemma-2-9b-it | Open-Source | General | 37.4 | 35.1 | 39.2 | 43.5 | 38.8 | 24.8 | 35.0 | 47.5 | 56.2 | 16.7 |
| Mistral-Small-Instruct-2409 | Open-Source | General | 37.3 | 34.1 | 38.2 | 43.0 | 33.6 | 30.6 | 27.7 | 55.8 | 51.1 | 22.0 |
| Baichuan-M1-14B-Instruct | Open-Source | Medical | 37.3 | 36.0 | 40.9 | 39.4 | 32.3 | 27.5 | 29.0 | 54.8 | 56.2 | 19.3 |
| Qwen3-4B-Instruct-2507 | Open-Source | General | 37.0 | 30.7 | 39.4 | 41.1 | 28.7 | 31.7 | 30.2 | 61.5 | 54.9 | 14.5 |
| gpt-35-turbo | Proprietary | General | 36.8 | 33.6 | 39.1 | 43.8 | 34.2 | 28.4 | 30.9 | 45.5 | 53.5 | 22.2 |
| Qwen3-30B-A3B-Non-Thinking | Open-Source | General | 36.8 | 36.4 | 41.9 | 38.1 | 42.5 | 26.3 | 21.0 | 52.9 | 55.1 | 17.1 |
| Llama-4-Scout-17B-16E-Instruct | Open-Source | General | 36.8 | 34.2 | 39.6 | 41.4 | 35.6 | 28.1 | 21.4 | 56.8 | 55.7 | 18.4 |
| HuatuoGPT-o1-70B | Open-Source | Medical | 36.4 | 30.2 | 37.7 | 28.1 | 38.7 | 24.3 | 37.8 | 52.8 | 57.5 | 20.2 |
| Qwen3-8B-Non-Thinking | Open-Source | General | 36.4 | 32.3 | 38.9 | 40.9 | 37.3 | 27.0 | 19.4 | 62.1 | 56.4 | 13.5 |
| Llama-3.1-Nemotron-70B-Instruct-HF | Open-Source | General | 35.9 | 31.0 | 35.4 | 37.9 | 34.7 | 24.4 | 29.2 | 56.3 | 56.0 | 18.5 |
| DeepSeek-R1-Distill-Qwen-14B | Open-Source | General | 35.8 | 33.9 | 38.5 | 37.7 | 34.3 | 23.3 | 26.2 | 55.3 | 53.9 | 19.0 |
| Qwen3-4B-Non-Thinking | Open-Source | General | 35.5 | 31.8 | 39.1 | 38.4 | 29.3 | 30.5 | 27.5 | 59.8 | 51.4 | 11.9 |
| Llama-3-70B-UltraMedical | Open-Source | Medical | 34.8 | 32.4 | 37.5 | 29.1 | 30.6 | 28.0 | 29.6 | 54.1 | 53.5 | 18.6 |
| QwQ-32B-Preview | Open-Source | General | 34.4 | 29.7 | 36.1 | 38.1 | 25.9 | 23.4 | 30.5 | 57.2 | 54.3 | 14.8 |
| MeLLaMA-70B-chat | Open-Source | Medical | 33.8 | 32.3 | 35.8 | 35.2 | 24.8 | 26.5 | 26.3 | 52.8 | 51.8 | 18.3 |
| Llama3-OpenBioLLM-70B | Open-Source | Medical | 33.8 | 31.7 | 38.4 | 35.6 | 26.5 | 23.6 | 22.7 | 55.8 | 52.7 | 17.5 |
| Qwen2.5-7B-Instruct | Open-Source | General | 33.7 | 31.3 | 34.8 | 30.0 | 31.3 | 22.7 | 27.6 | 56.3 | 55.0 | 14.8 |
| HuatuoGPT-o1-7B | Open-Source | Medical | 32.7 | 26.5 | 33.4 | 30.4 | 32.0 | 23.7 | 26.6 | 54.6 | 54.3 | 13.1 |
| Yi-1.5-34B-Chat-16K | Open-Source | General | 32.5 | 34.2 | 38.1 | 41.1 | 23.7 | 20.8 | 21.2 | 53.8 | 47.1 | 12.8 |
| Phi-3.5-MoE-instruct | Open-Source | General | 31.9 | 27.2 | 34.0 | 34.6 | 27.4 | 21.5 | 25.1 | 57.4 | 45.8 | 13.8 |
| Ministral-8B-Instruct-2410 | Open-Source | General | 31.0 | 27.6 | 36.3 | 38.7 | 26.8 | 21.7 | 22.0 | 40.6 | 50.6 | 15.0 |
| medgemma-4b-it | Open-Source | Medical | 30.6 | 28.0 | 35.5 | 33.0 | 20.3 | 21.5 | 22.9 | 54.5 | 48.3 | 11.0 |
| Llama-3.1-8B-Instruct | Open-Source | General | 29.9 | 26.9 | 34.4 | 33.8 | 21.2 | 18.5 | 21.5 | 51.4 | 47.5 | 14.3 |
| DeepSeek-R1-Distill-Llama-8B | Open-Source | General | 29.2 | 25.7 | 33.6 | 18.0 | 23.1 | 19.5 | 27.1 | 53.1 | 47.3 | 15.4 |
| gemma-3-4b-it | Open-Source | General | 28.7 | 27.9 | 33.8 | 29.4 | 23.4 | 21.5 | 25.4 | 33.8 | 49.4 | 13.8 |
| Yi-1.5-9B-Chat-16K | Open-Source | General | 28.6 | 31.8 | 34.3 | 30.3 | 23.6 | 19.6 | 23.6 | 37.7 | 46.1 | 10.3 |
| gpt-oss-20b-high | Open-Source | General | 28.6 | 30.7 | 34.5 | 21.4 | 8.5 | 11.7 | 27.0 | 55.2 | 52.9 | 15.4 |
| Qwen2.5-3B-Instruct | Open-Source | General | 26.7 | 25.9 | 31.7 | 18.0 | 18.0 | 20.6 | 21.4 | 44.8 | 47.5 | 12.1 |
| Llama-2-70b-chat | Open-Source | General | 26.1 | 21.2 | 30.8 | 30.2 | 14.8 | 18.1 | 23.4 | 40.2 | 44.8 | 11.4 |
| HuatuoGPT-o1-8B | Open-Source | Medical | 26.1 | 21.4 | 31.6 | 18.6 | 18.1 | 13.5 | 22.9 | 44.9 | 48.8 | 14.8 |
| Phi-3.5-mini-instruct | Open-Source | General | 25.4 | 23.2 | 31.0 | 9.3 | 18.1 | 19.5 | 18.4 | 50.5 | 46.6 | 11.6 |
| Qwen3-1.7B-Thinking | Open-Source | General | 25.3 | 27.7 | 31.7 | 18.2 | 19.9 | 21.6 | 21.9 | 25.7 | 50.7 | 10.5 |
| Phi-4-mini-instruct | Open-Source | General | 25.1 | 21.6 | 30.0 | 28.1 | 11.7 | 20.7 | 16.8 | 43.1 | 43.5 | 10.7 |
| DeepSeek-R1-Distill-Qwen-7B | Open-Source | General | 24.3 | 25.7 | 30.6 | 22.3 | 13.9 | 15.2 | 17.0 | 39.8 | 41.6 | 12.8 |
| Llama-3.2-3B-Instruct | Open-Source | General | 23.1 | 16.5 | 30.5 | 22.3 | 13.6 | 14.2 | 13.9 | 47.0 | 41.1 | 8.7 |
| Qwen2.5-1.5B-Instruct | Open-Source | General | 22.9 | 23.1 | 26.1 | 19.8 | 9.1 | 13.6 | 13.8 | 48.8 | 43.3 | 8.5 |
| Phi-4-mini-reasoning | Open-Source | General | 22.1 | 12.7 | 27.7 | 15.8 | 16.4 | 8.6 | 17.6 | 45.3 | 43.1 | 11.6 |
| Llama-2-13b-chat | Open-Source | General | 20.8 | 13.4 | 28.1 | 21.9 | 10.4 | 15.2 | 16.7 | 33.5 | 38.7 | 9.4 |
| Qwen3-1.7B-Non-Thinking | Open-Source | General | 20.7 | 22.6 | 27.9 | 15.4 | 16.8 | 18.1 | 19.9 | 17.5 | 41.3 | 6.5 |
| Llama-3.1-8B-UltraMedical | Open-Source | Medical | 20.3 | 14.1 | 26.9 | 18.1 | 7.8 | 8.9 | 7.5 | 44.4 | 47.4 | 8.0 |
| MMed-Llama-3-8B | Open-Source | Medical | 19.5 | 10.6 | 30.3 | 14.4 | 9.8 | 11.2 | 14.2 | 37.6 | 41.8 | 5.8 |
| MeLLaMA-13B-chat | Open-Source | Medical | 19.5 | 16.4 | 28.5 | 18.0 | 4.0 | 16.4 | 9.9 | 37.2 | 37.3 | 7.5 |
| BioMistral-7B | Open-Source | Medical | 18.7 | 17.5 | 26.9 | 15.6 | 6.6 | 13.3 | 10.7 | 37.0 | 30.2 | 10.8 |
| Qwen3-0.6B-Thinking | Open-Source | General | 18.1 | 23.0 | 26.2 | 12.7 | 3.2 | 10.4 | 4.4 | 38.3 | 36.6 | 7.8 |
| MedReason-8B | Open-Source | Medical | 17.7 | 13.2 | 23.8 | 19.0 | 8.8 | 8.3 | 7.6 | 32.3 | 36.2 | 9.7 |
| OpenThinker3-7B | Open-Source | General | 16.9 | 12.7 | 28.8 | 14.5 | 3.6 | 11.1 | 4.1 | 28.6 | 38.1 | 10.5 |
| meditron-70b | Open-Source | Medical | 15.5 | 10.5 | 20.7 | 6.8 | 6.2 | 10.3 | 13.7 | 30.6 | 32.4 | 7.9 |
| Llama-2-7b-chat | Open-Source | General | 15.4 | 11.7 | 23.9 | 15.0 | 4.1 | 7.3 | 9.8 | 34.3 | 27.7 | 5.0 |
| Qwen3-0.6B-Non-Thinking | Open-Source | General | 14.0 | 15.8 | 19.7 | 7.5 | 1.0 | 9.6 | 1.8 | 35.9 | 29.6 | 5.2 |
| gemma-3-1b-it | Open-Source | General | 13.9 | 16.7 | 20.0 | 8.1 | 5.2 | 14.5 | 6.5 | 18.3 | 29.3 | 6.5 |
| Llama3-OpenBioLLM-8B | Open-Source | Medical | 13.3 | 8.4 | 19.7 | 7.8 | 1.2 | 8.1 | 0.0 | 31.2 | 36.0 | 7.5 |
| DeepSeek-R1-Distill-Qwen-1.5B | Open-Source | General | 13.3 | 13.7 | 18.6 | 10.3 | 0.9 | 7.3 | 2.9 | 32.5 | 27.5 | 5.9 |
| Llama-3.2-1B-Instruct | Open-Source | General | 10.9 | 11.4 | 18.1 | 10.5 | 0.2 | 2.9 | 0.0 | 25.5 | 25.0 | 4.8 |
| meditron-7b | Open-Source | Medical | 9.2 | 4.7 | 13.5 | 5.9 | 0.0 | 4.1 | 0.0 | 27.9 | 21.3 | 5.2 |

*Abbr for Language: DE = German, EN = English, ES = Spanish, FR = French, JA = Japanese, PT = Portuguese, NO = Norwegian, RU = Russian, and ZH = Chinese.*

## **Table S6.** Subgroup Performance of LLMs Across Different Clinical Specialties (Zero-shot)

| Model | Model Type | Model Domain | Score Avg | Score CAR | Score END | Score GAS | Score GEN | Score CC | Score NEU | Score ONC | Score OTH | Score PED | Score PHA | Score PUL | Score RAD |
|---|---|---|---|---|---|---|---|---|---|---|---|---|---|---|---|
| gemini-2.5-flash | Proprietary | General | **48.9** | 48.7 | 52.0 | **65.8** | **44.5** | 43.5 | **76.0** | 22.3 | 34.7 | 47.8 | **39.7** | **62.8** | 48.8 |
| DeepSeek-R1 | Open-Source | General | 48.5 | 49.3 | 53.0 | 64.9 | 44.1 | 41.7 | 72.5 | 21.9 | **35.7** | 49.2 | 38.3 | 61.4 | **49.8** |
| Qwen3-Next-80B-A3B-Thinking | Open-Source | General | 48.2 | 49.9 | 52.1 | 64.2 | 43.1 | 43.6 | 74.2 | 17.6 | 35.6 | 49.7 | 38.1 | 60.9 | 49.2 |
| gemini-1.5-pro | Proprietary | General | 47.9 | 47.9 | 52.2 | 62.3 | 42.8 | **46.6** | 71.8 | **22.5** | 33.1 | **50.4** | 38.5 | 58.7 | 47.6 |
| gpt-4o | Proprietary | General | 47.5 | **50.4** | 52.9 | 62.5 | 44.4 | 46.0 | 71.3 | 19.1 | 32.0 | 47.3 | 37.7 | 58.1 | 48.1 |
| Qwen3-30B-A3B-Thinking-2507 | Open-Source | General | 46.4 | 48.2 | **54.6** | 64.1 | 41.3 | 41.4 | 71.9 | 16.6 | 31.2 | 42.1 | 36.1 | 60.5 | 49.0 |
| Mistral-Large-Instruct-2411 | Open-Source | General | 45.8 | 43.7 | 50.5 | 61.2 | 42.4 | 42.6 | 69.8 | 22.3 | 26.4 | 48.8 | 35.7 | 58.4 | 47.4 |
| gemini-2.0-flash | Proprietary | General | 45.7 | 46.2 | 50.7 | 63.7 | 43.6 | 45.8 | 71.9 | 20.3 | 22.7 | 38.9 | 36.9 | 60.8 | 47.0 |
| Qwen3-235B-A22B-Thinking | Open-Source | General | 45.5 | 45.7 | 49.0 | 63.2 | 42.2 | 39.9 | 72.7 | 12.7 | 30.6 | 44.9 | 36.1 | 59.8 | 49.5 |
| Qwen3-30B-A3B-Thinking | Open-Source | General | 45.0 | 46.9 | 50.6 | 61.8 | 40.6 | 41.2 | 71.0 | 18.2 | 27.3 | 44.5 | 36.8 | 56.4 | 44.7 |
| Qwen2.5-72B-Instruct | Open-Source | General | 44.8 | 43.3 | 49.2 | 60.9 | 43.2 | 44.7 | 68.9 | 16.6 | 26.3 | 45.3 | 35.9 | 57.5 | 45.9 |
| Athene-V2-Chat | Open-Source | General | 44.8 | 44.5 | 49.8 | 61.0 | 43.2 | 44.2 | 68.9 | 15.7 | 26.4 | 44.6 | 35.7 | 57.6 | 46.4 |
| medgemma-27b-it | Open-Source | Medical | 44.7 | 41.2 | 47.8 | 59.9 | 40.2 | 44.2 | 71.1 | 19.7 | 28.2 | 46.1 | 36.4 | 56.5 | 44.5 |
| Qwen3-32B-Thinking | Open-Source | General | 44.6 | 46.2 | 49.8 | 61.3 | 41.9 | 39.5 | 70.2 | 13.8 | 27.1 | 44.7 | 35.0 | 58.2 | 47.6 |
| Qwen3-4B-Thinking-2507 | Open-Source | General | 44.3 | 46.7 | 49.4 | 61.6 | 39.1 | 40.1 | 69.1 | 10.9 | 29.4 | 46.0 | 35.8 | 56.5 | 47.1 |
| HuatuoGPT-o1-72B | Open-Source | Medical | 44.2 | 43.3 | 48.7 | 60.5 | 42.5 | 44.5 | 69.1 | 17.2 | 25.8 | 41.0 | 36.2 | 57.0 | 44.5 |
| Qwen3-8B-Thinking | Open-Source | General | 44.2 | 46.3 | 49.8 | 61.3 | 39.7 | 39.0 | 70.6 | 12.1 | 28.2 | 43.2 | 35.6 | 57.2 | 47.1 |
| Qwen3-14B-Thinking | Open-Source | General | 44.2 | 44.7 | 49.0 | 61.2 | 40.1 | 40.1 | 70.4 | 13.2 | 27.7 | 43.6 | 35.7 | 56.7 | 47.5 |
| gemma-3-27b-it | Open-Source | General | 43.9 | 41.3 | 48.2 | 58.7 | 38.9 | 41.5 | 67.8 | 21.5 | 28.5 | 46.3 | 34.8 | 54.8 | 44.1 |
| Qwen3-Next-80B-A3B-Instruct | Open-Source | General | 43.7 | 42.8 | 48.7 | 61.8 | 40.7 | 40.1 | 70.4 | 15.3 | 23.6 | 43.1 | 33.1 | 57.5 | 47.6 |
| DeepSeek-R1-Distill-Llama-70B | Open-Source | General | 43.7 | 43.9 | 49.5 | 60.4 | 39.7 | 41.2 | 69.1 | 15.3 | 25.0 | 45.1 | 32.9 | 55.4 | 47.3 |
| Baichuan-M2-32B | Open-Source | Medical | 43.4 | 44.4 | 47.8 | 63.8 | 38.3 | 37.8 | 73.2 | 11.4 | 28.6 | 44.0 | 30.0 | 54.3 | 47.4 |
| Llama-3.3-70B-Instruct | Open-Source | General | 43.4 | 41.6 | 48.4 | 58.3 | 40.1 | 42.7 | 67.6 | 20.6 | 25.8 | 42.1 | 32.0 | 55.8 | 46.0 |
| QwenLong-L1-32B | Open-Source | General | 43.4 | 40.9 | 47.4 | 61.7 | 40.8 | 37.8 | 70.6 | 12.6 | 26.4 | 44.9 | 31.7 | 58.3 | 47.3 |
| DeepSeek-R1-Distill-Qwen-32B | Open-Source | General | 43.3 | 43.0 | 48.6 | 59.4 | 40.6 | 39.3 | 69.8 | 14.3 | 24.6 | 44.8 | 32.5 | 56.3 | 46.5 |
| QWQ-32B | Open-Source | General | 43.3 | 42.8 | 47.8 | 62.3 | 40.3 | 38.9 | 71.3 | 13.3 | 22.4 | 43.4 | 30.6 | 58.6 | 47.4 |
| Qwen2.5-32B-Instruct | Open-Source | General | 42.8 | 42.7 | 48.0 | 57.8 | 39.9 | 42.7 | 67.7 | 16.4 | 20.2 | 44.7 | 34.7 | 54.1 | 44.3 |
| Mistral-Small-3.1-24B-Instruct-2503 | Open-Source | General | 42.2 | 43.0 | 48.1 | 57.5 | 39.9 | 44.1 | 61.4 | 21.5 | 20.2 | 39.4 | 34.7 | 51.4 | 44.7 |
| Qwen3-4B-Thinking | Open-Source | General | 42.1 | 43.8 | 48.0 | 58.7 | 38.8 | 38.1 | 66.4 | 10.3 | 25.1 | 41.0 | 34.2 | 54.0 | 46.6 |
| Qwen3-235B-A22B-Non-Thinking | Open-Source | General | 42.0 | 38.1 | 43.7 | 57.9 | 40.2 | 42.7 | 66.0 | 14.5 | 24.7 | 42.5 | 34.1 | 54.5 | 45.3 |
| Llama-3.1-70B-Instruct | Open-Source | General | 41.8 | 39.7 | 47.1 | 59.2 | 40.5 | 42.3 | 66.7 | 20.2 | 19.1 | 34.0 | 30.0 | 55.3 | 46.9 |
| Qwen3-32B-Non-Thinking | Open-Source | General | 41.8 | 39.3 | 44.3 | 57.6 | 40.8 | 42.4 | 66.4 | 15.8 | 21.9 | 41.4 | 33.8 | 53.4 | 44.6 |
| gpt-oss-120b-high | Open-Source | General | 41.5 | 42.6 | 44.8 | 62.5 | 38.2 | 40.4 | 68.6 | 9.9 | 20.8 | 36.0 | 30.2 | 55.3 | 48.3 |
| gemma-2-27b-it | Open-Source | General | 41.4 | 37.3 | 43.8 | 56.4 | 38.4 | 39.9 | 65.0 | 16.8 | 24.1 | 43.5 | 34.0 | 52.8 | 45.0 |
| DeepSeek-R1-0528-Qwen3-8B | Open-Source | General | 41.0 | 40.5 | 44.0 | 60.0 | 38.1 | 34.4 | 65.7 | 11.8 | 25.6 | 39.6 | 31.7 | 54.4 | 46.7 |
| gemma-3-12b-it | Open-Source | General | 40.7 | 38.3 | 43.2 | 56.8 | 37.7 | 37.6 | 66.9 | 15.6 | 22.7 | 40.8 | 32.4 | 52.7 | 44.2 |
| Mistral-Small-24B-Instruct-2501 | Open-Source | General | 40.4 | 41.4 | 47.2 | 52.8 | 36.1 | 41.3 | 62.4 | 19.2 | 21.0 | 37.2 | 32.8 | 48.6 | 44.5 |
| Qwen3-30B-A3B-Instruct-2507 | Open-Source | General | 39.9 | 39.0 | 44.3 | 56.1 | 37.8 | 37.8 | 64.8 | 13.6 | 18.1 | 38.1 | 31.7 | 53.4 | 43.8 |
| Baichuan-M1-14B-Instruct | Open-Source | Medical | 39.8 | 35.2 | 42.5 | 58.1 | 36.1 | 42.0 | 65.0 | 14.0 | 19.0 | 39.1 | 32.5 | 52.8 | 41.3 |
| Magistral-Small-2506 | Open-Source | General | 39.7 | 37.8 | 45.7 | 54.0 | 38.1 | 40.4 | 58.4 | 16.5 | 20.1 | 39.6 | 31.9 | 50.5 | 43.4 |
| Qwen3-14B-Non-Thinking | Open-Source | General | 39.6 | 35.3 | 40.4 | 55.5 | 37.9 | 41.8 | 65.8 | 14.7 | 21.1 | 35.2 | 33.8 | 51.1 | 42.4 |
| Qwen3-30B-A3B-Non-Thinking | Open-Source | General | 39.3 | 33.6 | 42.6 | 54.2 | 37.1 | 41.8 | 59.3 | 19.0 | 23.9 | 39.7 | 31.6 | 48.2 | 40.9 |
| Phi-4 | Open-Source | General | 39.2 | 37.1 | 42.1 | 55.0 | 36.6 | 40.8 | 62.6 | 15.1 | 18.0 | 38.4 | 28.2 | 52.2 | 44.8 |
| gpt-35-turbo | Proprietary | General | 38.6 | 37.2 | 41.5 | 53.3 | 36.0 | 33.4 | 62.6 | 17.7 | 20.5 | 38.8 | 31.2 | 46.9 | 44.1 |
| HuatuoGPT-o1-70B | Open-Source | Medical | 38.5 | 40.8 | 44.8 | 54.9 | 32.4 | 36.3 | 64.6 | 13.8 | 17.2 | 34.2 | 26.3 | 52.3 | 44.5 |
| gemma-2-9b-it | Open-Source | General | 38.4 | 35.4 | 40.8 | 54.3 | 35.5 | 34.7 | 64.2 | 13.4 | 20.4 | 41.6 | 31.4 | 49.4 | 40.2 |
| Llama-4-Scout-17B-16E-Instruct | Open-Source | General | 38.2 | 32.6 | 42.2 | 55.3 | 36.4 | 39.7 | 64.0 | 14.6 | 18.8 | 33.7 | 28.1 | 50.5 | 41.9 |
| DeepSeek-R1-Distill-Qwen-14B | Open-Source | General | 38.1 | 31.1 | 38.8 | 57.6 | 37.0 | 35.7 | 64.4 | 6.3 | 21.5 | 39.5 | 28.7 | 50.5 | 45.7 |
| Mistral-Small-Instruct-2409 | Open-Source | General | 37.6 | 32.0 | 38.1 | 54.1 | 37.1 | 36.8 | 60.0 | 15.2 | 16.3 | 35.6 | 30.1 | 50.3 | 46.0 |
| Qwen3-4B-Instruct-2507 | Open-Source | General | 36.8 | 35.8 | 41.2 | 53.4 | 36.0 | 36.6 | 63.9 | 7.6 | 14.8 | 33.5 | 31.3 | 50.0 | 37.9 |
| Qwen3-8B-Non-Thinking | Open-Source | General | 36.8 | 28.9 | 36.8 | 55.7 | 35.8 | 42.0 | 63.6 | 12.5 | 17.5 | 32.9 | 29.9 | 50.4 | 36.0 |
| Llama3-OpenBioLLM-70B | Open-Source | Medical | 36.1 | 30.6 | 37.0 | 55.4 | 35.0 | 38.0 | 63.0 | 16.0 | 12.5 | 30.3 | 28.7 | 46.2 | 41.1 |
| Llama-3-70B-UltraMedical | Open-Source | Medical | 35.8 | 29.2 | 34.2 | 51.7 | 35.1 | 37.2 | 61.4 | 15.1 | 12.3 | 33.0 | 30.0 | 49.0 | 40.9 |
| Llama-3.1-Nemotron-70B-Instruct-HF | Open-Source | General | 35.8 | 30.2 | 34.2 | 57.2 | 35.7 | 40.4 | 63.4 | 10.7 | 12.6 | 28.4 | 23.0 | 49.5 | 44.8 |
| Qwen3-4B-Non-Thinking | Open-Source | General | 35.2 | 32.9 | 39.9 | 50.1 | 35.6 | 37.8 | 57.2 | 11.1 | 16.1 | 33.7 | 30.9 | 44.3 | 32.8 |
| QwQ-32B-Preview | Open-Source | General | 34.9 | 29.7 | 33.3 | 52.2 | 33.1 | 38.4 | 62.4 | 9.0 | 15.2 | 35.6 | 25.6 | 46.6 | 37.7 |
| MeLLaMA-70B-chat | Open-Source | Medical | 34.6 | 29.2 | 35.5 | 49.7 | 33.7 | 36.0 | 58.5 | 11.5 | 12.0 | 33.7 | 30.8 | 42.0 | 43.1 |
| Yi-1.5-34B-Chat-16K | Open-Source | General | 34.6 | 24.0 | 32.8 | 51.9 | 35.1 | 41.1 | 59.7 | 11.7 | 15.3 | 35.5 | 26.9 | 44.8 | 36.5 |
| Qwen2.5-7B-Instruct | Open-Source | General | 34.0 | 29.3 | 34.5 | 51.0 | 31.9 | 34.9 | 56.7 | 9.7 | 14.3 | 34.9 | 28.1 | 45.0 | 37.9 |
| Ministral-8B-Instruct-2410 | Open-Source | General | 33.7 | 29.0 | 35.5 | 52.3 | 30.8 | 33.0 | 59.6 | 10.6 | 12.1 | 30.7 | 29.1 | 44.6 | 36.7 |
| HuatuoGPT-o1-7B | Open-Source | Medical | 32.4 | 27.3 | 32.4 | 50.0 | 31.2 | 37.8 | 55.7 | 9.8 | 17.1 | 27.5 | 23.4 | 41.6 | 34.8 |
| medgemma-4b-it | Open-Source | Medical | 31.8 | 25.4 | 31.3 | 48.0 | 31.3 | 38.7 | 56.2 | 7.7 | 12.7 | 32.0 | 26.7 | 37.9 | 33.8 |
| Llama-3.1-8B-Instruct | Open-Source | General | 31.6 | 25.6 | 31.9 | 51.3 | 31.5 | 33.2 | 60.3 | 9.9 | 8.2 | 26.4 | 22.9 | 45.0 | 33.2 |
| Yi-1.5-9B-Chat-16K | Open-Source | General | 31.5 | 25.9 | 32.7 | 45.9 | 30.0 | 34.9 | 56.9 | 11.3 | 11.0 | 34.4 | 25.8 | 35.2 | 33.7 |
| gemma-3-4b-it | Open-Source | General | 31.5 | 27.4 | 30.6 | 47.9 | 29.5 | 30.8 | 57.7 | 11.4 | 9.7 | 30.1 | 27.2 | 42.3 | 33.1 |
| gpt-oss-20b-high | Open-Source | General | 30.8 | 27.0 | 33.1 | 47.7 | 31.1 | 33.8 | 48.8 | 2.4 | 9.5 | 28.0 | 25.0 | 40.1 | 43.3 |
| DeepSeek-R1-Distill-Llama-8B | Open-Source | General | 30.7 | 26.7 | 31.8 | 46.5 | 30.4 | 31.3 | 55.1 | 7.5 | 9.8 | 29.6 | 22.3 | 39.7 | 37.5 |
| Phi-3.5-MoE-instruct | Open-Source | General | 30.1 | 23.9 | 27.0 | 42.0 | 32.9 | 38.9 | 45.2 | 10.7 | 15.9 | 26.4 | 24.1 | 39.6 | 34.2 |
| HuatuoGPT-o1-8B | Open-Source | Medical | 29.2 | 28.1 | 34.8 | 45.5 | 26.8 | 27.0 | 54.5 | 6.7 | 8.7 | 22.3 | 19.9 | 39.5 | 36.6 |
| Qwen3-1.7B-Thinking | Open-Source | General | 29.0 | 24.9 | 32.0 | 44.9 | 26.0 | 32.5 | 49.9 | 5.4 | 11.6 | 31.2 | 21.6 | 36.2 | 32.2 |
| Qwen2.5-3B-Instruct | Open-Source | General | 28.5 | 24.0 | 31.2 | 40.8 | 27.2 | 30.2 | 50.4 | 7.1 | 8.8 | 30.2 | 22.4 | 37.1 | 32.0 |
| Llama-2-70b-chat | Open-Source | General | 26.9 | 20.6 | 24.8 | 42.4 | 28.6 | 28.2 | 54.5 | 6.1 | 9.6 | 17.6 | 21.7 | 35.8 | 33.4 |
| Phi-4-mini-instruct | Open-Source | General | 26.7 | 17.5 | 24.2 | 40.5 | 27.2 | 29.6 | 54.0 | 6.1 | 7.6 | 24.7 | 22.8 | 35.8 | 29.8 |
| DeepSeek-R1-Distill-Qwen-7B | Open-Source | General | 25.9 | 21.6 | 28.0 | 38.7 | 27.9 | 28.3 | 36.5 | 5.3 | 9.5 | 26.7 | 21.8 | 33.9 | 32.8 |
| Phi-3.5-mini-instruct | Open-Source | General | 25.1 | 17.5 | 21.4 | 35.0 | 28.8 | 35.4 | 43.5 | 6.5 | 9.1 | 20.5 | 22.1 | 30.7 | 31.0 |
| Llama-3.2-3B-Instruct | Open-Source | General | 25.1 | 17.9 | 26.3 | 37.2 | 23.1 | 34.6 | 48.2 | 6.5 | 8.1 | 22.0 | 20.7 | 29.0 | 27.5 |
| MMed-Llama-3-8B | Open-Source | Medical | 24.5 | 18.3 | 23.7 | 42.0 | 20.5 | 32.9 | 54.7 | 4.3 | 7.2 | 13.5 | 20.0 | 30.9 | 26.0 |
| Phi-4-mini-reasoning | Open-Source | General | 23.8 | 17.4 | 24.1 | 37.4 | 22.8 | 27.5 | 49.5 | 4.5 | 11.7 | 16.6 | 12.6 | 31.8 | 29.4 |
| Qwen3-1.7B-Non-Thinking | Open-Source | General | 23.6 | 18.2 | 24.7 | 31.8 | 22.4 | 28.3 | 40.9 | 4.2 | 12.5 | 23.3 | 22.7 | 28.3 | 26.1 |
| Qwen2.5-1.5B-Instruct | Open-Source | General | 23.3 | 16.3 | 20.9 | 34.8 | 25.8 | 26.8 | 43.9 | 2.3 | 5.8 | 24.4 | 20.9 | 29.0 | 28.3 |
| OpenThinker3-7B | Open-Source | General | 22.7 | 13.6 | 21.9 | 41.2 | 22.3 | 23.6 | 52.7 | 1.6 | 3.8 | 11.2 | 17.7 | 33.4 | 29.3 |
| Llama-3.1-8B-UltraMedical | Open-Source | Medical | 22.2 | 14.6 | 20.5 | 38.3 | 23.2 | 26.1 | 50.3 | 2.9 | 3.7 | 22.1 | 11.8 | 30.6 | 22.2 |
| MeLLaMA-13B-chat | Open-Source | Medical | 22.2 | 14.6 | 23.8 | 36.1 | 22.7 | 27.8 | 48.8 | 0.9 | 2.6 | 19.7 | 19.2 | 26.2 | 24.4 |
| Llama-2-13b-chat | Open-Source | General | 21.4 | 14.5 | 19.9 | 30.8 | 23.6 | 27.1 | 46.2 | 2.9 | 8.9 | 13.9 | 17.1 | 23.5 | 28.5 |
| MedReason-8B | Open-Source | Medical | 21.1 | 19.4 | 27.7 | 37.6 | 19.6 | 21.1 | 45.7 | 2.8 | 2.4 | 10.8 | 9.4 | 30.3 | 26.5 |
| Qwen3-0.6B-Thinking | Open-Source | General | 20.8 | 14.2 | 22.2 | 34.2 | 23.2 | 28.2 | 31.5 | 3.0 | 3.0 | 21.7 | 19.7 | 23.5 | 25.6 |
| BioMistral-7B | Open-Source | Medical | 19.6 | 12.0 | 19.6 | 25.5 | 23.6 | 23.8 | 39.4 | 2.5 | 4.9 | 16.6 | 19.7 | 21.6 | 26.5 |
| meditron-70b | Open-Source | Medical | 17.3 | 11.7 | 13.6 | 22.6 | 16.5 | 27.0 | 37.9 | 2.4 | 3.1 | 11.2 | 14.7 | 22.8 | 24.5 |
| Llama-2-7b-chat | Open-Source | General | 16.3 | 9.8 | 11.9 | 19.8 | 21.0 | 23.7 | 37.5 | 1.3 | 6.0 | 11.2 | 15.6 | 17.3 | 20.5 |
| Qwen3-0.6B-Non-Thinking | Open-Source | General | 14.6 | 7.1 | 10.6 | 20.8 | 19.7 | 23.9 | 28.8 | 1.0 | 1.3 | 13.8 | 9.2 | 16.3 | 22.8 |
| Llama3-OpenBioLLM-8B | Open-Source | Medical | 14.4 | 7.2 | 14.3 | 28.0 | 17.1 | 25.2 | 24.7 | 1.3 | 0.2 | 7.1 | 5.8 | 20.9 | 21.3 |
| gemma-3-1b-it | Open-Source | General | 14.3 | 7.5 | 9.9 | 24.4 | 18.9 | 22.3 | 17.6 | 0.8 | 4.6 | 15.7 | 17.9 | 18.7 | 13.4 |
| Llama-3.2-1B-Instruct | Open-Source | General | 14.3 | 7.2 | 11.8 | 26.1 | 14.7 | 18.6 | 31.2 | 0.0 | 0.6 | 8.9 | 11.1 | 18.7 | 23.4 |
| DeepSeek-R1-Distill-Qwen-1.5B | Open-Source | General | 12.4 | 6.6 | 11.9 | 21.2 | 19.0 | 18.4 | 16.1 | 0.4 | 1.3 | 10.9 | 12.5 | 16.3 | 14.5 |
| meditron-7b | Open-Source | Medical | 8.1 | 1.6 | 5.1 | 8.5 | 12.0 | 18.4 | 17.9 | 0.0 | 0.2 | 4.4 | 4.9 | 9.6 | 15.1 |

*Abbr for Clinical Specialty: CAR = Cardiology, END = Endocrinology, GAS = Gastroenterology, GEN = General, CC = Critical Care, NEU = Neurology, ONC = Oncology, PED = Pediatrics, PHA = Pharmacology, PUL = Pulmonology, RAD = Radiology, and OTH = Other (used when the specialty appears in only one task).*

**Table S7.** Subgroup Performance of LLMs Across Different Clinical Stages (Zero-shot)

| Model | Model Type | Model Domain | Score Avg | Score D&A | Score D&P | Score IA | Score Research | Score T&I | Score T&R |
|---|---|---|---|---|---|---|---|---|---|
| gemini-2.5-flash | Proprietary | General | **46.2** | 25.0 | 74.8 | 35.2 | **56.5** | **39.6** | 46.4 |
| gpt-4o | Proprietary | General | 45.9 | **25.3** | 77.4 | 36.0 | 54.8 | 35.9 | 46.2 |
| DeepSeek-R1 | Open-Source | General | 45.8 | 25.2 | 75.5 | 36.0 | 55.3 | 36.7 | 46.4 |
| Qwen3-Next-80B-A3B-Thinking | Open-Source | General | 45.7 | 22.8 | 80.8 | 34.9 | 53.6 | 36.5 | 45.7 |
| gemini-1.5-pro | Proprietary | General | 45.7 | 23.7 | 79.3 | **37.1** | 50.7 | 37.3 | 46.0 |
| gemini-2.0-flash | Proprietary | General | 45.0 | 24.3 | 78.2 | 31.1 | 53.3 | 35.3 | **47.5** |
| Mistral-Large-Instruct-2411 | Open-Source | General | 44.0 | 23.8 | 74.3 | 31.9 | 53.0 | 34.6 | 46.6 |
| Qwen3-30B-A3B-Thinking-2507 | Open-Source | General | 43.9 | 22.4 | 79.5 | 31.6 | 50.7 | 33.5 | 45.7 |
| Qwen2.5-72B-Instruct | Open-Source | General | 43.7 | 22.6 | 78.5 | 31.9 | 50.6 | 32.6 | 45.9 |
| Athene-V2-Chat | Open-Source | General | 43.7 | 22.5 | 78.3 | 32.6 | 51.2 | 32.3 | 45.4 |
| Qwen3-235B-A22B-Thinking | Open-Source | General | 43.5 | 22.3 | 77.3 | 30.5 | 52.9 | 33.3 | 44.8 |
| HuatuoGPT-o1-72B | Open-Source | Medical | 43.1 | 21.9 | 78.4 | 30.2 | 49.3 | 33.2 | 45.5 |
| Qwen3-32B-Thinking | Open-Source | General | 43.0 | 21.3 | 76.0 | 30.8 | 52.7 | 31.8 | 45.3 |
| Qwen3-30B-A3B-Thinking | Open-Source | General | 42.9 | 20.2 | 76.8 | 31.7 | 49.2 | 34.3 | 44.9 |
| medgemma-27b-it | Open-Source | Medical | 42.7 | 21.9 | 76.4 | 30.1 | 48.0 | 35.2 | 44.5 |
| Qwen3-14B-Thinking | Open-Source | General | 42.2 | 21.0 | 76.9 | 29.6 | 49.6 | 32.1 | 44.0 |
| Qwen3-4B-Thinking-2507 | Open-Source | General | 42.1 | 17.9 | 77.3 | 32.6 | 47.4 | 32.5 | 45.1 |
| DeepSeek-R1-Distill-Llama-70B | Open-Source | General | 42.0 | 19.7 | 76.7 | 30.4 | 49.5 | 30.1 | 45.4 |
| Qwen3-8B-Thinking | Open-Source | General | 41.9 | 19.7 | 76.4 | 30.0 | 49.4 | 32.5 | 43.3 |
| Llama-3.3-70B-Instruct | Open-Source | General | 41.8 | 22.6 | 75.6 | 30.0 | 46.3 | 32.8 | 43.6 |
| Qwen3-Next-80B-A3B-Instruct | Open-Source | General | 41.8 | 21.8 | 75.0 | 28.9 | 51.1 | 29.8 | 44.2 |
| Qwen2.5-32B-Instruct | Open-Source | General | 41.8 | 19.9 | 74.6 | 30.5 | 48.9 | 32.3 | 44.7 |
| Mistral-Small-3.1-24B-Instruct-2503 | Open-Source | General | 41.7 | 21.7 | 75.2 | 31.1 | 47.0 | 31.7 | 43.7 |
| DeepSeek-R1-Distill-Qwen-32B | Open-Source | General | 41.7 | 20.3 | 74.0 | 29.9 | 50.2 | 30.8 | 45.2 |
| gemma-3-27b-it | Open-Source | General | 41.6 | 22.2 | 70.8 | 30.7 | 47.2 | 34.3 | 44.5 |
| QWQ-32B | Open-Source | General | 41.6 | 20.3 | 75.1 | 30.4 | 51.5 | 26.5 | 46.0 |
| QwenLong-L1-32B | Open-Source | General | 41.4 | 19.9 | 74.6 | 30.5 | 51.0 | 27.4 | 45.1 |
| Qwen3-235B-A22B-Non-Thinking | Open-Source | General | 41.4 | 20.9 | 77.8 | 26.6 | 48.1 | 30.8 | 43.9 |
| Qwen3-32B-Non-Thinking | Open-Source | General | 41.4 | 20.9 | 76.0 | 27.2 | 48.9 | 30.8 | 44.3 |
| Llama-3.1-70B-Instruct | Open-Source | General | 41.2 | 22.4 | 76.6 | 27.2 | 46.9 | 30.3 | 44.0 |
| Qwen3-4B-Thinking | Open-Source | General | 40.5 | 18.9 | 74.4 | 29.0 | 47.4 | 30.2 | 43.3 |
| Baichuan-M2-32B | Open-Source | Medical | 40.3 | 17.7 | 74.4 | 31.3 | 44.9 | 28.1 | 45.5 |
| gemma-2-27b-it | Open-Source | General | 40.0 | 22.1 | 69.4 | 29.7 | 45.4 | 31.0 | 42.3 |
| gpt-oss-120b-high | Open-Source | General | 39.9 | 19.0 | **81.0** | 27.7 | 46.5 | 22.9 | 42.3 |
| Magistral-Small-2506 | Open-Source | General | 39.6 | 21.1 | 72.6 | 28.3 | 45.3 | 28.4 | 41.9 |
| gemma-3-12b-it | Open-Source | General | 39.3 | 19.3 | 70.2 | 28.2 | 42.8 | 31.5 | 43.7 |
| Qwen3-30B-A3B-Instruct-2507 | Open-Source | General | 39.2 | 19.4 | 74.5 | 26.0 | 43.6 | 27.9 | 43.6 |
| Mistral-Small-24B-Instruct-2501 | Open-Source | General | 39.1 | 21.1 | 69.5 | 30.8 | 43.6 | 31.7 | 37.6 |
| DeepSeek-R1-0528-Qwen3-8B | Open-Source | General | 39.0 | 17.5 | 70.3 | 27.3 | 45.6 | 28.9 | 44.3 |
| Qwen3-14B-Non-Thinking | Open-Source | General | 39.0 | 19.8 | 75.6 | 22.7 | 43.2 | 30.1 | 42.5 |
| Phi-4 | Open-Source | General | 38.4 | 19.8 | 73.9 | 25.3 | 43.0 | 26.5 | 42.0 |
| Qwen3-30B-A3B-Non-Thinking | Open-Source | General | 38.4 | 18.9 | 73.4 | 24.4 | 40.9 | 29.4 | 43.6 |
| Baichuan-M1-14B-Instruct | Open-Source | Medical | 38.4 | 19.4 | 75.2 | 24.9 | 41.6 | 27.7 | 41.5 |
| Mistral-Small-Instruct-2409 | Open-Source | General | 37.6 | 20.7 | 74.8 | 21.9 | 40.4 | 26.2 | 41.5 |
| Llama-4-Scout-17B-16E-Instruct | Open-Source | General | 37.6 | 19.8 | 73.8 | 22.2 | 40.8 | 25.9 | 42.9 |
| gpt-35-turbo | Proprietary | General | 37.2 | 21.0 | 66.1 | 27.0 | 39.3 | 28.8 | 40.9 |
| gemma-2-9b-it | Open-Source | General | 37.0 | 18.0 | 66.0 | 26.4 | 41.3 | 28.5 | 41.6 |
| DeepSeek-R1-Distill-Qwen-14B | Open-Source | General | 36.8 | 19.5 | 71.4 | 18.5 | 44.1 | 22.8 | 44.5 |
| Qwen3-4B-Instruct-2507 | Open-Source | General | 36.2 | 15.4 | 72.8 | 22.9 | 41.3 | 26.2 | 38.9 |
| Qwen3-8B-Non-Thinking | Open-Source | General | 36.2 | 16.0 | 76.0 | 18.0 | 40.6 | 27.0 | 39.6 |
| HuatuoGPT-o1-70B | Open-Source | Medical | 36.0 | 17.1 | 71.7 | 21.5 | 41.8 | 25.7 | 38.3 |
| Llama-3-70B-UltraMedical | Open-Source | Medical | 35.7 | 14.3 | 74.6 | 20.0 | 40.1 | 26.0 | 39.5 |
| Llama3-OpenBioLLM-70B | Open-Source | Medical | 35.6 | 17.0 | 74.2 | 21.8 | 34.8 | 25.0 | 40.5 |
| Qwen3-4B-Non-Thinking | Open-Source | General | 35.6 | 13.9 | 71.8 | 23.3 | 37.8 | 26.5 | 40.3 |
| Llama-3.1-Nemotron-70B-Instruct-HF | Open-Source | General | 35.5 | 16.6 | 78.3 | 15.5 | 39.5 | 22.8 | 40.0 |
| Yi-1.5-34B-Chat-16K | Open-Source | General | 34.8 | 16.2 | 72.6 | 19.4 | 35.1 | 23.7 | 41.7 |
| MeLLaMA-70B-chat | Open-Source | Medical | 34.7 | 16.6 | 69.5 | 22.7 | 35.8 | 23.4 | 39.9 |
| QwQ-32B-Preview | Open-Source | General | 34.1 | 15.8 | 71.4 | 18.0 | 37.6 | 23.3 | 38.8 |
| Qwen2.5-7B-Instruct | Open-Source | General | 33.6 | 14.1 | 68.7 | 19.9 | 37.5 | 23.4 | 37.8 |
| Ministral-8B-Instruct-2410 | Open-Source | General | 32.4 | 15.4 | 65.0 | 20.2 | 34.4 | 23.7 | 35.8 |
| gpt-oss-20b-high | Open-Source | General | 32.0 | 13.2 | 71.2 | 17.9 | 34.1 | 15.6 | 40.0 |
| HuatuoGPT-o1-7B | Open-Source | Medical | 31.9 | 12.9 | 69.2 | 16.0 | 36.7 | 21.3 | 35.1 |
| medgemma-4b-it | Open-Source | Medical | 31.8 | 12.7 | 67.7 | 18.2 | 33.2 | 21.7 | 37.5 |
| Phi-3.5-MoE-instruct | Open-Source | General | 31.7 | 11.8 | 67.8 | 17.1 | 38.4 | 20.7 | 34.5 |
| Llama-3.1-8B-Instruct | Open-Source | General | 31.3 | 13.1 | 67.3 | 16.0 | 34.0 | 21.4 | 36.1 |
| Yi-1.5-9B-Chat-16K | Open-Source | General | 31.1 | 12.9 | 60.4 | 20.2 | 31.5 | 21.5 | 40.4 |
| DeepSeek-R1-Distill-Llama-8B | Open-Source | General | 30.9 | 12.6 | 64.6 | 16.7 | 36.1 | 18.1 | 37.0 |
| gemma-3-4b-it | Open-Source | General | 30.6 | 13.2 | 58.2 | 19.0 | 33.3 | 21.7 | 38.4 |
| Qwen2.5-3B-Instruct | Open-Source | General | 28.8 | 10.7 | 58.6 | 18.8 | 30.4 | 18.1 | 36.4 |
| Qwen3-1.7B-Thinking | Open-Source | General | 28.7 | 10.9 | 56.0 | 16.9 | 30.2 | 17.8 | 40.0 |
| HuatuoGPT-o1-8B | Open-Source | Medical | 28.1 | 9.8 | 61.0 | 13.3 | 33.7 | 16.6 | 34.3 |
| Phi-3.5-mini-instruct | Open-Source | General | 27.8 | 7.9 | 58.5 | 11.5 | 33.0 | 17.5 | 38.2 |
| DeepSeek-R1-Distill-Qwen-7B | Open-Source | General | 27.4 | 11.6 | 53.0 | 16.8 | 30.8 | 15.7 | 36.7 |
| Llama-2-70b-chat | Open-Source | General | 27.4 | 13.0 | 57.9 | 13.0 | 29.6 | 16.6 | 34.2 |
| Phi-4-mini-instruct | Open-Source | General | 26.8 | 10.0 | 58.0 | 13.3 | 28.5 | 16.7 | 34.6 |
| Llama-3.2-3B-Instruct | Open-Source | General | 25.0 | 10.1 | 59.7 | 13.5 | 23.0 | 17.1 | 26.7 |
| Qwen2.5-1.5B-Instruct | Open-Source | General | 24.4 | 8.3 | 50.9 | 13.6 | 25.5 | 13.8 | 34.3 |
| Qwen3-1.7B-Non-Thinking | Open-Source | General | 23.7 | 6.5 | 47.2 | 15.0 | 25.2 | 17.3 | 31.0 |
| Phi-4-mini-reasoning | Open-Source | General | 23.5 | 8.4 | 56.7 | 9.5 | 28.6 | 10.8 | 27.1 |
| MeLLaMA-13B-chat | Open-Source | Medical | 23.2 | 7.1 | 56.4 | 11.7 | 23.1 | 11.2 | 29.8 |
| Llama-2-13b-chat | Open-Source | General | 22.9 | 10.0 | 51.8 | 11.0 | 24.3 | 13.1 | 27.4 |
| Qwen3-0.6B-Thinking | Open-Source | General | 22.7 | 8.5 | 48.6 | 11.2 | 22.6 | 11.8 | 33.5 |
| BioMistral-7B | Open-Source | Medical | 22.7 | 7.8 | 52.6 | 12.7 | 21.7 | 11.9 | 29.6 |
| MMed-Llama-3-8B | Open-Source | Medical | 22.3 | 5.8 | 58.1 | 7.6 | 26.5 | 13.8 | 22.0 |
| Llama-3.1-8B-UltraMedical | Open-Source | Medical | 22.3 | 4.8 | 54.3 | 10.9 | 27.3 | 9.9 | 26.9 |
| OpenThinker3-7B | Open-Source | General | 22.1 | 8.8 | 52.0 | 6.1 | 25.4 | 10.3 | 29.9 |
| MedReason-8B | Open-Source | Medical | 20.3 | 6.7 | 49.6 | 7.5 | 23.2 | 8.8 | 26.1 |
| Llama-2-7b-chat | Open-Source | General | 18.2 | 8.1 | 39.6 | 8.7 | 19.0 | 9.4 | 24.3 |
| Qwen3-0.6B-Non-Thinking | Open-Source | General | 17.5 | 5.3 | 42.6 | 8.2 | 15.8 | 5.6 | 27.8 |
| gemma-3-1b-it | Open-Source | General | 17.4 | 4.7 | 33.5 | 8.7 | 17.8 | 11.2 | 28.3 |
| meditron-70b | Open-Source | Medical | 17.3 | 4.4 | 42.9 | 5.6 | 21.9 | 9.2 | 19.6 |
| Llama3-OpenBioLLM-8B | Open-Source | Medical | 16.5 | 4.8 | 45.7 | 3.0 | 18.7 | 3.4 | 23.3 |
| DeepSeek-R1-Distill-Qwen-1.5B | Open-Source | General | 16.0 | 4.1 | 34.2 | 5.6 | 19.3 | 7.0 | 25.8 |
| Llama-3.2-1B-Instruct | Open-Source | General | 14.7 | 5.6 | 38.3 | 5.6 | 11.3 | 5.9 | 21.2 |
| meditron-7b | Open-Source | Medical | 11.1 | 3.6 | 31.1 | 1.9 | 11.6 | 2.6 | 15.9 |

*Abbr for Clinical Stage: D&A = Discharge and Administration, D&P = Diagnosis and Prognosis, IA = Initial Assessment, T&I = Treatment and Intervention, and T&R = Triage and Referral.*

### 3.6 Performance analysis of output length under CoT prompting

To investigate the relationship between CoT output length and model performance, we analyzed model outputs generated under CoT prompting and quantified output length as the number of words per generated response. We focused on 21 recent large-scale LLMs ($\geq$ 30B parameters; released after 2025-01-01) and text classification tasks (with standardized and relatively short answers). Therefore, we include four languages with at least 2 text classification tasks: English (12), Chinese (6), Spanish (2), and Portuguese (2).

As shown in Figure S1, we found that reasoning LLMs typically generate longer CoT outputs and tend to achieve higher performance across these languages. Overall, longer CoT outputs were associated with improved accuracy, and the general trend was consistent across the four languages.

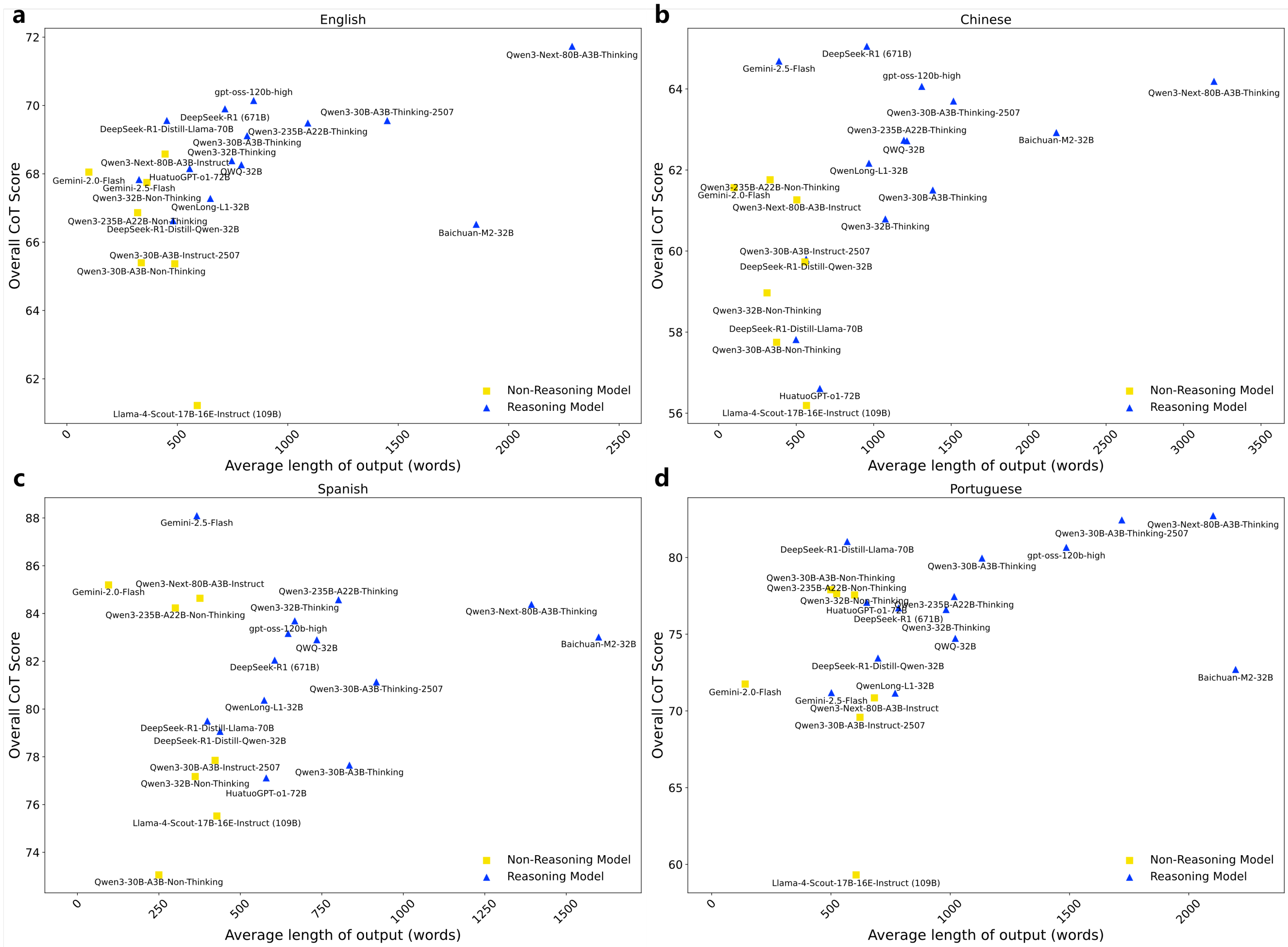

**Figure S1.** Analysis of the relationship between CoT output length and performance across languages: (a) English, (b) Chinese, (c) Spanish, and (d) Portuguese.

### 3.7 Analysis of Model Output Validity

We defined the standardized output format for each task using templates (See Section Methods and Supplementary Section), and developed automated scripts to extract results from the structured outputs accordingly. Outputs that failed to follow the required formatting were considered *invalid*. For each inference strategy, we calculated the valid response rate for each model, as shown in Table S8.

Under the zero-shot setting, 88 out of 95 models achieved a valid response rate exceeding 80%. After integrating few-shot prompting, 81 models exhibited improved structural adherence, resulting in 91 models surpassing the 80% threshold. In contrast, CoT prompting led to a decrease in valid response rates for 62 models, suggesting that CoT reasoning sometimes introduces formatting inconsistencies.

Moreover, larger LLMs were generally more capable of producing format-compliant outputs, reflecting stronger instruction-following abilities. However, some medical-specialized LLMs (e.g., BioMistral and Me-LLaMA) demonstrated poorer general instruction-following across diverse tasks compared to their general-purpose counterparts.

## **Table S8.** Overall Valid Rate of Generated response

| Model | Model Type | Model Domain | Score Zero-shot | Score CoT | Score 5-shot | Δ Score Few-shot | Δ Score (%) Few-shot | Δ Score CoT | Δ Score (%) CoT |
|---|---|---|---|---|---|---|---|---|---|
| Athene-V2-Chat | Open-Source | General | **99.9** | 98.3 | **99.9** | 0.0 | 0.0 | −1.6 | −1.6 |
| Qwen2.5-72B-Instruct | Open-Source | General | **99.9** | 98.2 | **99.9** | 0.0 | 0.0 | −1.7 | −1.7 |
| Qwen3-32B-Non-Thinking | Open-Source | General | 99.8 | 98.6 | **99.9** | 0.1 | 0.1 | −1.2 | −1.2 |
| HuatuoGPT-o1-72B | Open-Source | Medical | 99.8 | 99.4 | **99.9** | 0.1 | 0.1 | −0.4 | −0.4 |
| Qwen3-Next-80B-A3B-Instruct | Open-Source | General | 99.8 | 99.5 | **99.9** | 0.1 | 0.1 | −0.3 | −0.3 |
| Qwen2.5-32B-Instruct | Open-Source | General | 99.6 | 97.0 | 99.8 | 0.2 | 0.2 | −2.6 | −2.6 |
| Qwen3-Next-80B-A3B-Thinking | Open-Source | General | 99.6 | 99.4 | 99.8 | 0.2 | 0.2 | −0.2 | −0.2 |
| Mistral-Large-Instruct-2411 | Open-Source | General | 99.6 | 99.2 | 99.7 | 0.1 | 0.1 | −0.4 | −0.4 |
| Qwen3-32B-Thinking | Open-Source | General | 99.5 | 99.4 | 99.7 | 0.2 | 0.2 | −0.1 | −0.1 |
| gemini-1.5-pro | Proprietary | General | 99.4 | 98.9 | 99.7 | 0.3 | 0.3 | −0.5 | −0.5 |
| Qwen3-235B-A22B-Non-Thinking | Open-Source | General | 99.3 | 96.6 | **99.9** | 0.6 | 0.6 | −2.7 | −2.7 |
| Magistral-Small-2506 | Open-Source | General | 99.3 | 98.7 | 99.6 | 0.3 | 0.3 | −0.6 | −0.6 |
| DeepSeek-R1 | Open-Source | General | 99.3 | 98.9 | 99.0 | −0.3 | −0.3 | −0.4 | −0.4 |
| DeepSeek-R1-Distill-Qwen-32B | Open-Source | General | 99.2 | 98.2 | 97.4 | −1.8 | −1.8 | −1.0 | −1.0 |
| gpt-4o | Proprietary | General | 99.2 | 98.7 | 99.8 | 0.6 | 0.6 | −0.5 | −0.5 |
| medgemma-27b-it | Open-Source | Medical | 99.1 | 99.2 | **99.9** | 0.8 | 0.8 | 0.1 | 0.1 |
| gemini-2.5-flash | Proprietary | General | 98.9 | 98.9 | 99.7 | 0.8 | 0.8 | 0.0 | 0.0 |
| Phi-4 | Open-Source | General | 98.9 | 94.4 | 99.1 | 0.2 | 0.2 | −4.5 | −4.6 |
| DeepSeek-R1-Distill-Llama-70B | Open-Source | General | 98.9 | 98.7 | 99.4 | 0.5 | 0.5 | −0.2 | −0.2 |
| Llama-3.3-70B-Instruct | Open-Source | General | 98.8 | 99.3 | 99.7 | 0.9 | 0.9 | 0.5 | 0.5 |
| Baichuan-M2-32B | Open-Source | Medical | 98.8 | 98.3 | 99.5 | 0.7 | 0.7 | −0.5 | −0.5 |
| Mistral-Small-Instruct-2409 | Open-Source | General | 98.8 | 98.7 | 92.9 | −5.9 | −6.0 | −0.1 | −0.1 |
| Qwen3-235B-A22B-Thinking | Open-Source | General | 98.7 | 97.9 | **99.9** | 1.2 | 1.2 | −0.8 | −0.8 |
| gemma-2-27b-it | Open-Source | General | 98.7 | 95.8 | 98.1 | −0.6 | −0.6 | −2.9 | −2.9 |
| Baichuan-M1-14B-Instruct | Open-Source | Medical | 98.7 | 97.0 | 99.6 | 0.9 | 0.9 | −1.7 | −1.7 |
| Qwen3-30B-A3B-Instruct-2507 | Open-Source | General | 98.7 | 98.9 | 99.8 | 1.1 | 1.1 | 0.2 | 0.2 |
| Qwen3-4B-Thinking-2507 | Open-Source | General | 98.5 | 98.9 | 99.8 | 1.3 | 1.3 | 0.4 | 0.4 |
| gemma-3-27b-it | Open-Source | General | 98.5 | **99.7** | **99.9** | 1.4 | 1.4 | 1.2 | 1.2 |
| gpt-35-turbo | Proprietary | General | 98.4 | 97.3 | 98.9 | 0.5 | 0.5 | −1.1 | −1.1 |
| Qwen3-14B-Non-Thinking | Open-Source | General | 98.4 | 98.1 | 99.8 | 1.4 | 1.4 | −0.3 | −0.3 |
| Mistral-Small-3.1-24B-Instruct-2503 | Open-Source | General | 98.0 | 95.6 | **99.9** | 1.9 | 1.9 | −2.4 | −2.4 |
| Yi-1.5-34B-Chat-16K | Open-Source | General | 97.8 | 93.0 | 99.4 | 1.6 | 1.6 | −4.8 | −4.9 |
| gpt-oss-120b-high | Open-Source | General | 97.7 | 97.9 | 98.6 | 0.9 | 0.9 | 0.2 | 0.2 |
| gpt-oss-20b-high | Open-Source | General | 97.7 | 95.5 | 96.2 | −1.5 | −1.5 | −2.2 | −2.3 |
| Qwen3-14B-Thinking | Open-Source | General | 97.7 | 99.0 | 99.7 | 2.0 | 2.0 | 1.3 | 1.3 |
| QwenLong-L1-32B | Open-Source | General | 97.6 | 93.6 | 96.8 | −0.8 | −0.8 | −4.0 | −4.1 |
| HuatuoGPT-o1-70B | Open-Source | Medical | 97.6 | 96.1 | 99.4 | 1.8 | 1.8 | −1.5 | −1.5 |
| Ministral-8B-Instruct-2410 | Open-Source | General | 97.6 | 92.1 | 98.7 | 1.1 | 1.1 | −5.5 | −5.6 |
| Qwen3-8B-Thinking | Open-Source | General | 97.4 | 97.5 | 99.7 | 2.3 | 2.4 | 0.1 | 0.1 |
| Qwen3-4B-Thinking | Open-Source | General | 97.2 | 98.9 | 99.5 | 2.3 | 2.4 | 1.7 | 1.7 |
| Qwen3-4B-Instruct-2507 | Open-Source | General | 97.1 | 98.0 | 99.5 | 2.4 | 2.5 | 0.9 | 0.9 |
| Qwen3-30B-A3B-Thinking | Open-Source | General | 96.9 | 98.3 | 99.1 | 2.2 | 2.3 | 1.4 | 1.4 |
| gemma-3-12b-it | Open-Source | General | 96.7 | 99.2 | 99.7 | 3.0 | 3.1 | 2.5 | 2.6 |
| HuatuoGPT-o1-8B | Open-Source | Medical | 96.7 | 89.7 | 98.8 | 2.1 | 2.2 | −7.0 | −7.2 |
| DeepSeek-R1-0528-Qwen3-8B | Open-Source | General | 96.7 | 96.8 | 99.2 | 2.5 | 2.6 | 0.1 | 0.1 |
| gemma-2-9b-it | Open-Source | General | 96.4 | 89.7 | 98.3 | 1.9 | 2.0 | −6.7 | −7.0 |
| Yi-1.5-9B-Chat-16K | Open-Source | General | 96.2 | 91.5 | 99.1 | 2.9 | 3.0 | −4.7 | −4.9 |
| QwQ-32B-Preview | Open-Source | General | 95.6 | 87.9 | 99.6 | 4.0 | 4.2 | −7.7 | −8.1 |
| Qwen3-4B-Non-Thinking | Open-Source | General | 95.4 | 97.8 | 99.4 | 4.0 | 4.2 | 2.4 | 2.5 |
| QWQ-32B | Open-Source | General | 95.4 | 92.1 | 98.6 | 3.2 | 3.4 | −3.3 | −3.5 |
| Qwen3-30B-A3B-Non-Thinking | Open-Source | General | 95.3 | 96.8 | 99.7 | 4.4 | 4.6 | 1.5 | 1.6 |
| Qwen2.5-3B-Instruct | Open-Source | General | 95.2 | 93.3 | 98.6 | 3.4 | 3.6 | −1.9 | −2.0 |
| Qwen3-30B-A3B-Thinking-2507 | Open-Source | General | 95.1 | 97.6 | 99.8 | 4.7 | 4.9 | 2.5 | 2.6 |
| Llama-4-Scout-17B-16E-Instruct | Open-Source | General | 94.6 | 94.8 | 99.1 | 4.5 | 4.8 | 0.2 | 0.2 |
| Llama-3.1-70B-Instruct | Open-Source | General | 94.5 | 95.9 | 99.7 | 5.2 | 5.5 | 1.4 | 1.5 |
| Qwen3-1.7B-Non-Thinking | Open-Source | General | 94.5 | 89.0 | 99.0 | 4.5 | 4.8 | −5.5 | −5.8 |
| gemini-2.0-flash | Proprietary | General | 94.4 | 96.5 | **99.9** | 5.5 | 5.8 | 2.1 | 2.2 |
| MeLLaMA-70B-chat | Open-Source | Medical | 94.2 | 87.4 | 83.7 | −10.5 | −11.1 | −6.8 | −7.2 |
| medgemma-4b-it | Open-Source | Medical | 93.8 | 94.2 | 98.4 | 4.6 | 4.9 | 0.4 | 0.4 |
| Llama3-OpenBioLLM-70B | Open-Source | Medical | 93.7 | 91.0 | 99.0 | 5.3 | 5.7 | −2.7 | −2.9 |
| Llama-3.2-3B-Instruct | Open-Source | General | 93.6 | 86.3 | 98.1 | 4.5 | 4.8 | −7.3 | −7.8 |
| Qwen3-1.7B-Thinking | Open-Source | General | 93.2 | 90.8 | 98.8 | 5.6 | 6.0 | −2.4 | −2.6 |
| Qwen3-8B-Non-Thinking | Open-Source | General | 92.8 | 95.1 | 99.7 | 6.9 | 7.4 | 2.3 | 2.5 |
| Llama-3.1-Nemotron-70B-Instruct-HF | Open-Source | General | 92.5 | 91.8 | 97.6 | 5.1 | 5.5 | −0.7 | −0.8 |
| gemma-3-4b-it | Open-Source | General | 91.9 | 95.7 | 98.5 | 6.6 | 7.2 | 3.8 | 4.1 |
| Phi-3.5-MoE-instruct | Open-Source | General | 91.8 | 86.7 | 96.1 | 4.3 | 4.7 | −5.1 | −5.6 |
| Qwen2.5-1.5B-Instruct | Open-Source | General | 91.6 | 82.3 | 99.0 | 7.4 | 8.1 | −9.3 | −10.2 |
| DeepSeek-R1-Distill-Qwen-7B | Open-Source | General | 91.5 | 89.6 | 76.5 | −15.0 | −16.4 | −1.9 | −2.1 |
| BioMistral-7B | Open-Source | Medical | 90.6 | 24.4 | 92.7 | 2.1 | 2.3 | −66.2 | −73.1 |
| Qwen2.5-7B-Instruct | Open-Source | General | 90.4 | 92.5 | 99.1 | 8.7 | 9.6 | 2.1 | 2.3 |
| HuatuoGPT-o1-7B | Open-Source | Medical | 90.1 | 94.1 | 99.2 | 9.1 | 10.1 | 4.0 | 4.4 |
| Mistral-Small-24B-Instruct-2501 | Open-Source | General | 90.1 | 78.7 | 99.7 | 9.6 | 10.7 | −11.4 | −12.7 |
| Llama-3-70B-UltraMedical | Open-Source | Medical | 89.8 | 93.9 | 98.3 | 8.5 | 9.5 | 4.1 | 4.6 |
| MedReason-8B | Open-Source | Medical | 89.6 | 89.0 | 98.1 | 8.5 | 9.5 | −0.6 | −0.7 |
| Phi-3.5-mini-instruct | Open-Source | General | 88.8 | 84.8 | 94.7 | 5.9 | 6.6 | −4.0 | −4.5 |
| Phi-4-mini-reasoning | Open-Source | General | 88.3 | 83.8 | 66.9 | −21.4 | −24.2 | −4.5 | −5.1 |
| DeepSeek-R1-Distill-Qwen-14B | Open-Source | General | 88.2 | 94.5 | 98.0 | 9.8 | 11.1 | 6.3 | 7.1 |
| DeepSeek-R1-Distill-Llama-8B | Open-Source | General | 88.0 | 91.4 | 93.1 | 5.1 | 5.8 | 3.4 | 3.9 |
| Llama-2-7b-chat | Open-Source | General | 86.7 | 72.3 | 86.6 | −0.1 | −0.1 | −14.4 | −16.6 |
| Phi-4-mini-instruct | Open-Source | General | 85.9 | 77.5 | 88.5 | 2.6 | 3.0 | −8.4 | −9.8 |
| MMed-Llama-3-8B | Open-Source | Medical | 85.8 | 67.4 | 98.8 | 13.0 | 15.2 | −18.4 | −21.4 |
| Llama-2-70b-chat | Open-Source | General | 84.8 | 74.2 | 84.0 | −0.8 | −0.9 | −10.6 | −12.5 |
| Llama-3.1-8B-Instruct | Open-Source | General | 84.4 | 92.2 | 98.5 | 14.1 | 16.7 | 7.8 | 9.2 |
| Qwen3-0.6B-Thinking | Open-Source | General | 84.3 | 85.5 | 97.0 | 12.7 | 15.1 | 1.2 | 1.4 |
| MeLLaMA-13B-chat | Open-Source | Medical | 83.8 | 83.2 | 82.1 | −1.7 | −2.0 | −0.6 | −0.7 |
| Llama-2-13b-chat | Open-Source | General | 83.6 | 76.3 | 80.8 | −2.8 | −3.3 | −7.3 | −8.7 |
| Qwen3-0.6B-Non-Thinking | Open-Source | General | 82.5 | 80.7 | 98.4 | 15.9 | 19.3 | −1.8 | −2.2 |
| OpenThinker3-7B | Open-Source | General | 80.3 | 81.3 | 89.1 | 8.8 | 11.0 | 1.0 | 1.2 |
| meditron-70b | Open-Source | Medical | 75.6 | 62.9 | 90.2 | 14.6 | 19.3 | −12.7 | −16.8 |
| gemma-3-1b-it | Open-Source | General | 74.6 | 73.3 | 94.3 | 19.7 | 26.4 | −1.3 | −1.7 |
| Llama-3.1-8B-UltraMedical | Open-Source | Medical | 73.5 | 75.8 | 82.1 | 8.6 | 11.7 | 2.3 | 3.1 |
| DeepSeek-R1-Distill-Qwen-1.5B | Open-Source | General | 69.7 | 66.6 | 65.8 | −3.9 | −5.6 | −3.1 | −4.4 |
| Llama-3.2-1B-Instruct | Open-Source | General | 62.6 | 58.2 | 95.1 | 32.5 | 51.9 | −4.4 | −7.0 |
| Llama3-OpenBioLLM-8B | Open-Source | Medical | 56.1 | 50.3 | 95.7 | 39.6 | 70.6 | −5.8 | −10.3 |
| meditron-7b | Open-Source | Medical | 50.8 | 52.1 | 72.0 | 21.2 | 41.7 | 1.3 | 2.6 |

*Note: Models are ranked by the average valid rate across all tasks under zero-shot inference ("Score Zero-shot") in descending order. The **bold blue** highlights the highest value for each column. Δ represents the difference compared to zero-shot, with green indicating improvement and red indicating a decline.*

## 4 Data Contamination Analysis

To investigate potential data contamination, each LLM was prompted with a partial segment from the input of every test sample and asked to generate the next five tokens.[16] This process was repeated five times per sample. A sample was flagged as potentially contaminated if the model reproduced an exact five-token match in at least three of the five trials. Based on this criterion, we calculated the proportion of potentially leaked samples within each dataset.

We report the contamination ratios across all tasks for the largest model within each model family. As shown in Figure S2, the overall risk of data contamination was low across most clinical text tasks, indicating that the majority of benchmark datasets were not previously exposed to the evaluated models. However, certain datasets were used for model training and exhibited elevated contamination likelihood, such as dataset HealthCareMagi-100k for Llama-3-70B-UltraMedical,[17] and n2c2 2018–ADE&Medication for Me-LLaMA-70B.[18]

Overall, these results support the integrity and reliability of the BRIDGE, suggesting that it provides a robust benchmark for evaluating clinical text understanding.

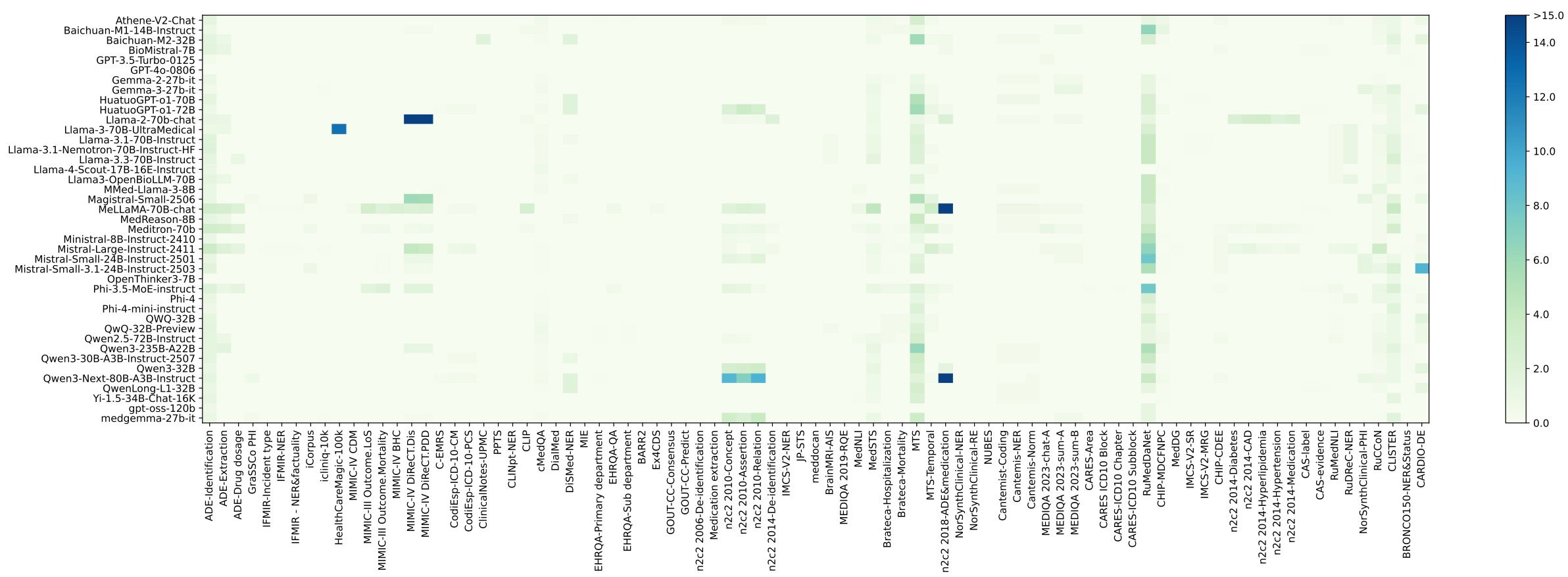

**Figure S2.** Data Contamination Analysis.
(Each cell represents the proportion of potentially leaked samples for a given model-task pair. Deeper color indicates a higher likelihood that the dataset was used during the model's training.)

## 5 BRIDGE Construction

### 5.1 Prompt Template

We designed a simple and standardized prompt template to generate appropriate instructions, inputs, and output formats for each task (see Section Methods for details). The metadata of all tasks can be found in Table S9. The general structure of the template is as follows:

> Given **[Input description]** in **[Language (if not English)]**, **[Task description]**.
> **[Requirement for inference strategy]**
> **[Output format]**

Here, **[Input description]** is a short phrase describing the input data (e.g., clinical text, radiology report). **[Language]** indicates the input language (e.g., Chinese, Spanish); this part is omitted if the input is in English. **Task description** provides a detailed explanation of the task and relevant information, including label definitions sourced from the original dataset. **[Output format]** defines the expected structure of the model's response to facilitate evaluation.

To investigate the effect of different inference strategies, we modified only the **[Requirement for inference strategy]** field while keeping the rest of the prompt unchanged. The corresponding instruction template is as below:

> **Zero / Few-shot**: *Return your answer in the following format. DO NOT GIVE ANY EXPLANATION:*
> **Chian-of-Thought**: *Solve it in a step-by-step fashion, return your answer in the following format, PROVIDE DETAILED ANALYSIS BEFORE THE RESULT:*

Specifically: For zero-shot and few-shot settings, the model is instructed to output the final answer directly. For CoT, the model is prompted to first provide reasoning before producing the final answer.

### 5.2 Task Taxonomy and Characteristics

To systematically investigate the performance of LLMs across different clinical scenarios, we extracted the following key characteristics for each task. Then, we categorized them into standardized taxonomies, enabling a comprehensive analysis of LLM capabilities among different settings. All metadata of the BRIDGE task is in Table S9.

#### *5.2.1 Language*

Each task is monolingual and is determined by the original source documents.

#### *5.2.2 Sourced Clinical Documents*

We record the originating clinical documents from which each dataset is constructed (e.g., admission notes and discharge summaries). If a dataset spans multiple document types, it is assigned to each relevant category. Datasets referencing more than five document types or without clear specifications (e.g., generic EHR references) are labeled "General".

#### *5.2.3 Clinical Specialty*

Tasks are categorized into specific clinical specialties, which are defined by the specific challenges they address (e.g., targeted diseases) and the originating department of the source data. Examples include cardiology, radiology, or intensive care units (ICUs). Similar to clinical document types, datasets spanning more than five clinical specialties or lacking explicit domain definitions are labeled as"General."

#### *5.2.4 Clinical Stages and Applications*

Tasks are first grouped based on their specific clinical applications and then mapped into six clinical stages. This categorization is informed by clinical expertise and reflects the typical course of patient care. It also considers both the source of the clinical note (e.g., admission notes or discharge summaries) and the nature and objective of the task itself.

**Table S9. Metadata of BRIDGE Tasks**

| Task name | Language | Task Type | Sourced Clinical Document | Clinical Specialty | Clinical Application | Clinical Stage | Number of Testing Samples | Word Count (Input / Output) |
|---|---|---|---|---|---|---|---|---|
| MIMIC-IV CDM | English | Text Classification | General EHR Note | Gastroenterology | Diagnosis | Diagnosis and Prognosis | 240 | 817.56 / 2.00 |
| MIMIC-III Outcome.LoS | English | Text Classification | Discharge Summary | Critical Care | Prognosis | Diagnosis and Prognosis | 1000 | 465.35 / 4.00 |
| MIMIC-III Outcome.Mortality | English | Text Classification | Discharge Summary | Critical Care | Prognosis | Diagnosis and Prognosis | 1000 | 458.88 / 4.00 |
| MIMIC-IV BHC | English | Summarization | General EHR Note | Critical Care | Summarization | Discharge and Administration | 1000 | 1223.33 / 335.24 |
| MIMIC-IV DiReCT.Dis | English | Text Classification | General EHR Note | Cardiology, Gastroenterology, Neurology, Pulmonology, Endocrinology | Diagnosis | Diagnosis and Prognosis | 485 | 588.39 / 3.09 |
| MIMIC-IV DiReCT.PDD | English | Text Classification | General EHR Note | Cardiology, Gastroenterology, Neurology, Pulmonology, Endocrinology | Diagnosis | Diagnosis and Prognosis | 485 | 589.93 / 3.78 |
| ADE-Identification | English | Text Classification | Case Report | Pharmacology | ADE & Incidents | Treatment and Intervention | 2093 | 19.55 / 4.00 |
| ADE-Extraction | English | Event Extraction | Case Report | Pharmacology | ADE & Incidents | Treatment and Intervention | 428 | 19.40 / 9.89 |
| ADE-Drug dosage | English | Event Extraction | Case Report | Pharmacology | Medication information | Treatment and Intervention | 193 | 24.89 / 6.97 |
| BARR2 | Spanish | Event Extraction | Case Report | General | Concept standarization | Research | 214 | 394.68 / 81.08 |
| BrainMRI-AIS | English | Text Classification | Radiology Report | Neurology, Radiology | Diagnosis | Diagnosis and Prognosis | 267 | 54.72 / 2.00 |
| Brateca-Hospitalization | Portugues | Text Classification | General EHR Note | General | Prognosis | Diagnosis and Prognosis | 3183 | 1514.92 / 4.00 |
| Brateca-Mortality | Portugues | Text Classification | General EHR Note | General | Prognosis | Diagnosis and Prognosis | 3170 | 1507.12 / 2.00 |
| Cantemist-Coding | Spanish | Normalization and Coding | Case Report | Oncology | Billing & Coding | Discharge and Administration | 300 | 743.44 / 12.86 |
| Cantemis-NER | Spanish | Named Entity Recognition | Case Report | Oncology | Billing & Coding | Discharge and Administration | 297 | 743.89 / 77.66 |
| Cantemis-Norm | Spanish | Normalization and Coding | Case Report | Oncology | Billing & Coding | Discharge and Administration | 297 | 743.89 / 77.90 |
| CARES-Area | Spanish | Text Classification | Radiology Report | Radiology | Billing & Coding | Discharge and Administration | 966 | 214.15 / 3.00 |
| CARES ICD10 Block | Spanish | Normalization and Coding | Radiology Report | Radiology | Billing & Coding | Discharge and Administration | 966 | 214.15 / 6.96 |
| CARES-ICD10 Chapter | Spanish | Normalization and Coding | Radiology Report | Radiology | Billing & Coding | Discharge and Administration | 966 | 214.15 / 4.57 |
| CARES-ICD10 Subblock | Spanish | Normalization and Coding | Radiology Report | Radiology | Billing & Coding | Discharge and Administration | 966 | 214.15 / 8.27 |
| CHIP-CDEE | Chinese | Event Extraction | General EHR Note | General | Temporal/Causality determination | Initial Assessment | 384 | 60.62 / 63.65 |
| C-EMRS | Chinese | Text Classification | General EHR Note | Radiology, Endocrinology, Pulmonology, Cardiology, Gastroenterology | Diagnosis | Diagnosis and Prognosis | 1819 | 577.47 / 4.71 |
| CodiEsp-ICD-10-CM | Spanish | Normalization and Coding | Case Report | General | Billing & Coding | Discharge and Administration | 250 | 378.92 / 116.88 |
| CodiEsp-ICD-10-PCS | Spanish | Normalization and Coding | Case Report | General | Billing & Coding | Discharge and Administration | 224 | 388.62 / 48.35 |
| ClinicalNotes-UPMC | English | Text Classification | General EHR Note | General | Negation identification | Research | 229 | 15.61 / 2.00 |
| PPTS | Spanish | Text Classification | General EHR Note | Pulmonology | Diagnosis | Diagnosis and Prognosis | 153 | 377.97 / 2.00 |
| CLINpt-NER | Portuguese | Named Entity Recognition | Case Report, Admission Note, Discharge Summary | Neurology | Procudure information | Treatment and Intervention | 260 | 183.81 / 297.13 |
| CLIP | English | Text Classification | Discharge Summary | Critical Care | Post-discharge patient management | Discharge and Administration | 933 | 20.84 / 3.90 |
| cMedQA | Chinese | Question Answering | Consultation Record | General | Screen & Consultation | Triage and Referral | 6184 | 49.16 / 91.76 |
| DialMed | Chinese | Text Classification | Consultation Record | Pulmonology, Gastroenterology, Dermatology, Pharmacology | Medication information | Treatment and Intervention | 1199 | 188.23 / 9.20 |
| DiSMed-NER | Spanish | Named Entity Recognition | Radiology Report | Radiology | De-identification | Research | 212 | 233.43 / 84.59 |
| MIE | Chinese | Event Extraction | Consultation Record | Cardiology | Phenotyping | Initial Assessment | 2084 | 97.73 / 28.34 |
| EHRQA-Primary department | Chinese | Text Classification | Consultation Record | General | Specialist recommendation | Triage and Referral | 5037 | 69.14 / 3.32 |
| EHRQA-QA | Chinese | Question Answering | Consultation Record | General | Screen & Consultation | Triage and Referral | 5097 | 118.78 / 92.99 |
| EHRQA-Sub department | Chinese | Text Classification | Consultation Record | General | Specialist recommendation | Triage and Referral | 5002 | 69.10 / 4.32 |
| Ex4CDS | German | Named Entity Recognition | General EHR Note | Nephrology | Procudure information | Treatment and Intervention | 398 | 20.35 / 33.24 |
| GOUT-CC-Consensus | English | Text Classification | Admission Note | Endocrinology | Diagnosis | Diagnosis and Prognosis | 425 | 22.85 / 3.00 |
| n2c2 2006-De-identification | English | Named Entity Recognition | Discharge Summary | Pulmonology | De-identification | Research | 220 | 738.49 / 141.48 |
| Medication extraction | English | Event Extraction | Discharge Summary | Pharmacology | Medication information | Treatment and Intervention | 229 | 1198.59 / 456.20 |
| n2c2 2010-Concept | English | Named Entity Recognition | Discharge Summary, Progress Note | Critical Care | Procudure information | Treatment and Intervention | 210 | 804.84 / 459.21 |
| n2c2 2010-Assertion | English | Named Entity Recognition | Discharge Summary, Progress Note | Critical Care | Post-discharge patient management | Discharge and Administration | 253 | 1004.96 / 259.55 |
| n2c2 2010-Relation | English | Event Extraction | Discharge Summary, Progress Note | Critical Care | Procudure information | Treatment and Intervention | 236 | 1042.94 / 267.09 |
| n2c2 2014-De-identification | English | Named Entity Recognition | General EHR Note | Endocrinology | De-identification | Research | 514 | 705.28 / 134.95 |
| IMCS-V2-NER | Chinese | Named Entity Recognition | Consultation Record | Pediatrics | Phenotyping | Initial Assessment | 2362 | 38.65 / 20.80 |
| JP-STS | Japanese | Semantic Similarity | Case Report | General | Semantic relation | Research | 363 | 53.57 / 3.00 |
| meddocan | Spanish | Named Entity Recognition | Case Report | General | De-identification | Research | 250 | 465.63 / 145.28 |
| MEDIQA 2019-RQE | English | Natural Language Inference | Consultation Record | General | Screen & Consultation | Triage and Referral | 230 | 58.78 / 2.00 |
| MedNLI | English | Natural Language Inference | General EHR Note | Critical Care | Semantic relation | Research | 1421 | 26.78 / 2.00 |
| MedSTS | English | Semantic Similarity | General EHR Note | General | Semantic relation | Research | 730 | 44.04 / 3.00 |
| MTS | English | Text Classification | Case Report | General | Data organization | Research | 235 | 524.53 / 4.93 |
| MTS-Temporal | English | Named Entity Recognition | Discharge Summary | Pediatrics, Psychology | Temporal/Causality determination | Initial Assessment | 274 | 672.05 / 105.01 |
| n2c2 2018-ADE&medication | English | Event Extraction | Discharge Summary | Pharmacology | ADE & Incidents | Treatment and Intervention | 202 | 2175.39 / 326.34 |
| NorSynthClinical-NER | Norwegian | Named Entity Recognition | General EHR Note | Cardiology | Temporal/Causality determination | Initial Assessment | 457 | 15.90 / 22.51 |
| NorSynthClinical-RE | Norwegian | Event Extraction | General EHR Note | Cardiology | Temporal/Causality determination | Initial Assessment | 457 | 15.90 / 37.89 |
| NUBES | Spanish | Event Extraction | General EHR Note | General | Negation identification | Research | 427 | 853.73 / 174.73 |
| MEDIQA 2023-chat-A | English | Summarization | Consultation Record | General | Encounter summarization | Initial Assessment | 200 | 111.25 / 43.88 |
| MEDIQA 2023-sum-A | English | Text Classification | Consultation Record | General | Data organization | Research | 199 | 118.51 / 3.85 |
| MEDIQA 2023-sum-B | English | Summarization | Consultation Record | General | Encounter summarization | Initial Assessment | 198 | 119.05 / 45.27 |
| RuMedDaNet | Russian | Natural Language Inference | General EHR Note | General | Screen & Consultation | Triage and Referral | 256 | 70.20 / 2.00 |
| CBLUE-CDN | Chinese | Normalization and Coding | General EHR Note | General | Billing & Coding | Discharge and Administration | 2000 | 10.51 / 11.97 |
| CHIP-CTC | Chinese | Text Classification | General EHR Note | General | Clinical trial matching | Research | 6080 | 17.78 / 6.36 |
| CHIP-MDCFNPC | Chinese | Event Extraction | Consultation Record | General | Phenotyping | Initial Assessment | 11060 | 25.56 / 13.64 |
| MedDG | Chinese | Question Answering | Consultation Record | Gastroenterology | Screen & Consultation | Triage and Referral | 2747 | 198.88 / 26.79 |
| IMCS-V2-SR | Chinese | Event Extraction | Consultation Record | Pediatrics | Phenotyping | Initial Assessment | 832 | 622.62 / 47.05 |
| IMCS-V2-MRG | Chinese | Summarization | Consultation Record | Pediatrics | Encounter summarization | Initial Assessment | 833 | 622.20 / 73.88 |
| IMCS-V2-DAC | Chinese | Text Classification | Consultation Record | Pediatrics | Screen & Consultation | Triage and Referral | 19333 | 16.74 / 7.58 |
| n2c2 2014-Diabetes | English | Event Extraction | General EHR Note | Cardiology, Endocrinology | Risk factor extraction | Initial Assessment | 364 | 765.71 / 33.77 |
| n2c2 2014-CAD | English | Event Extraction | General EHR Note | Cardiology, Endocrinology | Risk factor extraction | Initial Assessment | 226 | 750.21 / 40.48 |
| n2c2 2014-Hyperlipidemia | English | Event Extraction | General EHR Note | Cardiology, Endocrinology | Risk factor extraction | Initial Assessment | 249 | 812.20 / 28.11 |
| n2c2 2014-Hypertension | English | Event Extraction | General EHR Note | Cardiology, Endocrinology | Risk factor extraction | Initial Assessment | 393 | 747.99 / 32.04 |
| n2c2 2014-Medication | English | Event Extraction | General EHR Note | Cardiology, Endocrinology | Medication information | Treatment and Intervention | 451 | 746.75 / 66.87 |
| CAS-label | French | Event Extraction | Case Report | General | Post-discharge patient management | Discharge and Administration | 696 | 365.62 / 6.10 |
| CAS-evidence | French | Summarization | Case Report | General | Summarization | Discharge and Administration | 696 | 365.62 / 33.19 |
| RuMedNLI | Russian | Natural Language Inference | General EHR Note | Critical Care | Semantic relation | Research | 1421 | 23.90 / 2.00 |
| RuDReC-NER | Russian | Named Entity Recognition | Consultation Record | Pharmacology | ADE & Incidents | Treatment and Intervention | 251 | 16.19 / 9.52 |
| NorSynthClinical-PHI | Norwegian | Named Entity Recognition | General EHR Note | Cardiology | De-identification | Research | 211 | 17.71 / 9.00 |
| RuCCoN | Russian | Named Entity Recognition | General EHR Note | Pulmonology | Procudure information | Treatment and Intervention | 846 | 210.19 / 135.85 |
| CLISTER | French | Semantic Similarity | Case Report | General | Semantic relation | Research | 400 | 32.73 / 3.00 |
| BRONCO150-NER&Status | German | Event Extraction | Discharge Summary | Oncology | Procudure information | Treatment and Intervention | 880 | 78.44 / 66.56 |
| CARDIO-DE | German | Named Entity Recognition | General EHR Note | Cardiology | Medication information | Treatment and Intervention | 380 | 1902.40 / 259.12 |
| GraSSCo PHI | German | Named Entity Recognition | Discharge Summary, Case Report | General | De-identification | Research | 329 | 101.38 / 21.10 |
| IFMIR-Incident type | Japanese | Text Classification | Case Report | Pharmacology | ADE & Incidents | Treatment and Intervention | 5833 | 98.41 / 4.93 |
| IFMIR-NER | Japanese | Named Entity Recognition | Case Report | Pharmacology | ADE & Incidents | Treatment and Intervention | 5747 | 98.52 / 42.94 |
| IFMIR-NER&factuality | Japanese | Event Extraction | Case Report | Pharmacology | ADE & Incidents | Treatment and Intervention | 5747 | 98.52 / 58.87 |
| iCorpus | Japanese | Named Entity Recognition | Case Report | General | Procudure information | Treatment and Intervention | 217 | 91.15 / 147.10 |
| icliniq-10k | English | Question Answering | Consultation Record | General | Screen & Consultation | Triage and Referral | 733 | 96.14 / 94.87 |
| HealthCareMagic-100k | English | Question Answering | Consultation Record | General | Screen & Consultation | Triage and Referral | 11199 | 90.33 / 117.97 |

## 6 BRIDGE Dataset and Task Information

This section introduces the dataset sources and corresponding task information included in the BRIDGE benchmark. For each dataset, we provide a concise summary describing its background and purpose, followed by a list of associated metadata attributes. Unless otherwise specified, the listed metadata apply to all tasks derived from the dataset. If certain tasks differ in specific metadata attributes, these differences are explicitly noted within the dataset's metadata section.

For each task, we include a brief description together with the prompt template used during model inference. All templates correspond to either zero-shot or few-shot prompting settings, as defined in our evaluation framework.

All datasets were transformed and standardized to make them easy for LLMs to understand and execute, including detailed instructions and structured input and output. In general, task construction followed the original dataset design or established derivations in related studies. Additionally, we have newly designed and constructed 13 tasks as part of the BRIDGE benchmark. For previously established tasks, readers are referred to the corresponding citations for details, whereas for newly developed ones, we explicitly describe the task design and construction process in each task description with the format: [New task construction setting]: <description>.

### 6.1 MIMIC-IV CDM

The MIMIC-IV CDM (Clinical Decision Making) dataset[19] is sourced from the MIMIC-IV database and focuses on patients presenting with acute abdominal pain. It includes 2,400 real patient cases covering four common abdominal pathologies and incorporates multiple types of clinical notes to simulate a realistic clinical setting, such as the history of present illness, physical examination findings, and radiology reports.

- **Language:** English
- **Clinical Stage:** Diagnosis and Prognosis
- **Sourced Clinical Document Type:** General EHR Note
- **Clinical Specialty:** Gastroenterology
- **Application Method:** Link of MIMIC-IV CDM Dataset

#### 6.1.1 Task: MIMIC-IV CDM

This task is to determine the most likely diagnosis based on the clinical notes of a patient with acute abdominal pain.

> **Task type:** *Text Classification*
> **Instruction:** *Given the history of present illness, physical examination, and radiology reports of a patient with acute abdominal pain, determine the most likely diagnosis from the following pathologies: Appendicitis, Cholecystitis, Diverticulitis, Pancreatitis.*
> *Return your answer in the following format. DO NOT GIVE ANY EXPLANATION:*
> *Diagnosis: disease*
> *The optional list for "disease" is ["Appendicitis", "Cholecystitis", "Diverticulitis", "Pancreatitis"].*
> **Input:** *[Clinical note of a patient]*
> **Output:** *Diagnosis: [Appendicitis / Cholecystitis / Diverticulitis / Pancreatitis]*

### 6.2 MIMIC-III Outcome

The MIMIC-III Outcome dataset[20] contains data from 42,808 patients sourced from the MIMIC-III database and combines patients' clinical notes with their clinical outcomes. Admission-related information, such as chief complaint and medical history, was extracted from discharge summaries to simulate a realistic scenario for prognosis prediction at the time of hospital admission. This dataset supports research on clinical outcome prediction and risk stratification. We followed the official data split (29,839 for training, 4,300 for validation, and 8,669 for testing), and further selected 1,000 cases from the test set for evaluation in our benchmark.

- **Language:** English

- **Clinical Stage:** Diagnosis and Prognosis
- **Sourced Clinical Document Type:** Discharge Summary
- **Clinical Specialty:** Critical Care
- **Application Method:** Link of Code MIMIC-III Outcome

#### 6.2.1 Task: MIMIC-III Outcome.LoS

This task is to predict the patient's length of stay during the current hospital admission based on information from the admission note.

> **Task type:** *Text Classification*
> **Instruction:** *Given a patient's basic information and admission notes, predict the patient's length of stay (LOS) in the hospital, which is the hospitalization duration required by the patient's condition. The LoS is grouped into four categories:*
> *- "A": Under 3 days*
> *- "B": 3 to 7 days*
> *- "C": 1 week to 2 weeks*
> *- "D": more than 2 weeks*
> *Return your answer in the following format. DO NOT GIVE ANY EXPLANATION:*
> *Length of Stay: label*
> *The optional list for "label" is ["A", "B", "C", "D"].*
> **Input:** *[Clinical note of a patient]*
> **Output:** *Length of Stay: [A / B / C / D]*

#### 6.2.2 Task: MIMIC-III Outcome.Mortality

This task is to predict the in-hospital mortality for the current admission based on the patient's admission note.

> **Task type:** *Text Classification*
> **Instruction:** *Given a patient's basic information and admission notes, predict the patient's in-hospital mortality, which means whether the patient will die during the current admission.*
> *Return your answer in the following format. DO NOT GIVE ANY EXPLANATION:*
> *In-Hospital Mortality: label*
> *The optional list for "label" is ["Yes", "No"].*
> **Input:** *[Clinical note of a patient]*
> **Output:** *In-Hospital Mortality: [Yes/No]*

### 6.3 MIMIC-IV BHC

The MIMIC-IV BHC (Brief Hospital Course) dataset[21] focuses on generating brief hospital course summaries from patients' clinical notes. It is derived from the MIMIC-IV-Note dataset and includes 270,033 clinical notes. The dataset was constructed through data translation to transforms raw, unstructured clinical text into a standardized format suitable for brief hospital course generation. This process involves steps such as whitespace removal, section identification, tokenization, and other structural transformations, ultimately producing aligned pairs of clinical notes and corresponding brief hospital course summaries.

- **Language:** English
- **Clinical Stage:** Discharge and Administration
- **Sourced Clinical Document Type:** General EHR Note
- **Clinical Specialty:** Critical Care
- **Application Method:** Link of MIMIC-IV BHC Dataset

#### 6.3.1 Task: MIMIC-IV BHC

The objective of this task is to generate the brief hospital course based on the provided clinical notes of a patient.

> **Task type:** *Summarization*
> **Instruction:** *Given the patient's clinical notes from electronic health records, summarize the patient's clinical notes into a brief hospital course (BHC) summary, which is a concise summary of the patient's hospital stay. The BHC summary commonly is a paragraph that includes key information such as the patient's diagnosis, treatment, and any significant events that occurred during their hospital stay.*
> *Return your answer in the following format. DO NOT GIVE ANY EXPLANATION:*
> *BHC: ...*
> **Input:** *[Progress note of a patient]*
> **Output:** *BHC: [brief hospital course for the patient]*

### 6.4 MIMIC-IV DiReCT

The MIMIC-IV DiReCT (Diagnostic Reasoning dataset for Clinical noTes) dataset[22] focuses on disease diagnosis through diagnostic reasoning based on clinical notes. It is derived from the MIMIC-IV database and contains 511 clinical notes, each meticulously annotated by physicians to document the step-by-step diagnostic reasoning process—from clinical observations to final diagnosis. It covers 25 diseases across five clinical specialties: Cardiology, Gastroenterology, Neurology, Pulmonology, and Endocrinology.

- **Language:** English
- **Clinical Stage:** Diagnosis and Prognosis
- **Sourced Clinical Document Type:** General EHR Note
- **Clinical Specialty:** Cardiology, Gastroenterology, Neurology, Pulmonology, Endocrinology
- **Application Method:** Link of MIMIC-IV DiReCT Dataset

#### 6.4.1 Task: MIMIC-IV DiReCT.Dis

This task is to determine which disease the patient has based on their current clinical condition.

> **Task type:** *Text Classification*
> **Instruction:** *Given a patient's clinical notes from electronic health records (EHR), determine which disease the patient has. Specifically, the clinical notes include 6 parts: CHIEF COMPLAINT, HISTORY OF PRESENT ILLNESS, PAST MEDICAL HISTORY, FAMILY HISTORY, PHYSICAL EXAM, and PERTINENT RESULTS. The possible disease is one of the following list.*
> *Return your answer in the following format. DO NOT GIVE ANY EXPLANATION:*
> *Diagnosis: disease*
> *The optional list for "disease" is ["Acute Coronary Syndrome", "Heart Failure", "Gastro-oesophageal Reflux Disease", "Pulmonary Embolism", "Hypertension", "Peptic Ulcer Disease", "Stroke", "Gastritis", "Multiple Sclerosis", "Adrenal Insufficiency", "Pneumonia", "Chronic Obstructive Pulmonary Disease", "Aortic Dissection", "Asthma", "Diabetes", "Pituitary Disease", "Alzheimer", "Atrial Fibrillation", "Thyroid Disease", "Cardiomyopathy", "Epilepsy", "Upper Gastrointestinal Bleeding", "Tuberculosis", "Migraine", "Hyperlipidemia"].*
> **Input:** *[Clinical note of a patient]*
> **Output:** *Diagnosis: [valid disease name]*

#### 6.4.2 Task: MIMIC-IV DiReCT.PDD

This task is to determine the patient's primary discharge diagnosis (PDD) based on their current clinical condition.

> **Task type:** *Text Classification*
> **Instruction:** *Given a patient's clinical notes from electronic health records (EHR), determine which disease the patient has. Specifically, the clinical notes include 6 parts: CHIEF COMPLAINT, HISTORY OF PRESENT ILLNESS, PAST MEDICAL HISTORY, FAMILY HISTORY, PHYSICAL EXAM, and PERTINENT RESULTS. The possible disease is one of the following list.*
> *Return your answer in the following format. DO NOT GIVE ANY EXPLANATION:*
> *Diagnosis: disease*
> *The optional list for "disease" is ["Heart Failure", "Gastro-oesophageal Reflux Disease", "Hypertension", "Non-ST-Elevation Myocardial Infarction", "Relapsing-Remitting Multiple Sclerosis", "Unstable Angin", "Low-risk Pulmonary Embolism", "Gastric Ulcers", "Chronic Obstructive Pulmonary Disease", "Bacterial Pneumonia", "ST-Elevation Myocardial Infarction", "Hemorrhagic Stroke", "Acute Gastritis", "Ischemic Stroke", "Submassive Pulmonary Embolism", "Pituitary Macroadenoma", "Secondary Adrenal Insufficiency", "Alzheimer", "Type B Aortic Dissection", "Duodenal Ulcers", "Chronic Atrophic Gastritis", "Paroxysmal Atrial Fibrillation", "Primary Adrenal Insufficiency", "Upper Gastrointestinal Bleeding", "Type I Diabetes", "Type II Diabetes", "Chronic Non-atrophic Gastritis", "Type A Aortic Dissection", "Non-epileptic Seizure", "Tuberculosis", "Viral Pneumonia", "Severe Asthma Exacerbation", "Hyperthyroidism", "Dilated Cardiomyopathy", "Congenital Adrenal Hyperplasia", "Epilepsy", "Non-Allergic Asthma", "Migraine With Aura", "Secondary Progressive Multiple Sclerosis", "Hypothyroidism", "Massive Pulmonary Embolism", "Hyperlipidemia", "Restrictive Cardiomyopathy", "Asthma", "COPD Asthma", "Hypertrophic Cardiomyopathy", "Persistent Atrial Fibrillation", "Thyroid Nodules", "Primary Progressive Multiple Sclerosis", "Allergic Asthma", "Cough-Variant Asthma", "Arrhythmogenic Right Ventricular Cardiomyopathy", "Migraine Without Aura", "Thyroiditis", "Pituitary Microadenoma"].*
> **Input:** *[Clinical note of a patient]*
> **Output:** *[Diagnosis: [valid disease name]]*

### 6.5 ADE

The ADE dataset[23] was constructed from 3,000 English clinical case reports focused on adverse drug effects. Through multi-round systematic annotation, the dataset captures mentions of drugs, adverse effects, dosages, and the relationships among them. All entities and relations were annotated with rigorous quality control to ensure the dataset's reliability for medication information extraction research. This dataset supports three distinct tasks.

- **Language:** English
- **Clinical Stage:** Treatment and Intervention
- **Sourced Clinical Document Type:** Case Report
- **Clinical Specialty:** Pharmacology
- **Application Method:** Link of ADE Dataset

#### 6.5.1 Task: ADE-Identification

This task is to determine whether a sentence contains information about ADEs.

> **Task type:** *Text Classification*
> **Instruction:** *Given the clinical text, determine whether the text mentions adverse drug effects.*
> *Return your answer in the following format. DO NOT GIVE ANY EXPLANATION:*
> *adverse drug effect: label*

> *The optional list for "label" is ["Yes", "No"].*
> **Input:** *[A snippet from a case report]*
> **Output:** *adverse drug effect: [Yes / No]*

#### 6.5.2 Task: ADE-Extraction

This task is to extract all potential drugs and their associated adverse effects mentioned in the sentence.

> **Task type:** *Event Extraction*
> **Instruction:** *Given the clinical text, identify all the drugs and their corresponding adverse effect mentioned in the text.*
> *Return your answer in the following format. DO NOT GIVE ANY EXPLANATION:*
> *drug: ..., adverse effect: ...;*
> *...*
> *drug: ..., adverse effect: ...;*
> **Input:** *[A snippet from a case report. ]*
> **Output:** *drug: [drug name], adverse effect: [text of ADE information];*

#### 6.5.3 Task: ADE-Drug dosage

This task is to extract all potential drugs and their associated adverse effects mentioned in the sentence.

> **Task type:** *Event Extraction*
> **Instruction:** *Given the clinical text, identify all the drugs and their corresponding dosage information mentioned in the text.*
> *Return your answer in the following format. DO NOT GIVE ANY EXPLANATION:*
> *drug: ..., dosage: ...;*
> *...*
> *drug: ..., dosage: ...;*
> **Input:** *[A snippet from a case report]*
> **Output:** *drug: [drug name], dosage: [text of dosage information];*

### 6.6 BARR2

The BARR2 (Biomedical Abbreviation Recognition and Resolution, 2nd Edition) dataset[24] is a Spanish dataset designed for the recognition and resolution of biomedical abbreviations. It includes 24.5 million tokens of Spanish clinical case study sections compiled from various clinical disciplines, and forms part of a larger 1.1 billion-token biomedical corpus created by the Barcelona Supercomputing Center (BSC). The dataset comprises clinical case reports extracted from Spanish medical literature, annotated and structured for training and evaluation of biomedical language models. The annotations in derived tasks like PharmaCoNER and CANTEMIST were conducted using curated guidelines for biomedical and oncological entities.

- **Language:** Spanish
- **Clinical Stage:** Research
- **Sourced Clinical Document Type:** Case Report
- **Clinical Specialty:** General
- **Application Method:** Link of BARR2 Dataset

#### 6.6.1 Task: BARR2

This task is to extract the clinical abbreviations from the text and resolve each of them with their definitions.

> **Task type:** *Event Extraction*
> **Instruction:** *Given the clinical text of a patient in Spanish, extract the clinical abbreviations from the text and resolve each of them with their definitions. Return your answer in the following format. DO NOT GIVE ANY EXPLANATION:*
> *abbreviation: ..., definition: ...;*
> *...*
> *abbreviation: ..., definition: ...;*
> **Input:** *[A snippet from a case report]*
> **Output:** *abbreviation: [clinical abbreviation], definition: [resolved clinical abbreviation];*

### 6.7 BrainMRI-AIS

BrainMRI-AIS[25] dataset comprises 3,024 brain MRI radiology reports, focusing on the identification of acute ischemic stroke (AIS). It consists of free-text radiology reports collected between January 2015 and December 2016 from Hallym University Sacred Heart Hospital in Korea. Reports were manually annotated with AIS or non-AIS labels by medical experts to ensure high-quality labeling.

- **Language:** English
- **Clinical Stage:** Diagnosis and Prognosis
- **Sourced Clinical Document Type:** Radiology Report
- **Clinical Specialty:** Neurology, Radiology
- **Application Method:** Link of BrainMRI-AIS Dataset

#### 6.7.1 Task: BrainMRI-AIS

This task is to determine if the patient has an acute ischemic stroke.

> **Task type:** *Text Classification*
> **Instruction:** *Given a brain Magnetic Resonance Imaging (MRI) radiology report, determine whether the patient has acute ischemic stroke (AIS).*
> *Return your answer in the following format. DO NOT GIVE ANY EXPLANATION:*
> *AIS: label*
> *The optional list for "label" is ["Yes", "No"].*
> **Input:** *[A brain magnetic resonance imaging (MRI) radiology report of a patient]*
> **Output:** *AIS: [Yes / No]*

### 6.8 Brateca

Brateca[26] is a Portuguese-language dataset consisting of over 70,000 admissions and 2.5 million clinical notes and structured documents, including health records, prescription data, and exam results. The dataset was collected from 10 hospitals across two Brazilian states. Documents were annotated and deidentified with BiLSTM-CRF models and manually reviewed by clinical researchers, following HIPAA guidelines. The following tasks were newly constructed based on this dataset as part of our BRIDGE benchmark.

- **Language:** Portuguese (Brazilian)
- **Clinical Stage:** Diagnosis and Prognosis
- **Sourced Clinical Document Type:** General EHR Note

- **Clinical Specialty:** General
- **Application Method:** Link of Brateca Dataset

#### 6.8.1 Task: Brateca-Hospitalization

This task is to predict whether the patient will require more than seven days of hospitalization.
[New task construction setting]: For each patient, we extracted all clinical notes recorded during hospital visits and selected the admission-day notes, including both examination results and admission summaries. The input features include age, skin color, sex, weight, and height, while the output label indicates whether the patient's length of stay during the current hospitalization exceeded seven days.

> **Task type:** *Text Classification*
> **Instruction:** *Given a patient's basic information and clinical notes in Portuguese, predict whether the patient will require more than seven days of hospitalization.*
> *Return your answer in the following format. DO NOT GIVE ANY EXPLANATION:*
> *Hospitalization > 7 days: label*
> *The optional list for "label" is ["Yes", "No"].*
> **Input:** *[Clinical notes of a patient]*
> **Output:** *Hospitalization > 7 days: [Yes / No]*

#### 6.8.2 Task: Brateca-Mortality

This task is to predict whether the clinical outcome for the patient is survival or death.
[New task construction setting]: For every patient, we collected all notes corresponding to the day of admission (including examination results and admission notes), together with basic demographic variables including age, skin color, sex, weight, and height. The target variable represents the clinical outcome, categorized as either survival or death during hospitalization.

> **Task type:** *Text Classification*
> **Instruction:** *Given a patient's basic information and clinical notes in Portuguese, predict whether the clinical outcome for this patient is survival or death.*
> *Return your answer in the following format. DO NOT GIVE ANY EXPLANATION:*
> *Survival: status*
> *The optional list for "status" is ["Yes", "No"].*
> **Input:** *[Clinical notes of a patient]*
> **Output:** *Survival: [Yes / No]*

### 6.9 Cantemist

The Cantemist corpus[27] comprises 3,000 clinical cases centered on cancer-related data in Spanish. This corpus was curated by human clinical coding experts, who manually annotated tumor morphology entities and mapped them to the Spanish version of the International Classification of Diseases for Oncology (ICD-O). The annotation process was conducted using the BRAT annotation tool, adhering to well-defined annotation guidelines established by the Spanish Ministry of Health.

- **Language:** Spanish
- **Clinical Stage:** Discharge and Administration
- **Sourced Clinical Document Type:** Case Report
- **Clinical Specialty:** Oncology
- **Application Method:** Link of Cantemist Dataset

#### 6.9.1 Task: Cantemist-Coding

This task is to generate a ranked list of all morphology codes for tumor morphologies mentioned in the text according to CIE-O (Clasificación Internacional de Enfermedades para Oncología), the Spanish adaptation of the International Classification of Diseases for Oncology (ICD-O).

> **Task type:** *Normalization and Coding*
> **Instruction:** *Given the clinical document in Spanish, identify the entities of tumor morphology and determine the corresponding morphology codes. Specifically, the entities of tumor morphology can be linked to a morphology code from CIE-O (Clasificación Internacional de Enfermedades para Oncología, i.e., the Spanish equivalent of ICD-O, version 3.1).*
> *- "Tumor morphology": Una neoplasia es un crecimiento o formación de tejido nuevo, anormal, especialmente de carácter tumoral, benigno o maligno. La clasificación de las neoplasias según su morfología o características histológicas hace referencia a la forma y estructura de las células tumorales que se estudian para clasificar las neoplasias según su tejido de origen. El tejido de origen y el tipo de células que componen una morfología determinan a menudo la tasa de crecimiento esperada, la gravedad de la enfermedad y el tipo de tratamiento recomendado.*
> *- "Morphology code": The morphology code is a code from CIE-O that represents the morphology of the tumor. This code is used to classify the tumor morphology in a standardized way. A code consists of a four-digit morphology code indicating the tumor's histological type, followed by a slash and a single digit indicating the tumor's behavior.*
> *Assuming the number of normalized morphology codes is N, return the N normalized codes in the output.*
> *Return your answer in the following format. DO NOT GIVE ANY EXPLANATION:*
> *Morphology code: code 1, code 2, ..., code N*
> *The optional list for "code" is the normalized code from CIE-O.*
> **Input:** *[Clinical case of cancer patient in Spanish]*
> **Output:** *Morphology code: [N normalized codes]*

#### 6.9.2 Task: Cantemis-NER

This task is to extract all entities of tumor morphology mentioned in the text.

> **Task type:** *Named Entity Recognition*
> **Instruction:** *Given the clinical text related to cancer in Spanish, extract all entities about tumor morphology mentioned in the text.*
> *- "Tumor morphology": Una neoplasia es un crecimiento o formación de tejido nuevo, anormal, especialmente de carácter tumoral, benigno o maligno. La clasificación de las neoplasias según su morfología o características histológicas hace referencia a la forma y estructura de las células tumorales que se estudian para clasificar las neoplasias según su tejido de origen. El tejido de origen y el tipo de células que componen una morfología determinan a menudo la tasa de crecimiento esperada, la gravedad de la enfermedad y el tipo de tratamiento recomendado.*
> *Return your answer in the following format. DO NOT GIVE ANY EXPLANATION:*
> *entity: ..., type: tumor morphology;*
> *...*
> *entity: ..., type: tumor morphology;*
> **Input:** *[Clinical case of cancer patient in Spanish]*
> **Output:** *entity: [tumor morphology mention], type: tumor morphology;*

#### *6.9.3 Task: Cantemis-Norm*

This task is to extract all entities about tumor morphology mentioned in the text and identify their corresponding morphology codes according to CIE-O (Clasificación Internacional de Enfermedades para Oncología), the Spanish adaptation of the International Classification of Diseases for Oncology (ICD-O).

> **Task type:** *Normalization and Coding*
> **Instruction:** *Given the clinical text related to cancer in Spanish, extract all entities about tumor morphology and identify their corresponding morphology codes. Specifically, every entity of tumor morphology is linked to a morphology code from CIE-O (Clasificación Internacional de Enfermedades para Oncología, i.e., the Spanish equivalent of ICD-O, version 3.1).*
> *- "Tumor morphology": Una neoplasia es un crecimiento o formación de tejido nuevo, anormal, especialmente de carácter tumoral, benigno o maligno. La clasificación de las neoplasias según su morfología o características histológicas hace referencia a la forma y estructura de las células tumorales que se estudian para clasificar las neoplasias según su tejido de origen. El tejido de origen y el tipo de células que componen una morfología determinan a menudo la tasa de crecimiento esperada, la gravedad de la enfermedad y el tipo de tratamiento recomendado.*
> *- "Morphology code": The morphology code is a code from CIE-O that represents the morphology of the tumor. This code is used to classify the tumor morphology in a standardized way. A code consists of a four-digit morphology code indicating the tumor's histological type, followed by a slash and a single digit indicating the tumor's behavior. Return your answer in the following format. DO NOT GIVE ANY EXPLANATION:*
> *entity: ..., code: ...;*
> *...*
> *entity: ..., code: ...;*
> *The optional list for "code" is the normalized code from CIE-O.*
> **Input:** *[Clinical case of cancer patient in Spanish]*
> **Output:** *entity: [tumor morphology mention], code: [CIE-O code];*

### 6.10 CARES

The CARES[28] is a Spanish-language dataset including 3,219 anonymized radiological reports and 6,907 sub-code annotations. It was designed for the ICD-10 classification of free-text radiological reports. Each report was labeled with one or more ICD-10 sub-codes from a total of 223 unique sub-codes, which correspond to 156 distinct ICD-10 codes and 16 different chapters of the ontology. Annotations were carried out by a team of four expert radiologists with over 10 years of experience, following the official ICD-10 coding standards.

- **Language:** Spanish
- **Clinical Stage:** Discharge and Administration
- **Sourced Clinical Document Type:** Radiology Report
- **Clinical Specialty:** Radiology
- **Application Method:** Link of CARES Dataset

#### *6.10.1 Task: CARES-Area*

This task is to determine which anatomical area of the body that a radiographic report refers to.
[New task construction setting]: Based on the report source and type information, we link the raw report to its corresponding anatomical region as described in the report, which demonstrates the clinical focus of each report.

**Task type:** *Text Classification*
**Instruction:** *Given a radiology report in Spanish, determine which anatomical area of the body the report refers to.*
*Return your answer in the following format. DO NOT GIVE ANY EXPLANATION:*
*anatomical area: label*
*The optional list for "label" is ['Columna', 'Neuro', 'Musculoskeletal', 'Body'].*
**Input:** *[A radiology report in Spanish]*
**Output:** *anatomical area: [Columna / Neuro / Musculoskeletal / Body]*

#### 6.10.2 Task: CARES-ICD10 Chapter

This task is to identify the appropriate chapters of ICD-10 that correspond to the condition mentioned in the radiology report.
[Task Setting]: Following the design of the original study, each report was mapped to one or more ICD-10 chapters, as multiple conditions could be mentioned in a single report.

**Task type:** *Normalization and Coding*
**Instruction:** *Given a radiology report in Spanish, determine the appropriate chapters of ICD-10 corresponding to the conditions mentioned in the report. Specifically, the chapter is the highest level of the ICD-10 classification and each chapter represents a group of related diseases. Each chapter is identified by a Roman numeral from I to XXII, including the following chapters:*
*- "I": Certain infectious and parasitic diseases (A00–B99)*
*- "II": Neoplasms (C00–D49)*
*- "III": Diseases of the blood and blood-forming organs and certain disorders involving the immune mechanism (D50–D89)*
*- "IV": Endocrine, nutritional, and metabolic diseases (E00–E89)*
*- "V": Mental, Behavioral and Neurodevelopmental disorders (F01–F99)*
*- "VI": Diseases of the nervous system (G00–G99)*
*- "VII": Diseases of the eye and adnexa (H00–H59)*
*- "VIII": Diseases of the ear and mastoid process (H60–H95)*
*- "IX": Diseases of the circulatory system (I00–I99)*
*- "X": Diseases of the respiratory system (J00–J99)*
*- "XI": Diseases of the digestive system (K00–K95)*
*- "XII": Diseases of the skin and subcutaneous tissue (L00–L99)*
*- "XIII": Diseases of the musculoskeletal system and connective tissue (M00–M99)*
*- "XIV": Diseases of the genitourinary system (N00–N99)*
*- "XV": Pregnancy, childbirth, and the puerperium (O00–O9A)*
*- "XVI": Certain conditions originating in the perinatal period (P00–P96)*
*- "XVII": Congenital malformations, deformations, and chromosomal abnormalities (Q00–Q99)*
*- "XVIII": Symptoms, signs, and abnormal clinical and laboratory findings, not elsewhere classified (R00–R99)*
*- "XIX": Injury, poisoning, and certain other consequences of external causes (S00–T88)*
*- "XX": External causes of morbidity (V00–Y99)*
*- "XXI": Factors influencing health status and contact with health services (Z00–Z99)*
*- "XXII": Codes for special purposes (U00–U85)*
*This report may contain multiple conditions and is related to multiple chapters. Assuming the number of appropriate chapters is N, return the codes of N appropriate chapters in the output.*
*Return your answer in the following format. DO NOT GIVE ANY EXPLANATION:*
*ICD-10 chapter: code 1, code 2, ..., code N*

*The optional list for "code" is ["I", "II", "III", "IV", "V", "VI", "VII", "VIII", "IX", "X", "XI", "XII", "XIII", "XIV", "XV", "XVI", "XVII", "XVIII", "XIX", "XX", "XXI", "XXII"].*
**Input:** *[A radiology report in Spanish]*
**Output:** *ICD-10 chapter: chapter 1, chapter 2, ..., chapter N (with each code is from ICDO-10 sub block code)*

#### 6.10.3 Task: CARES ICD10 Block

This task is to identify the appropriate block code of ICD-10 that corresponds to the condition mentioned in the radiology report.
[New task construction setting]: Each report was linked to one or more ICD-10 block codes, capturing the best matched to the diagnostic descriptions contained in each report.

**Task type:** *Normalization and Coding*
**Instruction:** *Given a radiology report in Spanish, determine the appropriate ICD-10 block codes corresponding to the conditions mentioned in the report. Specifically, the block code is the second level of the ICD-10 classification and represents a group of related diseases. Each block code is identified by a code containing a character and two digits, which indicates its chapter and detailed block.*
*This report may contain multiple conditions and is related to multiple block codes. Assuming the number of appropriate block codes is N, return the codes for the N appropriate blocks in the output. Notably, the required block code is a combination of the chapter and the block, such as "I00", "I01", rather than the coarse range of the block, such as "I00-I99".*
*Return your answer in the following format. DO NOT GIVE ANY EXPLANATION:*
*ICD-10 block: code 1, code 2, ..., code N*
**Input:** *[A radiology report in Spanish]*
**Output:** *[ICD-10 block: code 1, code 2, ..., code N (each code is from ICDO-10 block code)]*

#### 6.10.4 Task: CARES-ICD10 Sub-Block

This task is to identify the appropriate subblock code of ICD-10 that corresponds to the condition mentioned in the radiology report.
[New task construction setting]: Each report was linked to one or more ICD-10 sub-block codes, capturing finer-grained diagnostic distinctions within the broader ICD-10 block categories.

**Task type:** *Normalization and Coding*
**Instruction:** *Given a radiology report in Spanish, determine the appropriate ICD-10 sub-block codes corresponding to the conditions mentioned in the report. Specifically, the sub-block code is the third level of the ICD-10 classification and represents several related diseases. Each sub-block code is identified by a code containing a character, two digits, and a decimal, which indicates its chapter, block, and detailed sub-block.*
*This report may contain multiple conditions and is related to multiple sub-block codes. Assuming the number of appropriate sub-blocks is N, return the codes for N appropriate sub-blocks in the output. Notably, the required sub-block code is a combination of the chapter, the block, and the sub-block, such as "I00.0", "I01.0", rather than the coarse range of the sub-block, such as "I00.0-I99.9".*
*Return your answer in the following format. DO NOT GIVE ANY EXPLANATION:*
*ICD-10 sub-block: code 1, code_2, ..., code N.*
**Input:** *[A radiology report in Spanish]*
**Output:** *ICD-10 sub-block: code 1, code 2, ..., code N. (with each code is from ICDO-10 chapter code)*

### 6.11 CHIP-CDEE

The CHIP-CDEE (CHIP- Clinical Discovery Event Extraction dataset) dataset[29] consists of 1,971 clinical text sourced from disease history and imaging reports in Chinese EHRs. It focuses on the extraction of clinical findings, referring broadly to patient-reported symptoms and examination-derived abnormalities, including signs and manifestations. The dataset requires identifying four attributes for each clinical finding: the subject term, descriptive term, anatomical location, and occurrence status. It was one of the shared tasks in the CHIP-2021 challenge and is included in the CBLUE benchmark[30]. We adapted the CBLUE version for downstream task construction.

- **Language:** Chinese
- **Clinical Stage:** Initial Assessment
- **Sourced Clinical Document Type:** General EHR Note
- **Clinical Specialty:** General
- **Application Method:** Link of CHIP-CDEE Dataset

#### 6.11.1 Task: CHIP-CDEE

This task is to extract clinical findings and their associated attributes: description, location, and status.

> **Task type:** *Event Extraction*
> **Instruction:** *Given the clinical text from electronic healthcare records in Chinese, extract all the clinical findings and their attributes.* 具体而言，给定一段现病史或者医学影像所见报告，要求从中抽取临床发现事件的四个属性*:* 解剖部位、主体词、描述词，以及发生状态*:*
> *1.* 主体词*(subject)*：指患者的电子病历中的疾病名称或者由疾病引发的症状，也包括患者的一般情况如饮食，二便，睡眠等。主体词尽可能完整并是专有名词，比如"麻木，疼痛，发烧，囊肿"等；专有名词，如"头晕"，晕只能发生在头部，"胸闷"，闷只能发生在胸部，所以不进行拆分，保留完整的专有名词。涉及泛化的症状不做标注，如"无其他不适"，句子中的"不适"不需要标注，只针对具体的进行标注。注意：有较小比例的主体词会映射到*ICD*标准术语，所使用的*ICD*的版本为"国际疾病分类 *ICD-10*北京临床版*v601.xIsx*"
> *2.* 描述词*(description)*：对主体词的发生时序特征、轻重程度、形态颜色等多个维度的刻画，也包括疾病的起病缓急、突发 *3.* 解剖部位*(location)*：指主体词发生在患者的身体部位，也包括组织，细胞，系统等，也包括部位的方向和数量
> *4.* 发生状态*(status)*："肯定"，"否定"或"不确定"，表示该主体词在患者的电子病历中的存在状态
> *Return your answer in the following format. DO NOT GIVE ANY EXPLANATION:*
> *subject: ..., description: [..., ...], location: [..., ...], status: ...;*
> *...*
> *subject: ..., description: [..., ...], location: [..., ...], status: ...;*
> *The optional list for "status" is ["*肯定*", "*否定*", "*不确定*"]; If there is not a description or location, they should be "[]"; if there are multiple descriptions or locations, they should be separated by commas in the list, like [..., ...].*
> **Input:** *[Clinical text from EHRs]*
> **Output:** *subject: [subject terms], description: [descriptive words], location: [location mention], status: [*肯定*/*否定*/*不确定*];*

### 6.12 C-EMRS

C-EMRs[31] is a collection of 18,590 Chinese electronic medical records, designed for automated clinical diagnosis. It was collected from Huangshi Central Hospital, consisting of a variety of medical cases such as hypertension, diabetes, COPD, etc. across departments. Records in the dataset include fields such as chief complaint, physical examination, and labels such as diseases. Annotations were obtained through expert diagnosis and manually aligned for classification tasks.

- **Language:** Chinese
- **Clinical Stage:** Diagnosis and Prognosis
- **Sourced Clinical Document Type:** General EHR Note
- **Clinical Specialty:** Radiology, Endocrinology, Pulmonology, Cardiology, Gastroenterology
- **Application Method:** Link of C-EMRS Dataset

#### 6.12.1 Task: C-EMRS

This task is to diagnose the disease this patient has.
[New task construction setting]: Based on the structured EHRs provided in the dataset, we constructed a comprehensive input by aggregating multiple categories of medical information, including the chief complaint, surgical history, vital signs, specialty findings, general condition, allergy history, nutritional status, suicidal tendency, specialty examinations, surgical or trauma history, comorbidities, present illness, obstetric history, auxiliary examinations, personal history, past medical history, and family history. These elements were integrated and reformulated into a standardized clinical case report. The output label represents the physician-diagnosed disease.

> **Task type:** *Text Classification*
> **Instruction:** *Given a clinical electronic health record in Chinese, diagnose the disease this patient has. Return your answer in the following format. DO NOT GIVE ANY EXPLANATION:*
> *diagnosis: disease*
> *The optional list for "disease" is ["胃息肉", "泌尿道感染", "慢性阻塞性肺病", "痛风", "胃溃疡", "高血压", "哮喘", "胃炎", "心律失常", "糖尿病"].*
> **Input:** *[Clinical electronic health record in Chinese]*
> **Output:** *diagnosis: [胃息肉 / 泌尿道感染 / 慢性阻塞性肺病 / 痛风 / 胃溃疡 / 高血压 / 哮喘 / 胃炎 / 心律失常 / 糖尿病];*

### 6.13 CLEF eHealth 2020 - CodiEsp

The CodiEsp dataset[32] is a Spanish dataset designed for automatic clinical coding, particularly the assignment of ICD-10-CM (diagnoses) and ICD-10-PCS (procedures) codes to medical documents. It includes 1,000 manually annotated clinical case reports comprising over 18435 annotations and 3427 unique ICD-10 codes, curated by professional clinical coders from the Barcelona Supercomputing Center. These documents span diverse medical specialties and contain both diagnostic and procedural information. The dataset consists of plain text clinical narratives, with annotations including ICD-10 codes and the corresponding textual evidence justifying each code. Annotations were conducted in accordance with the 2018 Spanish ICD-10 coding manuals, using an iterative validation process to ensure annotation quality.

- **Language:** Spanish
- **Clinical Stage:** Discharge and Administration
- **Sourced Clinical Document Type:** Case Report
- **Clinical Specialty:** General
- **Application Method:** Link of CLEF eHealth 2020 - CodiEsp Dataset

#### 6.13.1 Task: CodiEsp-ICD-10-CM

This task is to extract clinical diagnosis from the clinical records and convert each of them into ICD-10-CM codes.

**Task type:** *Normalization and Coding*
**Instruction:** *Given the clinical text of a patient in Spanish, extract the clinical diagnosis from the clinical records and convert each of them into ICD-10-CM codes.*
*Return your answer in the following format. DO NOT GIVE ANY EXPLANATION:*
*diagnosis: ..., ICD-10-CM: ...;*
*...*
*diagnosis: ..., ICD-10-CM: ...;*
**Input:** *[Clinical text of a patient]*
**Output:** *diagnosis: [clinical diagnosis], ICD-10-CM: [ICD-10-CM code];*

#### 6.13.2 Task: CodiEsp-ICD-10-PCS

This task is to extract clinical procedures from the clinical records and convert each of them into ICD-10-PCS codes.

**Task type:** *Normalization and Coding*
**Instruction:** *Given the clinical text of a patient in Spanish, extract clinical procedures from the clinical records and convert each of them into ICD-10-PCS codes.*
*Return your answer in the following format. DO NOT GIVE ANY EXPLANATION:*
*procedure: ..., ICD-10-PCS: ...;*
*...*
*procedure: ..., ICD-10-PCS: ...;*
**Input:** *[Clinical text of a patient]*
**Output:** *procedure: [clinical procedure], ICD-10-PCS: [ICD-10-PCS code];*

### 6.14 ClinicalNotes-UPMC

ClinicalNotes-UPMC[33] is a dataset comprising 2376 clinical phrases and sentences, collected from the University of Pittsburgh Medical Center. It was extracted from 120 reports of 6 types (emergency department, discharge summaries, surgical pathology, radiology, operative notes, echocardiograms). Each sentence was labeled by physicians, indicating whether the context is negated or affirmed.

- **Language:** English
- **Clinical Stage:** Research
- **Sourced Clinical Document Type:** General EHR Note
- **Clinical Specialty:** General
- **Application Method:** Link of ClinicalNotes-UPMC Dataset

#### 6.14.1 Task: ClinicalNotes-UPMC

This task is to determine whether the provided concept is Affirmed or Negated.

**Task type:** *Text Classification*
**Instruction:** *Given a sentence from clinical notes, determine whether the provided concept is Affirmed or Negated. Specifically, if the concept is mentioned in the sentence and it is negated, then the label is "No". If the concept is mentioned in the sentence and it is affirmed, then the label is "Yes".*
*Return your answer in the following format. DO NOT GIVE ANY EXPLANATION:*
*affirmed: label*
*The optional list for "label" is ["Yes", "No"].*
**Input:** *[Sentence in clinical notes]*

> **Output:** *affirmed: [Yes / No]*

### 6.15 Mexican Clinical Records

The Mexican Clinical Records dataset[34] is a Spanish dataset designed for automatic classification of pneumonia and pulmonary thromboembolism (PTE). It includes 173 clinical records compiled from the Mexican Social Security Institute (IMSS). The dataset contains electronic health records, including structured data (e.g., laboratory studies, vital signs) and unstructured data (e.g., clinical notes and discharge summaries), covering diagnostic challenges in respiratory conditions. Annotations were based on ICD-10 classifications, and data preprocessing included regular expression extraction and relational database modeling.

- **Language:** Spanish
- **Clinical Stage:** Diagnosis and Prognosis
- **Sourced Clinical Document Type:** General EHR Note
- **Clinical Specialty:** Pulmonology
- **Application Method:** Link of Mexican Clinical Records Dataset

#### 6.15.1 Task: PPTS

This task is to determine if the patient has pneumonia or pulmonary thromboembolism.

> **Task type:** *Text Classification*
> **Instruction:** *Given the clinical text of a patient in Spanish, label the patient into one of the following:*
> *- "pneumonia": The patient has pneumonia.*
> *- "thromboembolism": The patient has pulmonary thromboembolism.*
> *- "control": The patient has neither pneumonia nor pulmonary thromboembolism.*
> *Return your answer in the following format. DO NOT GIVE ANY EXPLANATION:*
> *answer: label*
> *The optional list for "label" is ["pneumonia", "thromboembolism", "control"].*
> **Input:** *[Clinical text of a patient]*
> **Output:** *answer: [pneumonia / thromboembolism / control]*

### 6.16 CLINpt

The CLINpt dataset[35] is a Portuguese dataset designed for named entity recognition in clinical text. The dataset is collected from the Sinapse journal and the Neurology service of the Coimbra University Hospital Centre. The dataset contains clinical narratives such as admission notes, diagnostic test reports, and discharge letters, focusing on neurology. Annotations were performed manually using the IOB format and revised by biomedical and data science experts, adhering to guidelines developed with physicians and linguists.

- **Language:** Portuguese
- **Clinical Stage:** Treatment and Intervention
- **Sourced Clinical Document Type:** Case Report, Admission Note, Discharge Summary
- **Clinical Specialty:** Neurology
- **Application Method:** Link of CLINpt Dataset

#### 6.16.1 Task: CLINpt-NER

This task is to extract the following types of entities from the clinical records: "Characterization", "Test", "Evolution", "Genetics", "Anatomical Site", "Negation", "Additional Observations", "Condition", "Results", "DateTime", "Therapeutics", "Value", "Route of Administration".

**Task type:** *Named Entity Recognition*
**Instruction:** *Given the clinical text of a patient in Portuguese, extract the following types of entities from the clinical text: "characterization", "test", "evolution", "genetics", "anatomical Site", "negation", "additional observations", "condition", "results", "dateTime", "therapeutics", "value", "route of administration".*
*Return your answer in the following format. DO NOT GIVE ANY EXPLANATION:*
*entity: ..., type: ...;*
*...*
*entity: ..., type: ...;*
*The optional list for "type" is ["characterization", "test", "evolution", "genetics", "anatomical Site", "negation", "additional observations", "condition", "results", "dateTime", "therapeutics", "value", "route of administration"].*
**Input:** *[Clinical text of a patient]*
**Output:** *entity: [clinical entity], type: [characterization / test / evolution / genetics / anatomical Site / negation / additional observations / condition / results / dateTime / therapeutics / value / route of administration];*

### 6.17 CLIP

CLIP[36] is an English dataset focused on extracting actionable clinical items from hospital discharge summaries. It comprises 718 discharge notes with more than 107, 000 sentences, collected from the popular clinical dataset MIMIC-III. The notes were annotated with seven types of follow-up action items such as label tests, procedures, etc. Annotations were performed by five medical professionals with custom-built annotation tools.

- **Language:** English
- **Clinical Stage:** Discharge and Administration
- **Sourced Clinical Document Type:** Discharge Summary
- **Clinical Specialty:** Critical Care
- **Application Method:** Link of CLIP Dataset

#### 6.17.1 Task: CLIP

This task is to identify the clinical action items for physicians from hospital discharge notes.

**Task type:** *Text Classification*
**Instruction:** *Given the discharge summary of a patient, identify the clinical action items for physicians from hospital discharge notes. Specifically, the clinical action items include the following types:*
*- "Patient Instructions": Post-discharge instructions that are directed to the patient, so the PCP (primary care provider) can ensure the patient understands and performs them, such as: 'No driving until post-op visit and you are no longer taking pain medications.'*
*- "Appointment": Appointments to be made by the PCPs, or monitored to ensure the patient attends them, such as: 'The patient requires a neurology consult at XYZ for evaluation.'*
*- "Medications": Medications that the PCP either needs to ensure that the patient is taking correctly (e.g., time-limited medications) or new medications that may need dose adjustment, such as: 'The patient was instructed to hold ASA and refrain from NSAIDs for 2 weeks.'*
*- "Lab": Laboratory tests that either have results pending or need to be ordered by the PCP, such as: 'We ask that the patients' family physician repeat these tests in 2 weeks to ensure resolution.'*
*- "Procedure": Procedures that the PCP needs to either order, ensure another caregiver orders, or ensure the patient undergoes, such as: 'Please follow-up for EGD with GI.'*

*- "Imaging": Imaging studies that either have results pending or need to be ordered by the PCP, such as: 'Superior segment of the left lower lobe: rounded density which could have been related to infection, but follow-up for resolution recommended to exclude possible malignancy.'*
*- "Other": Other actionable information that is important to relay to the PCP but does not fall under existing aspects (e.g., the need to closely observe the patient's diet, or fax results to another provider), such as: 'Since the patient has been struggling to gain weight this past year, we will monitor his nutritional status and trend weights closely.'*
*Assuming the number of action items is N, return the N recognized action items in the output.*
*Return your answer in the following format. DO NOT GIVE ANY EXPLANATION:*
*action items: item 1, item 2, ..., item N*
*The optional list for "item" is ["Patient Instructions", "Appointment", "Medications", "Lab", "Procedure", "Imaging", "Other"].*
**Input:** *[Discharge summary of a patient]*
**Output:** *action items: one or more of the above labels*

### 6.18 cMedQA

The cMedQA datasets contain two versions called cMedQA v1.0[37] and cMedQA v2.0[38]. cMedQA v1.0 is a Chinese-language dataset comprising medical question answer (QA) from an online Chinese healthcare forum (http://www.xywy.com). In the corpus of the dataset, questions contain symptoms, diagnosis, treatment, drug usage, etc., and corresponding answers are written by certified doctors. This dataset consists of 54,000 questions and over 101,000 answers. cMedQA v2.0 is an expanded and refined version of cMedQA v1.0. The cMedQA v2.0 dataset contains double the amount of data compared to v1.0 and refined the data processing steps such as standardization and noise reduction.

- **Language:** Chinese
- **Clinical Stage:** Triage and Referral
- **Sourced Clinical Document Type:** Consultation Record
- **Clinical Specialty:** General
- **Application Method:** Link of cMedQA Dataset1, Link of cMedQA Dataset2

#### 6.18.1 Task: cMedQA

This task is to generate answers from a doctor's perspective in Chinese for a medical consultation.

**Task type:** *Question Answering*
**Instruction:** *Given the medical consultation in Chinese, generate the anser from a doctor's perspective in Chinese.*
*Return your answer in the following format. DO NOT GIVE ANY EXPLANATION:*
**Input:** *[Medical consultation]*
**Output:** *[responses for the medical consultation]*

### 6.19 DialMed

DialMed[39] is a collection of 11,996 annotated medical dialogues in Chinese language, designed for medication recommendation in telemedicine consultations. Records in this dataset are collected from the Chunyu-Doctor platform and consist of high-quality interactions between doctors and patients. This dataset covers 16 common diseases from three departments and 70 standardized medications. The annotations of the dataset were performed by three annotators with relevant medical backgrounds and under the guidance of a doctor, and the consistency was checked with Cohen's kappa coefficient.

- **Language:** Chinese
- **Clinical Stage:** Treatment and Intervention
- **Sourced Clinical Document Type:** Consultation Record
- **Clinical Specialty:** Pulmonology, Gastroenterology, Dermatology, Pharmacology
- **Application Method:** Link of DialMed Dataset

#### 6.19.1 Task: DialMed

This task is to recommended medications for a medical consultation record in Chinese.

> **Task type:** *Text Classification*
> **Instruction:** *Given the medical consultation record in Chinese, where the recommended medications from the doctor are masked as "[MASK]", predict those recommended medications. Note that the number of medications is equal to the number of "[MASK]", assumed to be N.*
> *Return your answer in the following format. DO NOT GIVE ANY EXPLANATION:*
> *medication: label 1, label 2, ..., label N*
> *The optional list for "label" is: ["酮康唑", "板蓝根", "右美沙芬", "莫沙必利", "风寒感冒颗粒", "双黄连口服液", "蒲地蓝消炎口服液", "水飞蓟素", "米诺环素", "氯雷他定", "布地奈德", "苏黄止咳胶囊", "胶体果胶铋", "哈西奈德", "谷胱甘肽", "二硫化硒", "泰诺", "硫磺皂", "对乙酰氨基酚", "奥司他韦", "甘草酸苷", "红霉素", "西替利嗪", "克拉霉素", "氢化可的松", "复方甲氧那明胶囊", "三九胃泰", "替诺福韦", "健胃消食片", "炉甘石洗剂", "蒙脱石", "曲美布汀", "阿奇霉素", "扶正化瘀胶囊", "依巴斯汀", "感冒灵", "他克莫司", "氨溴索", "康复新液", "多烯磷脂酰胆碱", "恩替卡韦", "桉柠蒎肠溶软胶囊", "曲安奈德", "甘草片", "左氧氟沙星", "奥美拉唑", "铝镁化合物", "复方消化酶", "头孢类", "甲氧氯普胺", "地塞米松", "美沙拉秦", "双环醇", "肠炎宁", "抗病毒颗粒", "阿莫西林", "川贝枇杷露", "谷氨酰胺", "山莨菪碱", "阿达帕林", "孟鲁司特", "糠酸莫米松", "快克", "布洛芬", "益生菌", "通窍鼻炎颗粒", "阿昔洛韦", "生理氯化钠溶液", "连花清瘟胶囊", "黄连素"].*
> **Input:** *[Medical consultation record]*
> **Output:** *medication: [one or more of the above labels];*

### 6.20 DiSMed

The DiSMed dataset[40] is a Spanish dataset designed for named entity recognition applied to the de-identification of medical texts. It includes 692 manually annotated radiology reports, compiled from the Medical Imaging Databank of the Valencian Region (BIMCV). The dataset contains brain imaging radiology reports, covering a range of patient-related data that may include identifying information. Annotations were performed manually by three annotators, using a set of six named entity categories derived from the HIPAA-defined Protected Health Information (PHI) and additional relevant entities.

- **Language:** Spanish
- **Clinical Stage:** Research
- **Sourced Clinical Document Type:** Radiology Report
- **Clinical Specialty:** Radiology
- **Application Method:** Link of DiSMed Dataset

#### 6.20.1 Task: DiSMed-NER

This task is to extract the following types of entities from the clinical records: "NAME", "DIR", "LOC", "NUM", "FECHA", "INST".

**Task type:** *Named Entity Recognition*
**Instruction:** *Given the clinical text of a patient in Spanish, extract the following types of entities from the clinical text:*
*- "NAME": Names and surnames (patient and others)*
*- "DIR": Full addresses, including streets, numbers and zip codes.*
*- "LOC": Cities, inside and outside addresses.*
*- "NUM": Numbers or alphanumeric strings that might identify someone, including digital signatures, patient numbers, medical numbers, medical license numbers, and others.*
*- "FECHA": Dates.*
*- "INST": Hospitals, healthcare centers, or other institutions that might point to someone's location.*
*Return your answer in the following format. DO NOT GIVE ANY EXPLANATION:*
*entity: ..., type: ...;*
*...*
*entity: ..., type: ...;*
*The optional list for "type" is ["NAME", "DIR", "LOC", "NUM", "FECHA", "INST"].*
**Input:** *[Clinical text of a patient]*
**Output:** *entity: [clinical entity], type: [NAME / DIR / LOC / NUM / FECHA / INST];*

## 6.21 MIE

MIE[41] is a Chinese language dataset designed for extracting medical information from online consultations. It comprises 1,120 medical dialogues segmented in over 18,000 windows and 46,000 labels. The corpus originates from the Chunyu Doctor online platform, focusing on cardiology-related cases. Data was annotated with four categories (symptoms, tests, surgeries, and other lifestyle information) with five statuses (patient-positive, patient-negative, doctor-positive, doctor-negative, unknown). The labeling was conducted by three trained annotators with the supervision of physicians.

- **Language:** Chinese
- **Clinical Stage:** Initial Assessment
- **Sourced Clinical Document Type:** Consultation Record
- **Clinical Specialty:** Cardiology
- **Application Method:** Link of MIE Dataset

### 6.21.1 Task: Medical entity type extraction

This task is to extract types of medical entities.

**Task type:** *Event Extraction*
**Instruction:** *Given the medical consultation in Chinese, extract the following types of medical entities:*
*1. "symptom": This type of entity refers to the symptoms mentioned by the patient and the doctor.*
*- The optional list of "entity" for "symptom" is ["行动不便", "战栗抽搐", "心慌", "背痛", "头晕", "呃逆", "腹部不适", "高血压", "高血糖", "呼吸困难", "胸闷", "高血脂", "恶心", "呕吐", "胸痛", "乏力", "出汗", "发热", "休克", "晕厥", "感冒","咳嗽", "流涕", "头痛", "胃部不适", "僵硬", "发绀", "糖尿病", "贫血", "水肿", "心绞痛", "甲亢", "早搏", "心律不齐", "房间隔缺损", "房颤", "心衰", "心肌梗死", "先天性心脏病", "心肌缺血", "室间隔缺损", "心肌炎", "冠心病", "心肌病", "心脏肥大"].*
*- The optional list of "status" for "symptom" is ["病人-阳性", "病人-阴性", "医生-阳性", "医生-阴性", "未知"], which means the symptom appeared in the patient, the symptom was not presented in the patient, the*

*symptom was diagnosed by the doctor, the symptom was excluded by the doctor, and the status is unknown, respectively.*

*2. "surgery": This type of entity refers to the surgery operations mentioned by the patient and the doctor.*
*- The optional list of "entity" for "surgery" is ["介入", "射频消融", "搭桥", "支架"].*
*- The optional list of "status" for "surgery" is ["病人-阳性", "病人-阴性", "医生-阳性", "医生-阴性", "未知"], which means the surgery was done on the patient, the surgery was not done on the patient, the surgery was recommended by the doctor, the surgery was deprecated by the doctor, and the status is unknown, respectively.*

*3. "examination": This type of entity refers to the medical tests mentioned by the patient and the doctor.*
*- The optional list of "entity" for "examination" is ["心电图", "彩超", "心肌酶", "体检", "造影", "超声", "ct", "血常规", "甲状腺功能", "胸片", "b超", "肾功能", "平板", "cta", "测血压", "核磁共振"].*
*- The optional list of "status" for "examination" is ["病人-阳性", "病人-阴性", "医生-阳性", "医生-阴性", "未知"], which means the examination was done on the patient, the examination was not done on the patient, the examination was recommended by the doctor, the examination was deprecated by the doctor, and the status is unknown, respectively.*

*4. "general information": This type of entity refers to some general information that is relevant to the patient:*
*- The optional list of "entity" for "general information" is ["睡眠", "饮食", "精神状态", "大小便", "吸烟", "饮酒"].*
*- The optional list of "status" for "general information" is ["病人-阳性", "病人-阴性", "未知"], which means the status of this entity was normal, the status of this entity was abnormal, and the status is unknown, respectively.*

*Return your answer in the following format. DO NOT GIVE ANY EXPLANATION:*
*entity: ..., type: ..., status: ...;*
*...*
*entity: ..., type: ..., status: ...;*
**Input:** *[Medical consultation in Chinese]*
**Output:** *entity: ..., type: ..., status: ...;*
*...*
*entity: ..., type: ..., status: ...;*

### 6.22 EHRQA

EHRQA[42] is a Chinese medical QA dataset comprising over 210,175 electronic health records (EHR) from a hospital in Zhejiang Province and 3.02 million question-answer pairs from 39 Health Networks. The EHR data contains patient disease histories in free text labeled with concepts such as diseases, symptoms, etc. The QA data includes questions from users and answers from doctors organized by medical departments. Annotations were conducted via a bootstrapping method and human review. The EHRQA-QA task was constructed following the original dataset specifications, while the EHRQA-Primary department and EHRQA-Sub department tasks were newly designed and derived from this dataset as part of our BRIDGE benchmark.

- **Language:** Chinese
- **Clinical Stage:** Triage and Referral

**- Sourced Clinical Document Type:** Consultation Record
**- Clinical Specialty:** General
**- Application Method:** Link of EHRQA Dataset

### 6.22.1 Task: EHRQA-Primary department

This task is to determine the hospital department the patient should visit based on the patient's medical consultation record.
[New task construction setting]: We used the patient's basic information and chief complaint as the model input and assigned the primary department of the hospital as the triage label.

> **Task type:** *Text Classification*
> **Instruction:** *Given the medical consultation record in Chinese, determine the hospital department the patient should visit.*
> *Return your answer in the following format. DO NOT GIVE ANY EXPLANATION:*
> *department: label*
> *The optional list for "label" is ["儿科", "妇产科", "传染病科", "皮肤性病科", "外科", "内科", "五官科"].*
> **Input:** *[Medical consultation record in Chinese]*
> **Output:** *department: [儿科 / 妇产科 / 传染病科 / 皮肤性病科 / 外科 / 内科 / 五官科]*

### 6.22.2 Task: EHRQA-Sub department

This task is to determine the detailed hospital department the patient should visit based on the patient's medical consultation record.
[New task construction setting]: The patient's information and chief complaint were used as input, and the detailed sub-department of the subsequent assigned physician was designated as the triage label.

> **Task type:** *Text Classification*
> **Instruction:** *Given the medical consultation record in Chinese, determine the detailed hospital department the patient should visit.*
> *Return your answer in the following format. DO NOT GIVE ANY EXPLANATION:*
> *department: label*
> *The optional list for "label" is ["泌尿外科", "性病科", "胃肠外科", "肝胆外科", "骨科", "小儿内科", "妇科", "小儿精神科", "普外科", "其他传染病", "皮肤病", "消化内科", "风湿免疫科", "口腔科", "产科", "肝病科", "肛肠外科", "肾内科", "眼科", "血管外科", "小儿外科", "乳腺外科", "心胸外科", "烧伤科", "血液科", "内分泌科", "新生儿科", "神经外科", "呼吸内科"].*
> **Input:** *[Medical consultation record in Chinese]*
> **Output:** *department: [泌尿外科 / 性病科 / 胃肠外科 / 肝胆外科 / 骨科 / 小儿内科 / 妇科 / 小儿精神科 / 普外科 / 其他传染病 / 皮肤病 / 消化内科 / 风湿免疫科 / 口腔科 / 产科 / 肝病科 / 肛肠外科 / 肾内科 / 眼科 / 血管外科 / 小儿外科 / 乳腺外科 / 心胸外科 / 烧伤科 / 血液科 / 内分泌科 / 新生儿科 / 神经外科 / 呼吸内科].*

### 6.22.3 Task: EHRQA-QA

This task is to provide the answer from the doctor's perspective in Chinese for a patient's information and consultation record.
[New task construction setting]: Based on each patient's EHR, we extracted relevant medical information, including department information, patient demographics, problem summary, and specific query, and used the physician's actual answer as the expected output.

> **Task type:** *Question Answering*
> **Instruction:** *Given the patient's information and consultation record in Chinese, provide the answer from the doctor's perspective in Chinese.*
> *Return your answer in the following format. DO NOT GIVE ANY EXPLANATION:*
> 医生*: ...*
> **Input:** *[patient's information and consultation record in Chinese]*
> **Output:** *[doctor's response]*

## 6.23 Ex4CDS

Ex4CDS[43] is a German-language dataset focusing on kidney disease. It was designed to generate explainable clinical decision support. The dataset includes 720 annotated textual justifications written by four junior and four senior physicians, involving 120 kidney transplant patients in three medical endpoints (rejection, infection, and graft loss). Annotations of this dataset cover multiple semantic layers including medical entities, relations, temporal markers, etc. They were labeled by automatic annotation tools and finalized by physician review.

- **Language:** German
- **Clinical Stage:** Treatment and Intervention
- **Sourced Clinical Document Type:** General EHR Note
- **Clinical Specialty:** Nephrology
- **Application Method:** Link of Ex4CDS Dataset

### 6.23.1 Task: Ex4CDS

This task is to extract the medical entities with their corresponding types, factuality, and progression from the physician's explanations.
[New task construction setting]: We selected clinically relevant entities from patients' clinical notes and linked each entity with its corresponding factuality and progression attributes. The task is formulated as an event extraction task, requiring the model to identify entities and simultaneously determine their associated status information within the same inference process.

> **Task type:** *Named Entity Recognition*
> **Instruction:** *Given the following physician's explanation, extract the medical entities with their corresponding types, factuality, and progression. Specifically, this explanation was generated by a physician to predict negative outcomes in kidney disease patients within the next 90 days, including rejection, death-censored graft loss, and infection. We need to extract all the following information for each entity:*
> *1. Entity types:*
> *- "Condition": A pathological medical condition of a patient, can describe for instance a symptom or a disease.*
> *- "DiagLab": Particular diagnostic procedures which have been carried out.*
> *- "LabValues": Mentions of lab values.*
> *- "HealthState": A positive condition of the patient.*
> *- "Measure": Mostly numeric values, often in the context of medications or lab values, but can also be a description if a value changes, e.g., raises.*
> *- "Medication": A medication.*

*- "Process": Describes a particular process, such as blood pressure or heart rate, often related to vital parameters.*
*- "TimeInfo": Describes temporal information, such as 2 weeks ago or January.*
*- "Other": Additional relevant information which influences the health condition and the risk.*
*2. Factuality:*
*- "positive": indicates that something is present.*
*- "negative": indicates that something is not present.*
*- "speculated": indicates that something is not present, but might occur in the future.*
*- "unlikely": defines a kind of speculation, but expresses a tendency towards negation.*
*- "minor": expresses that something is present, but to a lower extent or in a lower amount.*
*- "possible_future": expresses that something is not there, but might occur in the future.*
*- "None": if no factuality is given.*
*3. Progression:*
*- "increase_risk_factor": A state/process that causes the respective endpoint (upstream in a causal chain). Increases the risk that endpoint occurs causally and probabilistically.*
*- "decrease_risk_factor": A state/process that prevents the respective endpoint (upstream in a causal chain). Decreases the risk that endpoint occurs causally and probabilistically.*
*- "increase_symptom": A state/process whose absence is a consequence of the respective endpoint (downstream in a causal chain). Increases risk probabilistically, but not causally.*
*- "decrease_symptom": A state/process whose occurrence is a consequence of the respective endpoint (downstream in a causal chain). Decreases risk probabilistically, but not causally.*
*- "Conclusion": The physician makes a concluding statement.*
*- "None": if no progression is given.*
*Return your answer in the following format. DO NOT GIVE ANY EXPLANATION:*
*entity: ..., type: ..., factuality: ..., progression: ...;*
*...*
*entity: ..., type: ..., factuality: ..., progression: ...;*
*The optional list for "type" is ["Condition", "DiagLab", "LabValues", "HealthState", "Measure", "Medication", "Process", "TimeInfo", "Other"].*
*The optional list for "factuality" is ["positive", "negative", "speculated", "unlikely", "minor", "possible_future", "None"].*
*The optional list for "progression" is ["increase_risk_factor", "decrease_risk_factor", "increase_symptom", "decrease_symptom", "Conclusion", "None"].*
**Input:** *[Discharge summary of a patient]*
**Output:** *entity: ..., type: [Condition / DiagLab / LabValues / HealthState / Measure / Medication / Process / TimeInfo / Other], factuality: [positive / negative / speculated / unlikely / minor / possible_future / None], progression: [increase_risk_factor / decrease_risk_factor / increase_symptom / decrease_symptom / Conclusion / None];*
*...*
*entity: ..., type: [Condition / DiagLab / LabValues / HealthState / Measure / Medication / Process / TimeInfo / Other], factuality: [positive / negative / speculated / unlikely / minor / possible_future / None], progression: [increase_risk_factor / decrease_risk_factor / increase_symptom / decrease_symptom / Conclusion / None];*

### 6.24 GOUT-CC

Gout-CC dataset[44] is an English dataset focused on GOUT detection. It consists of two splits: GOUT-CC-2019-CORPUS which includes 300 chief complaints collected in 2019 and GOUT-CC-2020-CORPUS which includes 8037 chief complaints collected in one month period of 2020. The dataset originates from records of an academic medical center in the southern United States. Each complaint includes two kinds of annotations: A predicted gout flare annotation labeled by a practicing rheumatologist (MD) and a PhD informatician (JDO) and a reviewed consensus labeled by a rheumatologist (MD) and a post-doctoral fellow (GR). Cohen's Kappa coefficient was applied to calculate the consistency of the annotations.

- **Language:** English
- **Clinical Stage:** Diagnosis and Prognosis
- **Sourced Clinical Document Type:** Admission Note
- **Clinical Specialty:** Endocrinology
- **Application Method:** Link of GOUT-CC Dataset

#### 6.24.1 Task: GOUT-CC-Consensus

This task is to determine whether a patient was experiencing a gout flare at the time of the Emergency Department visit.
[New task construction setting]: We used the annotations from the "consensus" column as the ground-truth labels. Because instances labeled as "unknown" typically corresponded to notes lacking meaningful clinical information (e.g., note titles or administrative text) and "Unknown" may lead to label misinterpretation, we simplified the task into a binary classification problem, requiring the model to determine whether a gout flare was present (yes or no).

> **Task type:** *Text Classification*
> **Instruction:** *Given the patient's chief complaint, determine whether the patient was experiencing a gout flare at the time of the Emergency Department visit.*
> *Return your answer in the following format. DO NOT GIVE ANY EXPLANATION:*
> *Gout flare: label*
> *The optional list for "label" is ["Yes", "No"]*
> **Input:** *Chief complaint of a patient]*
> **Output:** *Gout flare: [Yes / No]*

### 6.25 n2c2 2006

The n2c2 2006[45] dataset is an English dataset designed for Protected Health Information (PHI) de-identification. It includes discharge summaries from Partners HealthCare, compiled to support the de-identification of PHI. The dataset contains clinical narratives centered on patient discharge, covering a broad spectrum of medical treatments and prescriptions. Annotations were performed using a custom annotation schema by domain experts, adhering to guidelines developed for the 2006 i2b2 NLP challenge.

- **Language:** English
- **Clinical Stage:** Research
- **Sourced Clinical Document Type:** Discharge Summary
- **Clinical Specialty:** Pulmonology
- **Application Method:** Link of n2c2 2006 Dataset

#### 6.25.1 Task: n2c2 2006-De-identification

This task is to identify the PHI context and their associated type within the clinical document.

> **Task type:** *Named Entity Recognition*
> **Instruction:** *Given the clinical document of a patient, identify the Personal Health Information (PHI) context and their associated type within the document.*

*Return your answer in the following format. DO NOT GIVE ANY EXPLANATION:*
*PHI context: ..., type: ...;*
*...*
*PHI context: ..., type: ...;*
*The optional list for "type" is ["PATIENT", "ID", "LOCATION", "AGE", "DOCTOR", "PHONE", "HOSPITAL", "DATE"].*
**Input:** *[Clinical text of a patient]*
**Output:** *PHI context: [PHI context], type: [PATIENT / ID / LOCATION / AGE / DOCTOR / PHONE / HOSPITAL / DATE];*

### 6.26 i2b2 2009

The i2b2 2009 dataset[46] is an English dataset designed for medication information extraction. It includes 1,249 patient discharge summaries, compiled from Partners Healthcare. The dataset contains clinical narratives and was developed as part of the i2b2 medication extraction challenge. Annotations were performed by a team consisting of a physician and a researcher, adhering to a sequential annotation strategy.

- **Language:** English
- **Clinical Stage:** Treatment and Intervention
- **Sourced Clinical Document Type:** Discharge Summary
- **Clinical Specialty:** Pharmacology
- **Application Method:** Link of i2b2 2009 Dataset

#### 6.26.1 Task: Medication extraction

This task is to extract the medications from the discharge summary. Then, for each extracted medication, further extract the following entities: "dosage", "mode", "frequency", "duration", "reason", "list/narrative".

**Task type:** *Event Extraction*
**Instruction:** *Given the discharge summary of a patient, extract the medications from the discharge summary. Then, for each extracted medication, further extract the following entities:*
*- "dosage": Indicating the amount of a medication used in each administration.*
*- "mode": Indicating the route for administering the medication.*
*- "frequency": Indicating how often each dose of the medication should be taken.*
*- "duration": Indicating how long the medication is to be administered.*
*- "reason": Stating the medical reason for which the medication is given.*
*- "list/narrative": Indicating whether the medication information appears in a list structure or in narrative running text in the discharge summary.*
*Return your answer in the following format. DO NOT GIVE ANY EXPLANATION:*
*medication: ..., dosage: ..., mode: ..., frequency: ..., duration: ..., reason: ..., list/narrative: ...;*
*...*
*medication: ..., dosage: ..., mode: ..., frequency: ..., duration: ..., reason: ..., list/narrative: ...;*
*The optional list for "list/narrative" is ["list", "narrative"].*
*For each extracted medication, if any of the above entity ("dosage", "mode", "frequency", "duration", "reason", "list/narrative") is not mentioned in the discharge summary, then please label this entity as "not mentioned".*
**Input:** *[Discharge summary of a patient]*
**Output:** *medication: [name of the medication], dosage: [medication dosage], mode: [medication route], frequency: [dose frequency], duration: [medication administration time length], reason: [reason for giving the medication], list/narrative: [list / narrative];*

### 6.27 i2b2 2010

The i2b2 2010 dataset[47] is an English dataset designed for named entity recognition. It was compiled from Partners Healthcare, Beth Israel Deaconess Medical Center, and the University of Pittsburgh Medical Center. The dataset contains discharge summaries and progress reports, covering a wide range of clinical concepts such as medical problems, treatments, and tests. Annotations were performed using a manually annotated reference standard corpus by the i2b2 team in collaboration with the VA Salt Lake City Health Care System, adhering to task-specific guidelines for concepts, assertions, and fine-grained relations.

- **Language:** English
- **Clinical Stage:** Treatment and Intervention (Task "n2c2 2010-Assertion" has clinical stage "Discharge and Administration")
- **Sourced Clinical Document Type:** Discharge Summary, Progress Note
- **Clinical Specialty:** Critical Care
- **Application Method:** Link of i2b2 2010 Dataset

#### 6.27.1 Task: n2c2 2010-Concept

This task is to identify and extract the text corresponding to patient medical problems, treatments, and tests.

> **Task type:** *Named Entity Recognition*
> **Instruction:** *Given the discharge summary of a patient, extract the following types of entities from the discharge summary:*
> *- "problem": Phrases that contain observations made by patients or clinicians about the patient's body or mind that are thought to be abnormal or caused by a disease.*
> *- "treatment": Phrases that describe procedures, interventions, and substances given to a patient in an effort to resolve a medical problem.*
> *- "test": Phrases that describe procedures, panels, and measures that are done to a patient or a body fluid or sample in order to discover, rule out, or find more information about a medical problem.*
> *Return your answer in the following format. DO NOT GIVE ANY EXPLANATION:*
> *entity: ..., type: ...;*
> *...*
> *entity: ..., type: ...;*
> *The optional list for "type" is ["problem", "treatment", "test"].*
> **Input:** *[Discharge summary of a patient]*
> **Output:** *entity: [clinical entity], type: [problem / treatment / test];*

#### 6.27.2 Task: n2c2 2010-Assertion

This task is to extract the medical problems and classify the corresponding assertions made on given medical concepts as being present, absent, or possible in the patient, conditionally present in the patient under certain circumstances, hypothetically present in the patient at some future point, and mentioned in the patient report but associated with someone other than the patient.

> **Task type:** *Named Entity Recognition*
> **Instruction:** *Given the discharge summary of a patient, extract the medical problems from the discharge summary, and for each medical problem, classify it into one of the following categories:*
> *- "present": Problems associated with the patient can be present.*
> *- "absent: The discharge summary asserts that the problem does not exist in the patient.*
> *- "possible": The discharge summary asserts that the patient may have a problem, but there is uncertainty*

*expressed in the discharge summary.*
*- "conditional": The mention of the medical problem asserts that the patient experiences the problem only under certain conditions.*
*- "hypothetical": Medical problems that the note asserts the patient may develop.*
*- "not associated with patient": The mention of the medical problem is associated with someone who is not the patient.*
*Return your answer in the following format. DO NOT GIVE ANY EXPLANATION:*
*problem: ..., type: ...;*
*...*
*problem: ..., type: ...;*
*The optional list for "type" is ["present", "absent", "possible", "conditional", "hypothetical", "not associated with patient"].*
**Input:** *[Discharge summary of a patient]*
**Output:** *problem: [medical problem], type: [present / absent / possible / conditional / hypothetical / not associated with patient];*

#### 6.27.3 Task: n2c2 2010-Relation

This task is to extract relations of pairs of given reference standard concepts from a sentence.

**Task type:** *Event Extraction*
**Instruction:** *Given the discharge summary of a patient, extract the following types of entities from the discharge summary:*
*- "problem": Phrases that contain observations made by patients or clinicians about the patient's body or mind that are thought to be abnormal or caused by a disease.*
*- "treatment": Phrases that describe procedures, interventions, and substances given to a patient in an effort to resolve a medical problem.*
*- "test": Phrases that describe procedures, panels, and measures that are done to a patient or a body fluid or sample in order to discover, rule out, or find more information about a medical problem.*
*Then, extract the following three types of relations if applicable:*
*1. relations between "treatment" and "problem":*
*- "TrIP": This includes mentions where the treatment improves or cures the problem.*
*- "TrWP": This includes mentions where the treatment is administered for the problem but does not cure the problem, does not improve the problem, or makes the problem worse.*
*- "TrCP": The implied context is that the treatment was not administered for the medical problem that it ended up causing.*
*- "TrAP": This includes mentions where a treatment is given for a problem, but the outcome is not mentioned in the sentence.*
*- "TrNAP": This includes mentions where treatment was not given or discontinued because of a medical problem that the treatment did not cause.*
*2. relations between "test" and "problem":*
*- "TeRP": This includes mentions where a test is conducted and the outcome is known.*
*- "TeCP": This includes mentions where a test is conducted and the outcome is not known.*
*3. relations between "problem" and "problem":*
*- "PIP": This includes medical problems that describe or reveal aspects of the same medical problem and those that cause other medical problems.*
*Return your answer in the following format. DO NOT GIVE ANY EXPLANATION:*
*entity_1: ..., entity_2: ..., relation: ...;*

> *...*
> *entity_1: ..., entity_2: ..., relation: ...;*
> *The optional list for "relation" is ["TrIP", "TrWP", "TrCP", "TrAP", "TrNAP", "TeRP", "TeCP", "PIP"].*
> **Input:** *[Discharge summary of a patient]*
> **Output:** *entity_1: [a problem/treatment/test entity], entity_2: [a problem/treatment/test entity], relation: [TrIP / TrWP / TrCP / TrAP / TrNAP / TeRP / TeCP / PIP];*

## 6.28 n2c2 2014 - De-identification

The n2c2 2014 - De-identification dataset[48] is an English dataset designed for de-identification of clinical narratives. It includes 1,304 medical records from 296 diabetic patients, compiled from the Research Patient Data Repository of Partners Healthcare. The dataset contains longitudinal clinical notes, reflecting the progression of heart disease in diabetic patients over time. Annotations were performed using a comprehensive set of i2b2-PHI categories, extending beyond HIPAA definitions, and were conducted with both automatic and manual checks to ensure accuracy. Authentic Protected Health Information (PHI) was replaced with realistic surrogates, and the annotated corpus was used as a gold standard for system evaluations in the 2014 i2b2/UTHealth NLP shared task.

- **Language:** English
- **Clinical Stage:** Research
- **Sourced Clinical Document Type:** General EHR Note
- **Clinical Specialty:** Endocrinology
- **Application Method:** Link of n2c2 2014 - De-identification Dataset

### 6.28.1 Task: n2c2 2014-De-identification

This task is to identify all the following type of PHI information within the document: 'STATE', 'LOCATION-OTHER', 'STREET', 'PHONE', 'FAX', 'ZIP', 'DOCTOR', 'DATE', 'EMAIL', 'DEVICE', 'COUNTRY', 'HOSPITAL', 'HEALTHPLAN', 'ORGANIZATION', 'USERNAME', 'BIOID', 'CITY', 'PROFESSION', 'PATIENT', 'URL', 'AGE', 'IDNUM', 'MEDICALRECORD'.

> **Task type:** *Named Entity Recognition*
> **Instruction:** *Given the clinical document of a patient, identify the Personal Health Information (PHI) context and their associated type within the document.*
> *Return your answer in the following format. DO NOT GIVE ANY EXPLANATION:*
> *PHI context: ..., type: ...;*
> *...*
> *PHI context: ..., type: ...;*
> *The optional list of PHI types is ["STATE", "LOCATION-OTHER", "STREET", "PHONE", "FAX", "ZIP", "DOCTOR", "DATE", "EMAIL", "DEVICE", "COUNTRY", "HOSPITAL", "HEALTHPLAN", "ORGANIZATION", "USERNAME", "BIOID", "CITY", "PROFESSION", "PATIENT", "URL", "AGE", "IDNUM", "MEDICALRECORD"].*
> **Input:** *[Clinical document of a patient]*
> **Output:** *PHI context: [PHI context], type: [STATE / LOCATION-OTHER / STREET / PHONE / FAX / ZIP / DOCTOR / DATE / EMAIL / DEVICE / COUNTRY / HOSPITAL / HEALTHPLAN / ORGANIZATION / USERNAME / BIOID / CITY / PROFESSION / PATIENT / URL / AGE / IDNUM / MEDICALRECORD];*

## 6.29 IMCS-V2

The IMCS-V2 (Intelligent Medical Consultation System - Version 2) dataset[49] comprises real-world doctor–patient dialogues collected from an online medical consultation platform. This dataset covers 10 common pediatric diseases, including pediatric bronchitis, fever, and diarrhea. Each dialogue includes interactions between patients (or their guardians) and physicians. Through systematic annotation, the dataset labels various medical information, such as medical entities and dialogue acts. It supports four tasks: named entity recognition (NER), dialogue act classification (DAC), symptom recognition (SR), and medical report generation (MRG). IMCS-V2 is included in the CBLUE benchmark[30]. We adapted the CBLUE version for downstream task construction.

- **Language:** Chinese
- **Clinical Stage:** Triage and Referral (IMCS-V2-DAC), Inital Assessment (IMCS-V2-NER, SR, MRG)
- **Sourced Clinical Document Type:** Consultation Record
- **Clinical Specialty:** Pediatrics
- **Application Method:** Link of IMCS-V2-NER Dataset

#### 6.29.1 Task: IMCS-V2-NER

This task is to identify five categories of medical entities from doctor–patient dialogue texts.

> **Task type:** *Named Entity Recognition*
> **Instruction:** *Given the text from medical consultation in Chinese, extract the medical entities mentioned by the patient and doctors, including the following types: - "symptom"(症状)：病人因患病而表现出来的异常状况，如"发热"、"呼吸困难"、"鼻塞"等。*
> *- "drug"(药品名)：具体的药物名称，如"妈咪爱"、"蒙脱石散"、"蒲地蓝"等。*
> *- "drug category"(药物类别)：根据药物功能进行划分的药物种类，如"消炎药"、"感冒药"、"益生菌"等。*
> *- "examination"(检查)：医学检验，如"血常规"、"x光片"、"CRP分析"等。*
> *- "operation"(操作)：相关的医疗操作，如"输液"、"雾化"、"接种疫苗"等。*
> *Return your answer in the following format. DO NOT GIVE ANY EXPLANATION: entity: ..., type: ...; ... entity: ..., type: ...; The optional list for "type" is ["symptom", "drug", "drug category", "examination", "operation"].*
> **Input:** *[Medical dialogue between doctor and patient]*
> **Output:** *entity: [entity span], type: [symptom / drug / drug category / examination / operation];*
> *...*
> *entity: [entity span], type: [symptom / drug / drug category / examination / operation];*

#### 6.29.2 Task: IMCS-V2-SR

This task is to recognize patient symptom information, including both normalized labels and status, from the conversation.

> **Task type:** *Event Extraction*
> **Instruction:** *Given the medical consultation in Chinese, recognize the normalized symptoms mentioned by the patient and doctors and identify the global status of symptoms based on the dialogue, including:*
> *- "positive": 代表确定病人患有该症状*
> *- "negative": 代表确定病人没有患有该症状*
> *- "uncertain": 代表无法根据上下文确定病人是否患有该症状*
> *Specifically, the status of the symptom is based on the entire dialogue, not just the current sentence.*
> *Return your answer in the following format. DO NOT GIVE ANY EXPLANATION:*
> *symptom: ..., status: ...;*
> *...*
> *symptom: ..., status: ...;*

*The optional list for "status" is ["positive", "negative", "uncertain"].*

**Input:** *[Medical dialogue between doctor and patient]*
**Output:** *symptom: [entity span], status: [ positive / negative / uncertain];*
*...*
*symptom: [entity span], status: [ positive / negative / uncertain];*

#### 6.29.3 Task: IMCS-V2-MRG

This task is to generate a structured summarization based on the patient's chief complaint and the full doctor–patient dialogue.

**Task type:** *Summarization*
**Instruction:** *Given the medical consultation in Chinese, generate the brief report based on the dialogue between the patient and doctor. The report should include the following sections:*
*1.* 主诉*(Chief complaint):* 病人自诉（*Self-report*）的总结，包括主要症状或体征；
*2.* 现病史*(Present illness history):* 对话中病人涉及到的现病史的总结，如主要症状的描述（发病情况，发病时间）；
*3.* 辅助检查*(Auxiliary examination):* 对话中病人涉及过的医疗检查的总结，如病人已有的检查项目、检查结果、会诊记录等；
*4.* 既往史*(Past history):* 对话中医生对病人的过去病史的总结，如既往的健康状况、过去曾经患过的疾病等；
*5.* 诊断*(Diagnosis):* 对话中医生对病人的诊断结果的总结，如对疾病的诊断；
*6.* 建议*(Suggestion):* 对话中医生对病人的建议的总结，如检查建议、药物治疗、注意事项。
*Return your answer in the following format. DO NOT GIVE ANY EXPLANATION:*
主诉*: ...*
现病史*: ...*
辅助检查*: ...*
既往史*: ...*
诊断*: ...*
建议*: ...*
**Input:** *[A whole medical dialogue between doctor and patient]*
**Output:** 主诉*: [summarized chief complaint]*
现病史*: [summarized illness history]*
辅助检查*: [summarized auxiliary examination ]*
既往史*: [summarized past history]*
诊断*: [summarized diagnosis]*
建议*: [summarized suggestion from physician]*

#### 6.29.4 Task: IMCS-V2-DAC

This task is to classify the intent of each utterance in the dialogue into one of 16 predefined categories.

**Task type:** *Text Classification*
**Instruction:** *[ Given the utterance from a medical consultation in Chinese, identify the act the speaker is performing.*
*Return your answer in the following format. DO NOT GIVE ANY EXPLANATION:*
*dialogue act: label*

*The optional list for "label" is ["提问-症状", "提问-病因", "提问-基本信息", "提问-已有检查和治疗", "告知-用药建议", "告知-就医建议", "告知-注意事项", "诊断", "告知-症状", "告知-病因", "告知-基本信息", "告知-已有检查和治疗", "提问-用药建议", "提问-就医建议", "提问-注意事项"].*
**Input:** *[A utterance from medical dialogue]*
**Output:** *dialogue act: [提问-症状/ 提问-病因/ 提问-基本信息/ 提问-已有检查和治疗/ 告知-用药建议/ 告知-就医建议/ 告知-注意事项/ 诊断/ 告知-症状/ 告知-病因/ 告知-基本信息/ 告知-已有检查和治疗/ 提问-用药建议/ 提问-就医建议/ 提问-注意事项]*

### 6.30 Japanese Case Reports

The Japanese Case Reports dataset is a Japanese dataset[50] designed for semantic textual similarity tasks. It includes approximately 4,000 annotated sentence pairs, compiled from case reports extracted from the CiNii database. The dataset contains clinical case report texts, covering a wide range of medical scenarios. Annotations were performed by staff with medical backgrounds, using a 6-point scale (0 to 5) to rate semantic similarity, and following guidelines from previous STS tasks.

- **Language:** Japanese
- **Clinical Stage:** Research
- **Sourced Clinical Document Type:** Case Report
- **Clinical Specialty:** General
- **Application Method:** Link of Japanese Case Reports Dataset

#### 6.30.1 Task: JP-STS

This task is to capture semantic textual similarity (STS) between Japanese clinical texts.

**Task type:** *Semantic Similarity*
**Instruction:** *Given the following two clinical sentences that are labeled as "Sentence A" and "Sentence B" in Japanese, decide the similarity of the two sentences according to the following rubric:*
*- "0": The two sentences are completely dissimilar.*
*- "1": The two sentences are not equivalent, but are on the same topic.*
*- "2": The two sentences are not equivalent, but share some details.*
*- "3": The two sentences are roughly equivalent, but some important information differs/missing.*
*- "4": The two sentences are mostly equivalent, but some unimportant details differ.*
*- "5": The two sentences are completely equivalent.*
*Return your answer in the following format. DO NOT GIVE ANY EXPLANATION:*
*similarity score: score*
*The optional list for "score" is ["0", "1", "2", "3", "4", "5"].*
**Input:** *Sentence A: [Clinical sentence A]*
*Sentence B: [Clinical sentence B]*
**Output:** *similarity score: [0 / 1 / 2 / 3 / 4 / 5]*

### 6.31 meddocan

The meddocan dataset[51] is a Spanish dataset designed for de-identification of sensitive information in clinical texts. It includes 1,000 manually selected and synthetically enriched clinical case studies, compiled under the Spanish National Plan for the Advancement of Language Technology (Plan TL) by the Barcelona Supercomputing Center and the Centro Nacional de Investigaciones Oncológicas. The dataset contains clinical case reports enriched with Protected Health Information (PHI), covering a broad range of sensitive data types including names, dates, locations, and identifiers. Annotations were performed using a customized version of AnnotateIt and corrected using the BRAT annotation tool by a hybrid team of healthcare and NLP experts, adhering to guidelines based on the EU GDPR and the US HIPAA standards.

- **Language:** Spanish
- **Clinical Stage:** Research
- **Sourced Clinical Document Type:** Case Report
- **Clinical Specialty:** General
- **Application Method:** Link of meddocan Dataset

#### 6.31.1 Task: meddocan

This task is to extract specific clinical entities from the clinical text.

> **Task type:** *Named Entity Recognition*
> **Instruction:** *Given the clinical text of a patient in Spanish, extract the following types of entities from the clinical text: "TERRITORIO", "FECHAS", "EDAD SUJETO ASISTENCIA", "NOMBRE SUJETO ASISTENCIA", "NOMBRE PERSONAL SANITARIO", "SEXO SUJETO ASISTENCIA", "CALLE", "PAIS", "ID SUJETO ASISTENCIA", "CORREO ELECTRONICO", "ID TITULACION PERSONAL SANITARIO", "ID ASEGURAMIENTO", "HOSPITAL", "FAMILIARES SUJETO ASISTENCIA", "INSTITUCION", "ID CONTACTO ASISTENCIAL", "NUMERO TELEFONO", "PROFESION", "NUMERO FAX", "OTROS SUJETO ASISTENCIA", "CENTRO SALUD", "ID EMPLEO PERSONAL SANITARIO", "IDENTIF VEHICULOS NRSERIE PLACAS", "IDENTIF DISPOSITIVOS NRSERIE", "NUMERO BENEF PLAN SALUD", "URL WEB", "DIREC PROT INTERNET", "IDENTF BIOMETRICOS", "OTRO NUMERO IDENTIF".*
> *Return your answer in the following format. DO NOT GIVE ANY EXPLANATION:*
> *entity: ..., type: ...;*
> *...*
> *entity: ..., type: ...;*
> *The optional list for "type" is ["TERRITORIO", "FECHAS", "EDAD SUJETO ASISTENCIA", "NOMBRE SUJETO ASISTENCIA", "NOMBRE PERSONAL SANITARIO", "SEXO SUJETO ASISTENCIA", "CALLE", "PAIS", "ID SUJETO ASISTENCIA", "CORREO ELECTRONICO", "ID TITULACION PERSONAL SANITARIO", "ID ASEGURAMIENTO", "HOSPITAL", "FAMILIARES SUJETO ASISTENCIA", "INSTITUCION", "ID CONTACTO ASISTENCIAL", "NUMERO TELEFONO", "PROFESION", "NUMERO FAX", "OTROS SUJETO ASISTENCIA", "CENTRO SALUD", "ID EMPLEO PERSONAL SANITARIO", "IDENTIF VEHICULOS NRSERIE PLACAS", "IDENTIF DISPOSITIVOS NRSERIE", "NUMERO BENEF PLAN SALUD", "URL WEB", "DIREC PROT INTERNET", "IDENTF BIOMETRICOS", "OTRO NUMERO IDENTIF"].*
> **Input:** *[Discharge summary of a patient]*
> **Output:** *entity: [clinical entity], type: [TERRITORIO / FECHAS / EDAD SUJETO ASISTENCIA / NOMBRE SUJETO ASISTENCIA / NOMBRE PERSONAL SANITARIO / SEXO SUJETO ASISTENCIA / CALLE / PAIS / ID SUJETO ASISTENCIA / CORREO ELECTRONICO / ID TITULACION PERSONAL SANITARIO / ID ASEGURAMIENTO / HOSPITAL / FAMILIARES SUJETO ASISTENCIA / INSTITUCION / ID CONTACTO ASISTENCIAL / NUMERO TELEFONO / PROFESION / NUMERO FAX / OTROS SUJETO ASISTENCIA / CENTRO SALUD / ID EMPLEO PERSONAL SANITARIO / IDENTIF VEHICULOS NRSERIE PLACAS / IDENTIF DISPOSITIVOS NRSERIE / NUMERO BENEF PLAN SALUD / URL WEB / DIREC PROT INTERNET / IDENTF BIOMETRICOS / OTRO NUMERO IDENTIF];*

### 6.32 MEDIQA_2019_Task2_RQE

The MEDIQA_2019_Task2_RQE dataset[52] is an English dataset designed for recognizing question entailment in the medical domain. It includes 230 test pairs of consumer health questions and frequently asked questions (FAQs), as well as 8,890 training and 302 validation medical question pairs. The dataset was compiled by the U.S. National Library of Medicine (NLM) using a collection of clinical and consumer health questions. The dataset contains medical question pairs and covers a variety of health-related topics. Annotations were manually validated by medical experts, adhering to the definition that a question A entails question B if every answer to B is also a complete or partial answer to A.

- **Language:** English
- **Clinical Stage:** Triage and Referral
- **Sourced Clinical Document Type:** Consultation Record
- **Clinical Specialty:** General
- **Application Method:** Link of MEDIQA_2019_Task2_RQE Dataset

#### 6.32.1 Task: MEDIQA 2019-RQE

This task is to identify entailment between two questions in the context of QA.

> **Task type:** *Natural Language Inference*
> **Instruction:** *Given the following two clinical questions labeled as "Question A" and "Question B", determine if the answer to "Question B" is also the answer to "Question A", either exactly or partially.*
> *Return your answer in the following format. DO NOT GIVE ANY EXPLANATION:*
> *answer: label*
> *The optional list for "label" is ["true", "false"].*
> **Input:** *Question A: [Question A context]*
> *Question B: [Question B context]*
> **Output:** *answer: [true / false]*

### 6.33 MedNLI

The MedNLI dataset[53] is an English dataset designed for natural language inference (NLI) in the clinical domain. It includes 14,049 unique sentence pairs, compiled from the MIMIC-III v1.3 database of de-identified clinical records. The dataset contains premise-hypothesis pairs derived from clinical notes, particularly the Past Medical History section, covering a wide range of medical conditions and concepts such as diseases, symptoms, and medications. Annotations were performed by board-certified clinicians, adhering to custom-developed guidelines to ensure consistency and quality.

- **Language:** English
- **Clinical Stage:** Research
- **Sourced Clinical Document Type:** General EHR Note
- **Clinical Specialty:** Critical Care
- **Application Method:** Link of MedNLI Dataset

#### 6.33.1 Task: MedNLI

This task is to classify the inference relation between the premise-hypothesis statements pair into one of the three classes: "entailment", "contradiction", or "neutral".

> **Task type:** *Natural Language Inference*
> **Instruction:** *Given the premise-hypothesis statements pair labeled as "Premise statement" and "Hypothesis statement", classify the inference relation between two statements into one of the three classes: "entailment", "contradiction", or "neutral".*

*Return your answer in the following format. DO NOT GIVE ANY EXPLANATION:*
*relation: label*
*The optional list for "label" is ["entailment", "contradiction", "neutral"].*
**Input:** *Premise statement: [Premise statement context]*
*Hypothesis statement: [Hypothesis statement context]*
**Output:** *relation: [entailment / contradiction / neutral]*

### 6.34 MedSTS

The MedSTS dataset[54] is an English dataset designed for semantic textual similarity tasks. It includes 174,629 sentence pairs compiled from de-identified clinical notes at the Mayo Clinic. The dataset contains clinical sentences reflecting a wide range of medical concepts, including signs and symptoms, disorders, procedures, and medications. A subset of 1,250 sentence pairs, called MedSTS_ann, was annotated with semantic similarity scores ranging from 0 to 5 by two experienced clinical experts. Annotations were performed using manual scoring guidelines adapted from SemEval shared tasks. The dataset was developed to support NLP research aimed at reducing redundancy in electronic health records and improving clinical decision-making.

- **Language:** English
- **Clinical Stage:** Research
- **Sourced Clinical Document Type:** General EHR Note
- **Clinical Specialty:** General
- **Application Method:** Email the corresponding author for access

#### 6.34.1 Task: MedSTS

This task is to determine the similarity of the two sentences in a score from 0 to 5, with 0 indicating that the medical semantics of the sentences are completely independent and 5 signifying semantic equivalence.

**Task type:** *Semantic Similarity*
**Instruction:** *Given two clinical sentences labeled as "Sentence A" and "Sentence B", determine the similarity of the two sentences in a score from 0 to 5, with 0 indicating that the medical semantics of the sentences are completely independent and 5 indicating significant semantic equivalence.*
*Return your answer in the following format. DO NOT GIVE ANY EXPLANATION:*
*similarity score: score*
*The optional list for "score" is ["0", "1", "2", "3", "4", "5"].*
**Input:** *Sentence A: [Clinical sentence A]*
*Sentence B: [Clinical sentence B]*
**Output:** *similarity score: [0 / 1 / 2 / 3 / 4 / 5]*

### 6.35 mtsamples

The mtsamples dataset[55] is an English dataset designed for clinical outcome prediction and self-supervised language model pre-training. It includes a large collection of publicly available medical transcriptions of medical dictations, covering a wide range of medical specialties and conditions. Annotations were used as part of the self-supervised pre-training process to integrate practical medical knowledge into clinical outcome models.

- **Language:** English
- **Clinical Stage:** Research
- **Sourced Clinical Document Type:** Case Report
- **Clinical Specialty:** General

**- Application Method:** Link of mtsamples Dataset

#### 6.35.1 Task: MTS

This task is to classify the clinical text into specific categories.

> **Task type:** *Text Classification*
> **Instruction:** *Given the transcribed clinical text, classify the clinical text into its corresponding clinical specialty or document type (can be more than one).*
> *Return your answer in the following format. DO NOT GIVE ANY EXPLANATION:*
> *answer: label_1, label_2, ..., label_n*
> *The optional list for "label" is ['Allergy / Immunology', 'Autopsy', 'Bariatrics', 'Cardiovascular / Pulmonary', 'Chiropractic', 'Consult - History and Phy.', 'Cosmetic / Plastic Surgery', 'Dentistry', 'Dermatology', 'Diets and Nutritions', 'Discharge Summary', 'ENT - Otolaryngology', 'Emergency Room Reports', 'Endocrinology', 'Gastroenterology', 'General Medicine', 'Hematology - Oncology', 'Hospice - Palliative Care', 'IME-QME-Work Comp etc.', 'Lab Medicine - Pathology', 'Letters', 'Nephrology', 'Neurology', 'Neurosurgery', 'Obstetrics / Gynecology', 'Office Notes', 'Ophthalmology', 'Orthopedic', 'Pain Management', 'Pediatrics - Neonatal', 'Physical Medicine - Rehab', 'Podiatry', 'Psychiatry / Psychology', 'Radiology', 'Rheumatology', 'SOAP / Chart / Progress Notes', 'Sleep Medicine', 'Speech - Language', 'Surgery', 'Urology'].*
> **Input:** *[Clinical text of a patient]*
> **Output:** *answer: [one or more of the above labels]*

### 6.36 mtsamples-temporal

The mtsamples-temporal dataset[56] is an English dataset designed for temporal information extraction and timeline reconstruction. It includes 286 medical transcription documents covering four clinical specialties: discharge summaries, psychiatry-psychology, paediatrics, and emergency, compiled from MTSamples, a public repository of clinical text. The dataset contains annotated time expressions (TIMEXes) such as dates, times, durations, frequencies, and age-related expressions, capturing how temporal information is expressed in clinical narratives. Annotations were performed manually by two annotators, following custom annotation guidelines based on the TimeML schema, with extensions for domain-specific temporal constructs.

**- Language:** English
**- Clinical Stage:** Initial Assessment
**- Sourced Clinical Document Type:** Discharge Summary
**- Clinical Specialty:** Pediatrics, Psychology
**- Application Method:** Link of mtsamples-temporal Dataset

#### 6.36.1 Task: MTS-Temporal

This task is to extract the following types of time expression from the clinical text: 'time', 'frequency', 'age_related', 'date', 'duration', 'other'.

> **Task type:** *Named Entity Recognition*
> **Instruction:** *Given the clinical text, extract the time expressions from the clinical text, and categorize each of them into one of the following categories: "time", "frequency", "age_related", "date", "duration", "other".*
> *Return your answer in the following format. DO NOT GIVE ANY EXPLANATION:*
> *time expression: ..., category: ...;*
> *...*

> *time expression: ..., category: ...;*
> *The optional list for "category" is ["time", "frequency", "age_related", "date", "duration", "other"].*
> **Input:** *[Clinical text of a patient]*
> **Output:** *time expression: [time expression within the clinical text], category: [time / frequency / age_related / date / duration / other];*

### 6.37 n2c2 2018 Track2

The n2c2 2018 Track2 dataset[57] is an English dataset designed for adverse drug event (ADE) and medication information extraction. It includes 505 discharge summaries drawn from the MIMIC-III clinical care database. The dataset contains narrative clinical records annotated for medication-related concepts (e.g., drug name, strength, dosage, frequency, route, duration, reason, and ADE) and their relations. Annotations were performed by two independent annotators with conflict resolution by a third annotator.

- **Language:** English
- **Clinical Stage:** Treatment and Intervention
- **Sourced Clinical Document Type:** Discharge Summary
- **Clinical Specialty:** Pharmacology
- **Application Method:** Link of n2c2 2018 Track2 Dataset

#### 6.37.1 Task: n2c2 2018-ADE&medication

This task is to identify the drugs and extract the following attributes from the clinical text: "adverse drug event", "dosage", "duration", "form", "frequency", "reason", "route", "strength".

> **Task type:** *Event Extraction*
> **Instruction:** *Given the longitudinal medical records of a patient, identify the drugs and extract the following attributes from the clinical text: "adverse drug event", "dosage", "duration", "form", "frequency", "reason", "route", "strength". Please note that a drug may have none or some of these attributes.*
> *Return your answer in the following format. DO NOT GIVE ANY EXPLANATION:*
> *drug: ..., attribute: ..., ..., attribute: ...;*
> *...*
> *drug: ..., attribute: ..., ..., attribute: ...;*
> *The optional list for "attribute" is ["adverse drug event", "dosage", "duration", "form", "frequency", "reason", "route", "strength"].*
> **Input:** *[longitudinal medical records of a patient]*
> **Output:** *drug: [name of the drug], adverse drug event / dosage / duration / form / frequency / reason / route / strength: [drug attribute], ..., adverse drug event / dosage / duration / form / frequency / reason / route / strength: [drug attribute];*

### 6.38 NorSynthClinical

The NorSynthClinical dataset[58] is a Norwegian dataset designed for family history information extraction. It includes 477 sentences and 6,030 tokens, compiled from synthetically produced clinical statements by a domain expert in genetic cardiology. The dataset contains synthetic descriptions of patients' family histories, focusing primarily on cardiac conditions. Annotations were performed using the Brat annotation tool by a clinician and independent annotators, adhering to iteratively developed guidelines informed by inter-annotator agreement evaluations and clinical domain knowledge.

- **Language:** Norwegian

- **Clinical Stage:** Initial Assessment
- **Sourced Clinical Document Type:** General EHR Note
- **Clinical Specialty:** Cardiology
- **Application Method:** Link of NorSynthClinical Dataset

#### 6.38.1 Task: NorSynthClinical-NER

This task is to extract the following types of entities from the clinical records:"FAMILY", "SELF", INDEX", "CONDITION", "EVENT", "SIDE", "AGE", "NEG", "AMOUNT", "TEMPORAL".

> **Task type:** *Named Entity Recognition*
> **Instruction:** *Given the clinical text of a patient in Norwegian, extract the following types of entities from the clinical records:*
> *- "FAMILY": Used to describes various family members.*
> *- "SELF": Used to refer to the patient.*
> *- "INDEX": Used to designate the property of being the index patient or proband.*
> *- "CONDITION": Used to describe a range of clinical conditions such as diseases, diagnoses, various types of mutations, test results, treatments, and vital state.*
> *- "EVENT": Used to describe clinical events.*
> *- "SIDE": Used to describe the side of the family and thus modify "FAMILY" entities.*
> *- "AGE": Used to describe the age of a family member.*
> *- "NEG": Used to mark lexical items that signal negation.*
> *- "AMOUNT": Used to describe quantifiers that describe numerical properties of clinical entities.*
> *- "EMPORAL": Used to modify typically position "CONDITION" or "EVENT" entities in time.*
> *Return your answer in the following format. DO NOT GIVE ANY EXPLANATION:*
> *entity: ..., type: ...;*
> *...*
> *entity: ..., type: ...;*
> *The optional list for "type" is ["FAMILY", "SELF", "INDEX", "CONDITION", "EVENT", "SIDE", "AGE", "NEG", "AMOUNT", "EMPORAL"].*
> **Input:** *[Clinical text of a patient]*
> **Output:** *entity: [clinical entity], type: [FAMILY / SELF / INDEX / CONDITION / EVENT / SIDE / AGE / NEG / AMOUNT / EMPORAL];*

#### 6.38.2 Task: NorSynthClinical-RE

This task is to extract the following types of relations from the clinical records:"Holder", "Modifier", "Related_to", "Subset", "Partner".

> **Task type:** *Event Extraction*
> **Instruction:** *Given the clinical text of a patient in Norwegian, extract the following types of relations from the clinical records:*
> *- "Holder": Relations between a "CONDITION" or "EVENT" entity and its holder, a "FAMILY" or "SELF" or "INDEX" entity.*
> *- "Modifier": Relations hold between modifier entities and clinical entities.*
> *- "Related_to": Relations specify relations between family members and always hold between "FAMILY" entities.*
> *- "Subset": Relations specify relations between family members, where one is a subset of the other.*

- *"Partner": Relations specify relations between entities of the "FAMILY" type, used to identify couples that are able to provide offspring.*
*Below are the definitions of the entity types mentioned above: - "FAMILY": Used to describes various family members.*
- *"SELF": Used to refer to the patient.*
- *"INDEX": Used to designate the property of being the index patient or proband.*
- *"CONDITION": Used to describe a range of clinical conditions such as diseases, diagnoses, various types of mutations, test results, treatments, and vital state.*
- *"EVENT": Used to describe clinical events.*
*Return your answer in the following format. DO NOT GIVE ANY EXPLANATION:*
*entity_1: ..., entity_2: ..., relation: ...;*
*...*
*entity_1: ..., entity_2: ..., relation: ...;*
*The optional list for "relation" is ["Holder", "Modifier", "Related_to", "Subset", "Partner"].*
**Input:** *[Clinical text of a patient]*
**Output:** *entity_1: [clinical entity 1], entity_2: [clinical entity 2], relation: [Holder / Modifier / Related_to / Subset / Partner];*

### 6.39 NUBES

The NUBES dataset[59] is a Spanish dataset designed for negation and uncertainty detection in clinical texts. It includes 29,682 sentences and 518,068 tokens, compiled from anonymized health records provided by a Spanish private hospital. The dataset contains excerpts from various clinical sections, such as Chief Complaint, Present Illness, and Diagnostic Tests, covering a wide range of medical contexts. Annotations were performed using the BRAT tool by a team of linguists and a medical expert, following guidelines adapted from the IULA-SCRC corpus.

- **Language:** Spanish
- **Clinical Stage:** Research
- **Sourced Clinical Document Type:** General EHR Note
- **Clinical Specialty:** General
- **Application Method:** Link of NUBES Dataset

#### 6.39.1 Task: NUBES

This task is to extract the negation and uncertainty cues and scope.
[New task construction setting]: We constructed the task by combining the extraction of entity text, determining its marker type, and analyzing its scope into a event extraction task.

**Task type:** *Event Extraction*
**Instruction:** *Given the clinical text of a patient in Spanish, extract the following types of entities from the clinical text:*
- *"NegSynMarker": Syntactic negation cue.*
- *"NegLexMarker": Lexical negation cue.*
- *"NegMorMarker": Morphological negation cue.*
- *"UncertSynMarker": Syntactic uncertainty cue.*
- *"UncertLexMarker": Lexical uncertainty cue.*
*Then, for each extracted entity, extract the scope that the entity affects.*
*Return your answer in the following format. DO NOT GIVE ANY EXPLANATION:*
*entity: ..., type: ..., scope: ...;*
*...*

*entity: ..., type: ..., scope: ...;*
*The optional list for "type" is ["NegSynMarker", "NegLexMarker", "NegMorMarker", "UncertSynMarker", "UncertLexMarker"].*
**Input:** *[Clinical text of a patient]*
**Output:** *entity: [clinical entity], type: [NegSynMarker / NegLexMarker / NegMorMarker / UncertSynMarker / UncertLexMarker], scope: [scope of the negation cue];*

## 6.40 MTS-Dialog-MEDIQA-2023

The MTS-Dialog-MEDIQA-2023 dataset[60] is an English dataset designed for automatic summarization of doctor-patient conversations into clinical notes. It includes 1,701 simulated dialogue-note pairs, comprising 15,969 dialogue turns and 5,870 summary sentences, created from de-identified clinical notes sourced from the public MTSamples collection. The dataset contains synthetic medical dialogues reflecting six specialties, generated by trained annotators with medical backgrounds using detailed annotation guidelines. Annotations were validated through a rigorous quality control process involving rubric-based manual review and corrections.

- **Language:** English
- **Clinical Stage:** Initial Assessment
- **Sourced Clinical Document Type:** Consultation Record
- **Clinical Specialty:** General
- **Application Method:** Link of MTS-Dialog-MEDIQA-2023 Dataset

### 6.40.1 Task: MEDIQA 2023-chat-A

This task involves summarizing various clinical sections based on the dialogue between a doctor and a patient. The clinical sections to be summarized include: fam/sochx [FAMILY HISTORY/SOCIAL HISTORY], genhx [HISTORY OF PRESENT ILLNESS], pastmedicalhx [PAST MEDICAL HISTORY], cc [CHIEF COMPLAINT], pastsurgical [PAST SURGICAL HISTORY], allergy, ros [REVIEW OF SYSTEMS], medications, assessment, exam, diagnosis, disposition, plan, edcourse [EMERGENCY DEPARTMENT COURSE], immunizations, imaging, gynhx [GYNECOLOGIC HISTORY], procedures, other_history, and labs.

**Task type:** *Summarization*
**Instruction:** *Given the clinical conversation between a doctor and a patient, summarize the clinical section based on the conversation.*
*Return your answer in the following format. DO NOT GIVE ANY EXPLANATION:*
*summary: ...*
**Input:** *[Clinical conversation between a doctor and a patient]*
**Output:** *summary: [summary of the specific clinical section]*

### 6.40.2 Task: MEDIQA 2023-sum-A

This task is to classify the topic of the conversation between a doctor and a patient into one of the following categories: "family history/social history", "history of patient illness", "past medical history", "chief complaint", "past surgical history", "allergy", "review of systems", "medications", "assessment", "exam", "diagnosis", "disposition", "plan", "emergency department course", "immunizations", "imaging", "gynecologic history", "procedures", "other history", "labs".

**Task type:** *Text Classification*
**Instruction:** *Given the clinical conversation between a doctor and a patient, determine the topic of the conversation.*

*Return your answer in the following format. DO NOT GIVE ANY EXPLANATION:*
*topic: label*
*The optional list for "label" is ["family history/social history", "history of patient illness", "past medical history", "chief complaint", "past surgical history", "allergy", "review of systems", "medications", "assessment", "exam", "diagnosis", "disposition", "plan", "emergency department course", "immunizations", "imaging", "gynecologic history", "procedures", "other history", "labs"].*
**Input:** *[Clinical conversation between a doctor and a patient]*
**Output:** *topic: [family history/social history / history of patient illness / past medical history / chief complaint / past surgical history / allergy / review of systems / medications / assessment / exam / diagnosis / disposition / plan / emergency department course / immunizations / imaging / gynecologic history / procedures / other history / labs]*

#### 6.40.3 Task: MEDIQA 2023-sum-B

**This task has the same description as "MTS-Dialog-MEDIQA-2023-chat-task-A". The only difference between the two tasks is the test set.**
This task involves summarizing various clinical sections based on the dialogue between a doctor and a patient. The clinical sections to be summarized include: fam/sochx [FAMILY HISTORY/SOCIAL HISTORY], genhx [HISTORY OF PRESENT ILLNESS], pastmedicalhx [PAST MEDICAL HISTORY], cc [CHIEF COMPLAINT], pastsurgical [PAST SURGICAL HISTORY], allergy, ros [REVIEW OF SYSTEMS], medications, assessment, exam, diagnosis, disposition, plan, edcourse [EMERGENCY DEPARTMENT COURSE], immunizations, imaging, gynhx [GYNECOLOGIC HISTORY], procedures, other_history, and labs.

**Task type:** *Summarization*
**Instruction:** *Given the clinical conversation between a doctor and a patient, summarize the clinical section based on the conversation.*
*Return your answer in the following format. DO NOT GIVE ANY EXPLANATION:*
*summary: ...*
**Input:** *[Clinical conversation between a doctor and a patient]*
**Output:** *summary: [summary of the specific clinical section]*

### 6.41 RuMedDaNet

The RuMedDaNet dataset[61] is a Russian language dataset designed for medical question answering. It includes 1,564 records compiled from diverse medical domains such as therapeutic medicine, physiology, anatomy, pharmacology, and biochemistry. The dataset contains medical-related context passages paired with yes/no questions, aiming to assess a model's understanding of domain-specific knowledge. Questions were manually created by assessors based on open-source medical texts to avoid template-like structures. Annotations were performed by human assessors to ensure balanced positive and negative responses, adhering to ethical standards of medical data handling.

- **Language:** Russian
- **Clinical Stage:** Triage and Referral
- **Sourced Clinical Document Type:** General EHR Note
- **Clinical Specialty:** General
- **Application Method:** Link of RuMedDaNet Dataset

#### 6.41.1 Task: RuMedDaNet

This task is to answer the clinical question with either "yes" or "no" based on the clinical text.

**Task type:** *Natural Language Inference*
**Instruction:** *Given the clinical text labeled as "Context" and the clinical question labeled as "Question" in Russian, answer the clinical question with either "yes" or "no". Return your answer in the following format. DO NOT GIVE ANY EXPLANATION: answer: label The optional list for "label" is ["yes", "no"].*
**Input:** *Context: [Clinical context]*
*Question: [Clinical question]*
**Output:** *answer: [yes / no]*

### 6.42 CHIP-CDN

The CHIP-CDN dataset[29] is sourced from diagnostic text of Chinese EHRs. It focuses on clinical terminology normalization by mapping original diagnostic expressions to standardized ICD codes through manual annotation. CHIP-CDN was one of the shared tasks in the CHIP-2021 challenges and is included in the CBLUE benchmark[30]. We adapted the CBLUE version for downstream task construction.

- **Language:** Chinese
- **Clinical Stage:** Discharge and Administration
- **Sourced Clinical Document Type:** General EHR Note
- **Clinical Specialty:** General
- **Application Method:** Link of CHIP-CDN Dataset

#### *6.42.1 Task: CBLUE-CDN*

This task is to normalize diagnostic terms in Chinese EHRs by mapping them to standard ICD codes (text).

**Task type:** *Normalization and Coding*
**Instruction:** *Given the original diagnostic text from electronic healthcare records in Chinese, normalize them to the corresponding standard diagnostic terms. Specifically, use the names of standardized terms from the 《国际疾病分类 ICD-10 北京临床版v601》, covering diagnosis, surgery, medication, examination, laboratory testing, and symptoms. There may be multiple appropriate normalized terms for the original diagnostic text. Assuming the number of normalized terms is N, return the names of N normalized terms in the output.*
*Return your answer in the following format. DO NOT GIVE ANY EXPLANATION:*
*Normalized terms: label 1, label 2, ..., label N*
*The optional list for "label" is the names of normalized terms (not code) from the 《国际疾病分类 ICD-10 北京临床版v601》, covering diagnosis, surgery, medication, examination, laboratory testing, and symptoms.*
**Input:** *[Diagnostic terms from EHRs]*
**Output:** *Normalized terms: [normalized ICD code]*

### 6.43 CHIP-CTC

The CHIP-CTC dataset[62] is derived from clinical trial documents and aims to identify matching clinical trial eligibility criteria. Clinical trials require screening appropriate participants, which involves retrieving and matching patient cases against predefined inclusion and exclusion criteria. CHIP-CTC was one of the shared tasks in the CHIP-2019 challenges and is included in the CBLUE benchmark[30]. We adapted the CBLUE version for downstream task construction.

- **Language:** Chinese
- **Clinical Stage:** Research
- **Sourced Clinical Document Type:** General EHR Note

- **Clinical Specialty:** General
- **Application Method:** Link of CHIP-CTC Dataset

#### *6.43.1 Task: CHIP-CTC*

This task is to determine whether a given clinical case matches specific inclusion or exclusion criteria from a clinical trial.

> **Task type:** *Text Classification*
> **Instruction:** *Given the clinical text in Chinese, identify the clinical trial criterion that this text meets. Return your answer in the following format. DO NOT GIVE ANY EXPLANATION:*
> *clinical trial criterion: label*
> *The optional list for "label" is ["疾病", "症状-患者感受", "体征-医生检测", "怀孕相关", "肿瘤进展", "疾病分期", "过敏耐受", "器官组织状态", "预期寿命", "口腔相关", "药物", "治疗或手术", "设备", "护理", "诊断", "实验室检查", "风险评估", "受体状态", "年龄", "特殊病人特征", "读写能力", "性别", "教育情况", "居住情况", "种族", "知情同意", "参与其它试验", "研究者决定", "能力", "伦理审查", "依存性", "成瘾行为", "睡眠", "锻炼", "饮食", "酒精使用", "性取向", "吸烟状况", "献血", "病例来源", "残疾群体", "健康群体", "数据可及性"].*
> **Input:** *[Clinical text]*
> **Output:** *clinical trial criterion: [疾病 / 症状-患者感受 / 体征-医生检测 / 怀孕相关 / 肿瘤进展 / 疾病分期 / 过敏耐受 / 器官组织状态 / 预期寿命 / 口腔相关 / 药物 / 治疗或手术 / 设备 / 护理 / 诊断 / 实验室检查 / 风险评估 / 受体状态 / 年龄 / 特殊病人特征 / 读写能力 / 性别 / 教育情况 / 居住情况 / 种族 / 知情同意 / 参与其它试验 / 研究者决定 / 能力 / 伦理审查 / 依存性 / 成瘾行为 / 睡眠 / 锻炼 / 饮食 / 酒精使用 / 性取向 / 吸烟状况 / 献血 / 病例来源 / 残疾群体 / 健康群体 / 数据可及性]*

### 6.44 CHIP-MDCFNPC

The CHIP-MDCFNPC (Medical Dialog Clinical Findings Negative and Positive Classification) dataset[29] is sourced from Chinese doctor–patient dialogues of online medical consultations. To objectively and comprehensively describe a patient's condition, this dataset uses the concept of clinical findings—specifically focusing on the classification of these findings as positive or negative, indicating the presence or absence of a specific condition. The definitions of positivity and negativity are generally derived from the patient's subjective complaint and the physician's diagnostic assessment. The annotation process follows the SOAP (Subjective, Objective, Assessment, Plan) framework, a widely adopted problem-oriented medical documentation method. The positive/negative classification is primarily applied to entities identified in the Subjective (S) and Assessment (A) sections. The preprocessing pipeline includes SOAP section alignment, named entity recognition in the S and A sections, and subsequent polarity labeling. This task was part of the CHIP-2021 shared tasks and is included in the CBLUE benchmark[30]. We adapted the CBLUE version for downstream task construction.

- **Language:** Chinese
- **Clinical Stage:** Initial Assessment
- **Sourced Clinical Document Type:** Consultation Record
- **Clinical Specialty:** General
- **Application Method:** Link of CHIP-MDCFNPC Dataset

#### *6.44.1 Task: CHIP-MDCFNPC*

This task is to identify clinical findings and determine their presence status (positive or negative) from doctor–patient dialogue history.

**Task type:** *Event Extraction*
**Instruction:** *Given the medical consultation in Chinese, extract the clinical findings mentioned by the patient and doctors and identify their status based on the dialogue, including:*
- "阳性": 已有症状疾/病等相关，医生诊断（包含多个诊断结论），以及假设未来可能发生的疾病等，如："如果不治疗的话，大概率会引起A疾病"，"A疾病"标注为阳性；
- "阴性": 未患有的疾病症状相关；
- "其他": 未知的标注其他，一般指用户没有回答、不知道或者回答不明确/模棱两可不好推断的情况。
- "不标注": 无实际意义的不标注，一般是医生的解释说的是一般知识，和病人当前的状态条件独立不具有标注意义，及有些检查项带疾病名称的，识别的疾病（乙肝五项/乙肝抗体），药品名中出现的"疾病"不标注。
*Return your answer in the following format. DO NOT GIVE ANY EXPLANATION:*
*findings: ..., status: ...;*
*...*
*findings: ..., status: ...;*
*The optional list for "status" is ["阳性", "阴性", "其他", "不标注"].*
**Input:** *[A utterance from medical consultation]*
**Output:** *findings: [findings mention], status: [阳性 / 阴性 / 其他 / 不标注]; ... findings: [findings mention], status: [阳性 / 阴性 / 其他 / 不标注];*

### 6.45 MedDG

The MedDG dataset[63] contains doctor–patient dialogues collected from Chinese online medical consultation platforms. It covers 12 gastrointestinal diseases and includes over 17,000 dialogues and 380,000 utterances. As an entity-centric dataset, the physicians identified and formalized 160 types medical entities through a systematic annotation. Each dialogue is annotated with entities across five major categories—diseases, symptoms, severity, examinations, and medications. This dataset is included in the CBLUE benchmark[30], and we adapted the CBLUE version for downstream task construction.

- **Language:** Chinese
- **Clinical Stage:** Triage and Referral
- **Sourced Clinical Document Type:** Consultation Record
- **Clinical Specialty:** Gastroenterology
- **Application Method:** Link of MedDG Dataset

#### *6.45.1 Task: MedDG*

This task is to generate the doctor's response based on the provided dialogue history of a medical consultation.

**Task type:** *Question Answering*
**Instruction:** *Given the medical consultation in Chinese, generate the next response of the doctor based on the dialogue context.*
*Return your answer in the following format. DO NOT GIVE ANY EXPLANATION:*
医生*: ...*
**Input:** *[Dialogue history of medical consultation]*
**Output:** 医生*: [generated response from doctor's perspective]*

### 6.46 n2c2 2014 - Heart Disease Challenge

The n2c2 2014 - Heart Disease Challenge dataset[64] is an English dataset designed for risk factor identification and temporal classification. It includes 1304 longitudinal medical records from 296 diabetic patients, compiled and de-identified for the 2014 i2b2/UTHealth NLP shared task. The dataset contains narrative clinical records, covering a wide range of heart disease risk factors such as hypertension, hyperlipidemia, obesity, smoking, family history of CAD, diabetes, and coronary artery disease (CAD). Annotations were performed using the Multi-purpose Annotation Environment (MAE) by a group of seven medical professionals, adhering to light annotation guidelines developed specifically for this task.

- **Language:** English
- **Clinical Stage:** Initial Assessment (Task "n2c2 2014-Medication" has clinical stage "Treatment and Intervention")
- **Sourced Clinical Document Type:** General EHR Note
- **Clinical Specialty:** Cardiology, Endocrinology
- **Application Method:** Link of n2c2 2014 - Heart Disease Challenge Dataset

#### 6.46.1 Task: n2c2 2014-Diabetes

This task is to extract the indicators that are related to Diabetes based on the clinical document of a patient and classify each extracted indicator into two types of categories.

> **Task type:** *Event Extraction*
> **Instruction:** *Given the clinical document of a patient.*
> *Firstly, extract the indicators that are related to Diabetes.*
> *Secondly, classify each extracted indicator into one of the following categories (denoted as category 1):*
> *- "mention": A diagnosis of Type 1 or Type 2 diabetes, or a mention of a preexisting diagnosis.*
> *- "high A1c": An A1c test value of over 6.5.*
> *- "high glucose": 2 fasting blood glucose measurements of over 126.*
> *Finally, for each extracted indicator, classify it into one of the following categories (denoted as category 2):*
> *- "before DCT": This attribute value is used to indicate that the indicator can only be stated to be present prior to the date of the record.*
> *- "during DCT": This attribute value is used to indicate that a risk factor indicator occurred the day of the date on the record.*
> *- "after DCT": This attribute value is used to indicate that the risk factor indicator applies to the days after the date of the record. Return your answer in the following format. DO NOT GIVE ANY EXPLANATION:*
> *indicator: ..., category_1: ..., category_2: ...;*
> *...*
> *indicator: ..., category_1: ..., category_2: ...;*
> *The optional list for "category_1" is ["mention", "high A1c", "high glucose"]. The optional list for "category_2" is ["before DCT", "during DCT", "after DCT"].*
> **Input:** *[Clinical document of a patient]*
> **Output:** *indicator: [indicators related to Diabetes], category_1: [mention / high A1c / high glucose], category_2: [before DCT / during DCT / after DCT];*

#### 6.46.2 Task: n2c2 2014-CAD

This task is to extract the indicators that are related to Coronary Artery Disease (CAD) based on the clinical document of a patient and classify each extracted indicator into two types of categories.

> **Task type:** *Event Extraction*
> **Instruction:** *Given the clinical document of a patient.*
> *Firstly, extract the indicators that are related to Coronary Artery Disease (CAD).*

*Secondly, classify each extracted indicator into one of the following categories (denoted as category 1):*
*- "mention": A diagnosis of CAD, or a mention of a history of CAD.*
*- "event": MI, STEMI, NSTEMI OR revascularization procedures (bypass surgery, CABG, percutaneous) OR cardiac arrest OR ischemic cardiomyopathy.*
*- "test": Exercise or pharmacologic stress test showing ischemia OR abnormal cardiac catheterization showing coronary stenoses (narrowing).*
*- "symptom": Chest pain consistent with angina.*
*Finally, for each extracted indicator, classify it into one of the following categories (denoted as category 2):*
*- "before DCT": This attribute value is used to indicate that the indicator can only be stated to be present prior to the date of the record.*
*- "during DCT": This attribute value is used to indicate that a risk factor indicator occurred the day of the date on the record.*
*- "after DCT": This attribute value is used to indicate that the risk factor indicator applies to the days after the date of the record. Return your answer in the following format. DO NOT GIVE ANY EXPLANATION:*
*indicator: ..., category_1: ..., category_2: ...;*
*...*
*indicator: ..., category_1: ..., category_2: ...;*
*The optional list for "category_1" is ["mention", "event", "test", "symptom"].*
*The optional list for "category_2" is ["before DCT", "during DCT", "after DCT"].*
**Input:** *[Clinical document of a patient]*
**Output:** *indicator: [indicators related to Diabetes], category_1: [mention / event / test / symptom], category_2: [before DCT / during DCT / after DCT];*

#### 6.46.3 Task: n2c2 2014-Hyperlipidemia

This task is to extract the indicators that are related to Hyperlipidemia/Hypercholesterolemia based on the clinical document of a patient and classify each extracted indicator into two types of categories.

**Task type:** *Event Extractionn*
**Instruction:** *Given the clinical document of a patient.*
*Firstly, extract the indicators that are related to Hyperlipidemia/Hypercholesterolemia.*
*Secondly, classify each extracted indicator into one of the following categories (denoted as category 1):*
*- "mention": A diagnosis of Hyperlipidemia or Hypercholesterolemia or a mention of the patient already having that condition.*
*- "high cholesterol": Total cholesterol of over 240.*
*- "high LDL": LDL measurement of over 100 mg/dL.*
*Finally, for each extracted indicator, classify it into one of the following categories (denoted as category 2):*
*- "before DCT": This attribute value is used to indicate that the indicator can only be stated to be present prior to the date of the record.*
*- "during DCT": This attribute value is used to indicate that a risk factor indicator occurred the day of the date on the record.*
*- "after DCT": This attribute value is used to indicate that the risk factor indicator applies to the days after the date of the record. Return your answer in the following format. DO NOT GIVE ANY EXPLANATION:*
*indicator: ..., category_1: ..., category_2: ...;*
*...*
*indicator: ..., category_1: ..., category_2: ...;*
*The optional list for "category_1" is ["mention", "high cholesterol", "high LDL"].*
*The optional list for "category_2" is ["before DCT", "during DCT", "after DCT"].*

**Input:** *[Clinical document of a patient]*
**Output:** *indicator: [indicators related to Diabetes], category_1: [mention / high cholesterol / high LDL], category_2: [before DCT / during DCT / after DCT];*

#### 6.46.4 Task: n2c2 2014-Hypertension

This task is to extract the indicators that are related to Hypertension based on the clinical document of a patient and classify each extracted indicator into two types of categories.

**Task type:** *Event Extraction*
**Instruction:** *Given the clinical document of a patient.*
*Firstly, extract the indicators that are related to Hypertension.*
*Secondly, classify each extracted indicator into one of the following categories (denoted as category 1):*
*- "mention": A diagnosis of Hypertension or a mention of a pre-existing condition.*
*- "high blood pressure": Blood pressure measurement of over 140/90 mm/hg (if either value is high, the patient has hypertension).*
*Finally, for each extracted indicator, classify it into one of the following categories (denoted as category 2):*
*- "before DCT": This attribute value is used to indicate that the indicator can only be stated to be present prior to the date of the record.*
*- "during DCT": This attribute value is used to indicate that a risk factor indicator occurred the day of the date on the record.*
*- "after DCT": This attribute value is used to indicate that the risk factor indicator applies to the days after the date of the record.*
*Return your answer in the following format. DO NOT GIVE ANY EXPLANATION:*
*indicator: ..., category_1: ..., category_2: ...;*
*...*
*indicator: ..., category_1: ..., category_2: ...;*
*The optional list for "category_1" is ["mention", "high blood pressure"].*
*The optional list for "category_2" is ["before DCT", "during DCT", "after DCT"].*
**Input:** *[Clinical document of a patient]*
**Output:** *indicator: [indicators related to Diabetes], category_1: [mention / high blood pressure], category_2: [before DCT / during DCT / after DCT];*

#### 6.46.5 Task: n2c2 2014-Medication

This task is to extract the indicators that are related to Medication based on the clinical document of a patient and classify each extracted indicator into one of the three categories.

**Task type:** *Event Extraction*
**Instruction:** *Given the clinical document of a patient, extract the indicators that are related to Medication.*
*Then, for each extracted indicator, classify it into one of the following categories:*
*- "before DCT": This attribute value is used to indicate that the indicator can only be stated to be present prior to the date of the record.*
*- "during DCT": This attribute value is used to indicate that a risk factor indicator occurred the day of the date on the record.*
*- "after DCT": This attribute value is used to indicate that the risk factor indicator applies to the days after the date of the record.*
*Return your answer in the following format. DO NOT GIVE ANY EXPLANATION:*
*indicator: ..., category: ...;*

...
*indicator: ..., category: ...;*
*The optional list for "category" is ["before DCT", "during DCT", "after DCT"].*
**Input:** *[Clinical document of a patient]*
**Output:** *indicator: [indicators related to Diabetes], category: [before DCT / during DCT / after DCT];*

## 6.47 CAS

The CAS dataset[65] was derived from French clinical case reports and contains 20,363 sentences and over 397,000 word occurrences. Each sentence is annotated with Concept Unique Identifiers (CUIs) corresponding to French terms in the UMLS, along with negation and uncertainty markers. The annotated corpus was used in the DEFT shared tasks in 2019 and 2020. Based on its annotations, we constructed two downstream tasks.

- **Language:** French
- **Clinical Stage:** Discharge and Administration
- **Sourced Clinical Document Type:** Case Report
- **Clinical Specialty:** General
- **Application Method:** Link of CAS Dataset

### 6.47.1 Task: CAS-label

This task is to extract patient age, gender, and clinical outcome from French clinical case reports.

**Task type:** *Event Extraction*
**Instruction:** *Given the clinical case report in French, extract the following medical information:*
*- "age": l'âge de la personne dont le cas est décrit, au moment du dernier élément clinique rapporté dans le cas clinique, normalisé sous la forme d'un entier (soit 0 pour un nourrisson de moins d'un an, 1 pour un enfant de moins de deux ans, y compris un an et demi, 20 pour un patient d'une vingtaine d'années, etc.).*
*- "genre": le genre de la personne dont le cas est décrit, parmi deux valeurs normalisées : féminin, masculin (il n'existe aucun cas de dysgénésie ou d'hermaphrodisme dans le corpus). si le genre n'est pas mentionné, retournez "None".*
*- "issue": l'issue parmi cinq valeurs possibles: (1) guérison (le problème clinique décrit dans le cas a été traité et la personne est guérie), (2) amélioration (l'état clinique est amélioré sans qu'on ne puisse conclure à une guérison), (3) stable (soit l'état clinique reste stationnaire, soit il est impossible de déterminer entre amélioration et détérioration), (4) détérioration (l'état clinique se dégrade), ou (5) décès (lorsque le décès concerne directement le cas clinique décrit). si le problème n'est pas mentionné, retournez "None".*
*Return your answer in the following format. DO NOT GIVE ANY EXPLANATION:*
*age: ..., genre: ..., issue: ...*
*The optional list for "genre" is ["féminin", "masculin", "None"].*
*The optional list for "issue" is ["guérison", "amélioration", "stable", "détérioration", "décès", "None"].*
**Input:** *[A clinical case report]*
**Output:** *age: [age mention], genre: [féminin / masculin / None], issue: [guérison / amélioration / stable / détérioration / décès / None]*

### 6.47.2 Task: CAS-evidence

This task is to extract evidences of information about gender, etiology, and clinical outcome from French clinical case reports.

**Task type:** *Summarization*
**Instruction:** *Given the clinical care report in French, extract the evidence of the following medical information from the original text:*
*- "genre": le genre de la personne dont le cas est décrit, parmi deux valeurs normalisées : féminin, masculin (il n'existe aucun cas de dysgénésie ou d'hermaphrodisme dans le corpus).*
*- "origine": l'origine (motif de la consultation ou de l'hospitalisation) pour le dernier événement clinique ayant motivé la consultation. Cette catégorie intègre généralement les pathologies, signes et symptômes (par exemple, "une tuméfaction lombaire droite, fébrile avec frissons" ou "un contexte d'asthénie et d'altération de l'état général"), plus rarement les circonstances d'un accident ("une chute de 12 mètres, par défénestration, avec réception ventrale", "un AVP moto" ou "pense avoir été violée"). Le suivi clinique se trouve dans la continuité d'événements précédents. Il ne constitue pas un motif de consultation.*
*- "issue": l'issue parmi cinq valeurs possibles: (1) guérison (le problème clinique décrit dans le cas a été traité et la personne est guérie), (2) amélioration (l'état clinique est amélioré sans qu'on ne puisse conclure à une guérison), (3) stable (soit l'état clinique reste stationnaire, soit il est impossible de déterminer entre amélioration et détérioration), (4) détérioration (l'état clinique se dégrade), ou (5) décès (lorsque le décès concerne directement le cas clinique décrit).*
*Return your answer in the following format. DO NOT GIVE ANY EXPLANATION:*
*evidence of genre: ...;*
*evidence of origine: ...;*
*evidence of issue: ...;*
*The optional evidence of "genre" is the word or shot phrase that indicates the genre of the person; The optional evidence of "origine" and "issue" is the sentence or shot paragraph that indicates the origine and issue of the person, respectively. If the evidence is not mentioned, return "None".*
**Input:** *[A clinical case report]*
**Output:** *evidence of genre: patient;*
*evidence of origine: None;*
*evidence of issue: None;*

## 6.48 RuMedNLI

The RuMedNLI dataset[61] is a Russian language dataset designed for natural language inference in the clinical domain. It includes 14,049 records compiled through translation and manual correction of the English MedNLI dataset. The dataset contains pairs of clinical sentences—premise and hypothesis—annotated with one of three labels: entailment, contradiction, or neutral. Annotations were generated by clinicians and translated into Russian using two automatic translation systems, followed by human post-editing to ensure domain accuracy, including adaptation of medical abbreviations and measurement units.

- **Language:** Russian
- **Clinical Stage:** Research
- **Sourced Clinical Document Type:** General EHR Note
- **Clinical Specialty:** Critical Care
- **Application Method:** Link of RuMedNLI Dataset

### *6.48.1 Task: RuMedNLI-NLI*

This task is to classify the inference relation between the premise-hypothesis statements pair into one of the three classes: "entailment", "contradiction", or "neutral".

**Task type:** *Natural Language Inference*
**Instruction:** *Given the premise-hypothesis statements pair labeled as "Premise statement" and "Hypothesis statement" in Russian, classify the inference relation between two statements into one of the three classes: "entailment", "contradiction", or "neutral".*
*Return your answer in the following format. DO NOT GIVE ANY EXPLANATION:*
*relation: label*
*The optional list for "label" is ["entailment", "contradiction", "neutral"].*
**Input:** *Premise statement: [Premise statement context]*
*Hypothesis statement: [Hypothesis statement context]*
**Output:** *relation: [entailment / contradiction / neutral]*

## 6.49 RuDReC

The RuDReC dataset[61] is a Russian language dataset designed for named entity recognition tasks in the pharmaceutical domain. It includes 2,587 annotated reviews compiled from the Russian drug review platform Otzovik. The dataset contains user-generated reviews about pharmaceutical products, covering aspects such as drug effectiveness, adverse drug reactions, and indications. Annotations were performed using a rule-based system and verified by domain experts, adhering to a scheme that identifies drug-related entities.

- **Language:** Russian
- **Clinical Stage:** Treatment and Intervention
- **Sourced Clinical Document Type:** Consultation Record
- **Clinical Specialty:** Pharmacology
- **Application Method:** Link of RuDReC Dataset

### 6.49.1 Task: RuDReC-NER

This task is to extract the following types of entities from the clinical text: "Adverse Drug Reaction", "Drugclass", "Finding", "Drugform", "Drugname", "Drug Interaction".

**Task type:** *Named Entity Recognition*
**Instruction:** *Given the clinical text of a patient in Russian, extract the following types of entities from the clinical text: "Adverse Drug Reaction", "Drugclass", "Finding", "Drugform", "Drugname", "Drug Interaction".*
*Return your answer in the following format. DO NOT GIVE ANY EXPLANATION:*
*entity: ..., type: ...;*
*...*
*entity: ..., type: ...;*
*The optional list for "type" is ["Adverse Drug Reaction", "Drugclass", "Finding", "Drugform", "Drugname", "Drug Interaction"].*
**Input:** *[Clinical text of a patient]*
**Output:** *entity: [clinical entity], type: [Adverse Drug Reaction / Drugclass / Finding / Drugform / Drugname / Drug Interaction];*

## 6.50 NorSynthClinical-PHI

The NorSynthClinical-PHI dataset[66] is a Norwegian dataset designed for Protected Health Information (PHI) de-identification. It includes 8,270 tokens and 409 annotated PHI instances, compiled by extending the NorSynthClinical synthetic clinical corpus. The dataset contains synthetic clinical text focusing on family history related to cardiac disease. Annotations were performed using Named Entity Tagging by two native Norwegian speakers, following guidelines adapted from previous de-identification research.

- **Language:** Norwegian
- **Clinical Stage:** Research
- **Sourced Clinical Document Type:** General EHR Note
- **Clinical Specialty:** Cardiology
- **Application Method:** Link of NorSynthClinical-PHI Dataset

#### 6.50.1 Task: NorSynthClinical-PHI

This task is to extract the following types of entities from the clinical records:"First_Name", "Last_Name", Age", "Health_Care_Unit", "Phone_Number", "Social_Security_Number", "Date_Full", "Date_Part", "Location".

> **Task type:** *Named Entity Recognition*
> **Instruction:** *Given the clinical text of a patient in Norwegian, extract the following types of entities from the clinical records:*
> *- "First_Name": An individual's first name.*
> *- "Last_Name": An individual's last name, including double last names and middle names that could be considered a last name. Any academic title (such as Dr.) should not be included.*
> *- "Age": Any age written with numbers, letters, or a mixture of both.*
> *- "Health_Care_Unit": All healthcare institutions where health care is provided (hospitals, healthcare centers, nursing homes, etc.) and their clinical units (emergency department, neuroclinic, department of medical genetics, etc.). Even abbreviations and other unofficial institution names should be included.*
> *- "Social_Security_Number ": A person's social security number.*
> *- "Phone_Number": A phone number. Any mentioned country codes should also be included.*
> *- "Date_Full": Including the combination of date, month, and year, written with numbers, letters, or a mixture of both.*
> *- "Date_Part": Including one or two of the instances date, month, and year, written with numbers, letters, or a mixture of both. Does not include weekdays, week numbers, or seasons.*
> *- "Location": Any geographical location, including countries, cities, towns, post codes, addresses, etc.*
> *Return your answer in the following format. DO NOT GIVE ANY EXPLANATION:*
> *entity: ..., type: ...;*
> *...*
> *entity: ..., type: ...;*
> *The optional list for "type" is ["First_Name", "Last_Name", "Age", "Health_Care_Unit", "Social_Security_Number", "Phone_Number", "Date_Full", "Date_Part", "Location"].*
> **Input:** *[Clinical text of a patient]*
> **Output:** *entity: [clinical entity], type: [First_Name / Last_Name / Age / Health_Care_Unit / Social_Security_Number / Phone_Number / Date_Full / Date_Part / Location];*

### 6.51 RuCCoN

RuCCoN[67] is a Russian-language dataset designed for clinical concept normalization. It comprises medical histories from more than 60 patients at the Scientific Center of Children's Health, focusing on allergic and pulmonary disorders. The dataset includes deidentified discharge summaries, radiology, echocardiography, and ultrasound diagnostic reports, as well as recommendations and other physician records. RuCCoN contains more than 16,028 entity mentions, manually mapped to 2,409 unique concepts from the Russian-language section of UMLS.

- **Language:** Russian
- **Clinical Stage:** Treatment and Intervention
- **Sourced Clinical Document Type:** General EHR Note

- **Clinical Specialty:** Pulmonology
- **Application Method:** Link of RuCCoN Dataset

#### 6.51.1 Task: RuCCoN

The objective of this task is to extract all clinical entities from the text and identify their corresponding type as one of the following categories: Disease, Symptom, Drug, Treatment, Body location, Severity, or Course.

> **Task type:** *Named Entity Recognition*
> **Instruction:** *Given the clinical text of a patient in Russian, extract the following types of entities from the clinical text:*
> *- "Disease": A definite pathologic process with a characteristic set of signs and symptoms.*
> *- "Symptom": Subjective evidence of disease perceived by the patient.*
> *- "Drug": A drug product that contains one or more active and/or inactive ingredients; it is intended to treat, prevent or alleviate the symptoms of disease.*
> *- "Treatment": Procedures concerned with the remedial treatment or prevention of diseases.*
> *- "Body location": Named locations of or within the body.*
> *- "Severity": The intensity of a specific adverse event evaluated based on the magnitude of clinical signs, symptoms and findings.*
> *- "Course": The course a disease typically takes from its onset, progression in time, and eventual resolution or death of the affected individual.*
> *Return your answer in the following format. DO NOT GIVE ANY EXPLANATION:*
> *entity: ..., type: ...;*
> *...*
> *entity: ..., type: ...;*
> *The optional list for "type" is ["Disease", "Symptom", "Drug", "Treatment", "Body location", "Severity", "Course"].*
> **Input:** *[Clinical text of a patient Russian]*
> **Output:** *entity: [clinical entity], type: [Disease / Symptom / Drug / Treatment / Body location / Severity / Course*

### 6.52 CLISTER

The CLISTER corpus[68] is a French Semantic Textual Similarity (STS) dataset comprising 1,000 sentence pairs from the clinical domain, each manually annotated with similarity scores. The clinical cases were sourced from CAS[69], a French corpus containing clinical case descriptions that encompass a variety of clinical information, including medical history, treatments, and follow-ups. Each pair of sentences was given a similarity score between 0 and 5, with 0 indicating no similarity and 5 indicating the highest level of similarity.

- **Language:** French
- **Clinical Stage:** Research
- **Sourced Clinical Document Type:** Case Report
- **Clinical Specialty:** General
- **Application Method:** Link of CLISTER Dataset

#### 6.52.1 Task: CLISTER

This task is to capture semantic textual similarity between French sentence pairs sourced from clinical cases.

**Task type:** *Semantic Similarity*
**Instruction:** *Given the following two clinical sentences that are labeled as "Sentence A" and "Sentence B" in French, decide the similarity of the two sentences. Specifically, analyze the potential similarity, including: Surface similarity: concerns the structural similarity. This similarity is based on grammatical words or words that are not related to the domain. Two sentences that have a surface similarity can be syntactically close but semantically distant. Semantic similarity: concerns medical concepts. The closer the concepts are to one another, the higher the similarity. These concepts can refer to medications, diseases, procedures, and others. Clinical compatibility: going further into the semantics, clinical compatibility is an assessment of whether sentences in a pair can refer to the same clinical case.*
*Then, assign a similarity score to the sentence pair based on the following scale:*
*- "0": For sentence pairs with only surface similarity, such as words non-specific to the medical domain or stop-words.*
*- "1": For sentence pairs with only surface similarity, concerning at most one medical entity.*
*- "2": For sentence pairs containing medical concepts with low semantic similarity, but no clinical compatibility. Typically, sentences in a pair can concern a disease, a procedure, or a drug.*
*- "3": For sentence pairs with semantic similarity on several medical concepts making them partially clinically compatible.*
*- "4": For sentence pairs with high semantic similarity and clinical compatibility. One sentence may contain more information than the other may, and vice-versa.*
*- "5": For sentence pairs with high semantic similarity and full clinical compatibility. The sentences have globally the same meaning, while one may be more specific than the other.*
*Return your answer in the following format. DO NOT GIVE ANY EXPLANATION:*
*similarity score: score*
*The optional list for "score" is ["0", "1", "2", "3", "4", "5"].*
**Input:** *[Clinical sentence pair in French]*
**Output:** *similarity score: [0 / 1 / 2 / 3 / 4 / 5]*

### 6.53 BRONCO150

BRONCO150[70] is a corpus containing 150 German discharge summaries from cancer patients diagnosed with hepatocellular carcinoma or melanoma, treated at Charite Universitaetsmedizin Berlin or Universitaetsklinikum Tuebingen. The corpus consists of 11,434 sentences and 89,942 tokens, with 11,124 annotations for medical entities and 3,118 annotations for related attributes. It is annotated with labels for diagnosis (ICD10), treatment (OPS), and medication (ATC), and the data is normalized to these terminologies. The corpus is provided in five splits (randomSentSet1-5) in both XML and CONLL formats. Annotation followed a structured, quality-controlled process involving two groups of medical experts to ensure consistency and high-quality annotations.

- **Language:** German
- **Clinical Stage:** Treatment and Intervention
- **Sourced Clinical Document Type:** Discharge Summary
- **Clinical Specialty:** Oncology
- **Application Method:** Link of BRONCO150 Dataset

#### *6.53.1 Task: BRONCO150-NER&Status*

The objective of this task is to extract all clinical entities from a cancer patient's discharge summary, classify their type, and identify their corresponding normalized terms.

**Task type:** *Event Extraction*
**Instruction:** *Given the following sentences from a discharge summary of a cancer patient (hepatocellular carcinoma or melanoma) in German, extract the medical entities with their types and identify their corresponding normalized terms.*
*- "diagnosis": A diagnosis is a disease, a symptom, or a medical observation that can be matched with the German Modification of the International Classification of Diseases-10 (ICD-10-GM: `https://www.bfarm.de/DE/Kodiersysteme/Klassifikationen/ICD/ICD-10-GM/_node.html`).*
*- "treatment": A treatment is a diagnostic procedure, an operation, or a systemic cancer treatment that can be found in the Operationen und Prozedurenschlu ̈ssel (OPS: `https://www.bfarm.de/DE/Kodiersysteme/Klassifikationen/OPS-ICHI/OPS/_node.html`).*
*- "medication": A medication names a pharmaceutical substance or a drug that can be related to the Anatomical Therapeutic Chemical Classification System (ATC: `https://www.bfarm.de/DE/Kodiersysteme/Klassifikationen/ATC/_node.html`).*
*Return your answer in the following format. DO NOT GIVE ANY EXPLANATION:*
*entity: ..., type: ..., normalized terms: ...;*
*...*
*entity: ..., type: ..., normalized terms: ...;*
*The optional list for "type" is ["diagnosis", "treatment", "medication"].*
*The normalized terms should be the corresponding text of normalized terms from the ICD-10-GM, OPS, or ATC classification systems.*
**Input:** *[Discharge summary of cancer patient in German]*
**Output:** *entity: [clinical entity], type: [diagnosis / treatment / medication], normalized terms: [ICD-10-GM / OPS / ATC normalized term]*

## 6.54 CARDIO:DE

The CARDIO:DE[71] dataset is a German clinical corpus containing 500 routine doctor's letters from Heidelberg University, representing a diverse range of cases from a tertiary care cardiovascular center during 2020 and 2021. The corpus includes 311 in-patient, 172 out-patient, and 17 cardiac emergency room letters, offering a comprehensive collection of clinical documentation. The dataset comprises a total of 993,143 tokens, with approximately 31,952 unique tokens. It is randomly divided into two subsets: CARDIO:DE400, which contains 400 documents with 805,617 tokens and 114,348 annotations, and CARDIO:DE100, which includes 100 documents, 187,526 tokens, and 26,784 annotations.

- **Language:** German
- **Clinical Stage:** Treatment and Intervention
- **Sourced Clinical Document Type:** General EHR Note
- **Clinical Specialty:** Cardiology
- **Application Method:** Link of CARDIO:DE Dataset

### *6.54.1 Task: CARDIO-DE*

The objective of this task is to extract the following types of entities from clinical documents related to the cardiovascular domain: "ACTIVEING", "DRUG", "DURATION", "FORM", "FREQUENCY", and "STRENGTH".

**Task type:** *Named Entity Recognition*
**Instruction:** *Given the following clinical document related to the cardiovascular domain in German, extract the following types of entities from the clinical text:*
*- "ACTIVEING": The primary ingredient in the medication responsible for its therapeutic effect.*

- *"DRUG": The name of the medication, including brand or generic name.*
- *"DURATION": The length of time the medication is to be taken.*
- *"FORM": The physical form of the medication, such as tablet, capsule, or liquid.*
- *"FREQUENCY": How often the medication should be taken within a specific time period.*
- *"STRENGTH": The concentration or dosage of the active ingredient in the medication.*

*Return your answer in the following format. DO NOT GIVE ANY EXPLANATION:*
*entity: ..., type: ...;*
*...*
*entity: ..., type: ...;*
*The optional list for "type" is ["ACTIVEING", "DRUG", "DURATION", "FORM", "FREQUENCY", "STRENGTH"].*
**Input:** *[Doctor's letter from the cardiology department in German]*
**Output:** *entity: [clinical entity], type: [ACTIVEING / DRUG / DURATION / FORM / FREQUENCY / STRENGTH]*

### 6.55 GraSSCo_PHI

GraSSCo_PHI[72] is a synthetic clinical dataset made up of carefully designed fictional discharge summaries. It comprises 63 clinical documents, containing approximately 5,000 sentences and over 43,000 tokens. The dataset features a nearly equal gender distribution, with about two-thirds of the cases involving hospitalized in-patients, and it covers a broad spectrum of medical specialties, including ophthalmology, oncology, and orthopedics. More than 1,400 instances of Protected Health Information (PHI) were automatically annotated using the Averbis Health Discovery (AHD) pipeline[73]. These annotations were then reviewed and refined by two trained annotators via the INCEpTION platform, who corrected any missed or misidentified PHI to improve annotation accuracy.

- **Language:** German
- **Clinical Stage:** Research
- **Sourced Clinical Document Type:** Discharge Summary, Case Report
- **Clinical Specialty:** General
- **Application Method:** Link of GraSSCo_PHI Dataset

#### 6.55.1 Task: GraSSCo PHI

The objective of this task is to extract all instances of Protected Health Information (PHI) from the synthetic patient discharge summaries and identify their corresponding PHI type.
[New task construction setting]: Based on the annotated PHI information, we formulated this task as an NER task, requiring the model to identify text spans corresponding to PHI entities and classify them into their respective types.

**Task type:** *Named Entity Recognition*
**Instruction:** *Given the clinical summaries of a patient in German, extract the following types of entities from the clinical text:*
- *"LOCATION_COUNTRY": Country name or reference*
- *"NAME_DOCTOR": Name of a medical professional*
- *"AGE": Numeric representation of a person's age*
- *"CONTACT_FAX": Fax number for communication*
- *"LOCATION_ZIP": Postal or ZIP code*
- *"LOCATION_ORGANIZATION": Name of an organization or institution*
- *"CONTACT_PHONE": Phone number for communication*
- *"DATE": Calendar date*

- *"LOCATION_CITY": Name of a city*
- *"CONTACT_EMAIL": Email address*
- *"NAME_PATIENT": Name of a patient*
- *"LOCATION_HOSPITAL": Name of a hospital or medical facility*
- *"PROFESSION": Job title or occupation*
- *"NAME_TITLE": Honorific or title before a name*
- *"NAME_USERNAME": Username for online identification*
- *"ID": Identification number or code*
- *"NAME_RELATIVE": Name of a family member*
- *"NAME_EXT": Name suffix, extension, or additional identifier*
- *"LOCATION_STREET": Street address or name*

*Return your answer in the following format. Do not output entities whose types do not exist in the clinical text. DO NOT GIVE ANY EXPLANATION:*
*entity: ..., type: ...;*
*...*
*entity: ..., type: ...;*
*The optional list for "type" is ["LOCATION_COUNTRY", "NAME_DOCTOR", "AGE", "CONTACT_FAX", "LOCATION_ZIP", "LOCATION_ORGANIZATION", "CONTACT_PHONE", "DATE", "LOCATION_CITY", "CONTACT_EMAIL", "NAME_PATIENT", "LOCATION_HOSPITAL", "PROFESSION", "NAME_TITLE", "NAME_USERNAME", "ID", "NAME_RELATIVE", "NAME_EXT", "LOCATION_STREET"]*
**Input:** *[Synthetic discharge summary of a patient in German]*
**Output:** *entity: [clinical entity], type: [LOCATION_COUNTRY / NAME_DOCTOR / AGE / CONTACT_FAX / LOCATION_ZIP / LOCATION_ORGANIZATION / CONTACT_PHONE / DATE / LOCATION_CITY / CONTACT_EMAIL / NAME_PATIENT / LOCATION_HOSPITAL / PROFESSION / NAME_TITLE / NAME_USERNAME / ID / NAME_RELATIVE / NAME_EXT / LOCATION_STREET]*

### 6.56 IFMIR

The IFMIR corpus[74] comprises 58,658 machine-annotated Japanese incident reports sourced from the Japan Council for Quality Health (JQ). These reports capture a range of medication-related incidents, such as incorrect administration, missed doses, and overdoses (https://www.med-safe.jp/index.html). The dataset includes 478,175 named entities and features three annotation types: medication entities, entity relations, and intention/factuality[75]. To validate the annotations, a random sample of 40 incident reports was manually reviewed.

- **Language:** Japanese
- **Clinical Stage:** Treatment and Intervention
- **Sourced Clinical Document Type:** Case Report
- **Clinical Specialty:** Pharmacology
- **Application Method:** Link of IFMIR Dataset

#### 6.56.1 Task: IFMIR-Incident type

The objective of this task is to identify and classify the types of incidents documented in the incident reports, recognizing that a single report may involve multiple incident types. The classification includes the following categories: "Wrong Drug", "Wrong Form", "Wrong Mode", "Wrong Strength amount", "Wrong Strength rate", "Wrong Strength concentration", "Wrong Timing", "Wrong Date", "Wrong Duration", "Wrong Frequency", "Wrong Dosage", "Wrong Route", "Others".

**Task type:** *Text Classification*
**Instruction:** *Given the medical incident report in Japanese, determine what type of incident occurred. One report might contain more than one incident type. The incident types and their definitions are as follows:*
*- "Wrong Drug": Wrong drug occurs when inappropriate medication or IV fluid is prescribed, dispensed, prepared or administered. Wrong drug applies when the intended drug and the actual drug are different. A generic substitution is not considered as a wrong drug.*
*- "Wrong Form": Wrong form occurs when the wrong form of drug is ordered, dispensed or administered.*
*- "Wrong Mode": Wrong mode occurs when the wrong mode of a medication is ordered, dispensed or administered.*
*- "Wrong Strength amount": Wrong amount is defined as a dose of medication or volume of IV fluid over or under the intended amount, taking into account the patient's age, weight, renal and liver function.*
*- "Wrong Strength rate": Wrong rate is defined as a rate, e.g., IV rate, being slower or faster than intended.*
*- "Wrong Strength concentration": Wrong concentration is defined as the concentration of a medication being higher or lower than intended. Concentration is also closely related to amount and rate; most cases of 'Wrong Strength concentration' co-occur with 'Wrong Strength rate' or 'Wrong Strength amount'. A wrong concentration might be reported as a wrong amount.*
*- "Wrong Timing": Timing-related errors are defined as administration too early or too late, relative to the time designated by the healthcare facility. There are three scenarios associated with wrong timing: No 'omission' or 'extra drug' results from wrong timing, 'omission' results from wrong timing, or 'extra drug' results from wrong timing.*
*- "Wrong Date": Wrong date refers to the medication being administered for a different date compared to the intended date.*
*- "Wrong Duration": Wrong duration refers to the medication being administered for a longer or shorter period than intended.*
*- "Wrong Frequency": A wrong frequency occurs when the prescribed or administered frequency of delivery for a drug or an IV rate falls outside of the recommended range or planned number. If the frequency is larger, it is often also labeled as an extra drug. If the frequency is smaller, then 'omission' is applicable. Wrong timing is also relevant is such cases.*
*- "Wrong Dosage": Patients may be subject to excessive or insufficient amounts of a drug.*
*- "Wrong Route": Wrong route occurs when a medication is prescribed or administered via an incorrect route of administration, e.g., a drug that creates strong vascular irritation and should be given via the central line is administered via the peripheral line.*
*- "Others": Other errors that are not covered by the current scope of the previous annotations, e.g., procedural errors such as forgetting to fill out a questionnaire before administrating a vaccine to a patient. For errors that are out of the scope of the above or the free text inputs does not present any error, the incident type is registered as 'Others'.*
*Assuming the number of incident types is N, return the N recognized incident types in the output.*
*Return your answer in the following format. DO NOT GIVE ANY EXPLANATION: incident type: type 1, type 2, ..., type N*
*The optional list for "type" is ["Wrong Drug", "Wrong Form", "Wrong Mode", "Wrong Strength amount", "Wrong Strength rate", "Wrong Strength concentration", "Wrong Timing", "Wrong Date", "Wrong Duration", "Wrong Frequency", "Wrong Dosage", "Wrong Route", "Others"].*
**Input:** *[Incident report in Japanese]*
**Output:** *incident type: [Wrong Drug / Wrong Form / Wrong Mode / Wrong Strength amount / Wrong Strength rate / Wrong Strength concentration / Wrong Timing / Wrong Date / Wrong Duration / Wrong Frequency / Wrong Dosage / Wrong Route / Others]*

#### 6.56.2 Task: IFMIR-NER

The objective of this task is to extract the following entity types from the Japanese incident reports: "Strength concentration", "Frequency", "Date", "Drug", "Dosage", "Strength rate", "Drug form", "Duration", "Strength amount", "Drug mode", "Route", "Timing".

> **Task type:** *Named Entity Recognition*
> **Instruction:** *Given the following medical incident report in Japanese, extract the following types of entities from the medical text.*
> *- "Strength concentration": Concentration is defined as diluted medication concentration with nominator and denominator or presented as percentage or IV fluid concentration.*
> *- "Frequency": Frequency is defined as how many times a drug is given per unit of time.*
> *- "Date": Date is defined as a time unit including a date and time unit longer than one day.*
> *- "Drug": The intended to deliver or actual delivered drug name, or entities described as drugs.*
> *- "Dosage": Dosage is defined as the number of units (e.g., tables, bottles, ampules) given to the patient as a single dose.*
> *- "Strength rate": Rate typically represents one measure against another quantity or measure.*
> *- "Drug form": The form of a drug (e.g., tablet, subcutaneous injection).*
> *- "Duration": Duration is defined as the period during which a drug is administered to the patient.*
> *- "Strength amount": The amount is defined as medication dose or IV fluid volume.*
> *- "Drug mode": The mode is a drug mode of action that is associated with pharmacologic action.*
> *- "Route": Route is defined as the route of drug administration to the patient, which may include the infusion sites, routes and pumps.*
> *- "Timing": Timing is defined as a scheduled administration time that is predefined as time interval.*
> *Return your answer in the following format. DO NOT GIVE ANY EXPLANATION:*
> *entity: ..., type: ...;*
> *...*
> *entity: ..., type: ...;*
> *The optional list for "type" is ["Strength concentration", "Frequency", "Date", "Drug", "Dosage", "Strength rate", "Drug form", "Duration", "Strength amount", "Drug mode", "Route", "Timing"].*
> **Input:** *[Incident report in Japanese]*
> **Output:** *entity: [clinical entity], type: [Strength concentration / Frequency / Date / Drug / Dosage / Strength rate / Drug form / Duration / Strength amount / Drug mode / Route / Timing]*

#### 6.56.3 Task: IFMIR - NER&factuality

The objective of this task is to extract the following entity types from Japanese incident reports: "Strength Concentration," "Frequency," "Date," "Drug," "Dosage," "Strength Rate," "Drug Form," "Duration," "Strength Amount," "Drug Mode," "Route," and "Timing." Additionally, this task requires determining the intention and factuality of the incident. Intention refers to whether the medication was intended to be administered, while factuality indicates whether the medication was actually administered.

> **Task type:** *Event Extraction*
> **Instruction:** *Given the following medical incident report in Japanese, extract the medical entities with their corresponding types and intention and factuality information. Specifically, need to extract all the following information for each entity:*
> *1. Entity types:*
> *- "Strength concentration": Concentration is defined as diluted medication concentration with nominator and denominator or presented as percentage or IV fluid concentration.*

*- "Frequency": Frequency is defined as how many times a drug is given per unit of time.*
*- "Date": Date is defined as a time unit including a date and time unit longer than one day.*
*- "Drug": The intended to deliver or actual delivered drug name, or entities described as drugs.*
*- "Dosage": Dosage is defined as the number of units (e.g., tables, bottles, ampules) given to the patient as a single dose.*
*- "Strength rate": Rate typically represents one measure against another quantity or measure.*
*- "Drug form": The form of a drug (e.g., tablet, subcutaneous injection). - "Duration": Duration is defined as the period during which a drug is administered to the patient.*
*- "Strength amount": The amount is defined as medication dose or IV fluid volume.*
*- "Drug mode": The mode is a drug mode of action that is associated with pharmacodynamic action.*
*- "Route": Route is defined as the route of drug administration to the patient, which may include the infusion sites, routes and pumps.*
*- "Timing": Timing is defined as a scheduled administration time that is predefined as time interval.*
*2. Intention / factuality information:*
*- "IA": Intended & Actual. The entity was intended to be given and was actually given. This indicates no error has occurred as to this entity.*
*- "IN": Intended & Not-actual. The entity was intended to be given but actually was not given. This indicates the intended medication was not delivered.*
*- "NA": Not-intended & Actual. The entity was not intended to be given but actually was. This indicates the not intended medication was mistakenly delivered.*
*Return your answer in the following format. DO NOT GIVE ANY EXPLANATION:*
*entity: ..., type: ..., intention: ...;*
*...*
*entity: ..., type: ..., intention: ...;*
*The optional list for "type" is ["Strength concentration", "Frequency", "Date", "Drug", "Dosage", "Strength rate", "Drug form", "Duration", "Strength amount", "Drug mode", "Route", "Timing"].*
*The optional list for "intention" is ["IA", "IN", "NA"].*
**Input:** *[Incident report in Japanese]*
**Output:** *entity: [clinical entity], type: [Strength concentration / Frequency / Date / Drug / Dosage / Strength rate / Drug form / Duration / Strength amount / Drug mode / Route / Timing], intention: [IA / IN / NA]*

### 6.57 iCorpus

iCorpus[76] is a Japanese-language dataset created from publicly accessible case reports sourced via J-STAGE, a Japanese platform for academic publications. The dataset contains text extracted from the case sections of reports that mention intractable diseases recognized by the Japanese Ministry of Health, Labor, and Welfare. iCorpus includes 179 case reports (across 183 files), representing 102 out of the 333 officially designated intractable diseases. Annotations were carried out using Brat (`https://github.com/aih-uth/brat_entity_linking`), an open-source annotation tool[77]. The annotation guidelines were developed, refined, and validated by experts in medical informatics.

- **Language:** Japanese
- **Clinical Stage:** Treatment and Intervention
- **Sourced Clinical Document Type:** Case Report
- **Clinical Specialty:** General
- **Application Method:** Link of iCorpus Dataset

#### *6.57.1 Task: iCorpus*

The objective of this task is to extract the following types of entities from the case reports: "age", "sex", "smoking", "drinking", "state", "body", "tissue", "item", "clinical test", "PN", "judge", "quantity evaluation", "quantity progress", "quality evaluation", "quality progress", "value", "unit", "time", or "time span".

> **Task type:** *Named Entity Recognition*
> **Instruction:** *Given the clinical report of a patient in Japanese, extract the following types of entities from the clinical text:*
> - *"age": 年齢を示す表現, 例: 63歳, 56歳, 1歳6か月, 高校生.*
> - *"sex": 性別を示す表現, 例: 男性, 女性, 男児, 女児.*
> - *"smoking": 喫煙に関する表現, 例: 喫煙, タバコ, 喫煙歴, 禁煙.*
> - *"drinking": 飲酒に関する表現, 例: 飲酒, アルコール, 飲酒歴, 禁酒.*
> - *"state": 患者の状態全般を示す表現. いわゆる, 病名, 症状 (患者の訴え), 所見(観察結果) などを含む, 例: 吐き気, 萎縮症, 糖尿病, 口渇.*
> - *"body": 人体部位. 特定の部位を示す表現, 例: 頭,胃,肝,手足, 眼瞼結膜.*
> - *"tissue": 人体組織.人体各所で繰り返し出現するもの, 例: 筋,筋肉,粘膜, 細胞, 繊維.*
> - *"item": 患者の状態を表すために参照される項目, 例: 血糖,血糖値, HbA1c, 食欲.*
> - *"clinical test": 臨床検査に関する表現. item との違いは計測法を含むか否か, 例: 神経学的検査, 徒手筋力検査.*
> - *"PN": 患者の状態が, ある (Positive), ない (Negative), わからない (None) ことを示す表現, 例: で、認め, 示す, 認めるなし, 認めず, ではなく, なく不明であった、詳細不明.*
> - *"judge": 医療者により, 患者の状態がある (Positive), あることが疑われ る (Suspicious), 将来あるもしれない (Future), ない (Negative), 不明 (None)であることが判断されたことを示す表現, 例: 診断された, 考えられた, 疑われた, 可能性も考え否定的, 明らかではなかった確定診断に至らなかった.*
> - *"quantity evaluation": 数値への評価。 高い (High), 正常 (Normal) 低い (Low), 例: 上昇, 異常高値, 増加, 正常, 基準値, 保たれて低下, 減少, 減弱.*
> - *"quantity progress": 数値の変化。 上昇 (Increase) 変化なし (NoChange)低下 (Decrease).*
> - *"quality evaluation": 数値以外の状態の程度を質的に示す表現. 主に疾患の重症度を示 す 軽度 (Mild) 中等度 (Moderate) 重度・高度 (Severe), 例: 軽度, 軽い, わずか, やや中等度, 中度, 中等症強い, 著名, 著しい, 重度.*
> - *"quality progress": 数値以外の状態の時間的な変化を質的に示す表現. 出現 (Start) 悪化 (Worsen), 持続 (NoChange), 改善 (Improve), 軽快 (Recover), 例: 出現した, なった, きたした悪化, 増悪, 進行, 顕在化持続, 保たれて, 変わらず改善, 軽快, 回復落ち着き, 復帰, 軽快, 回復.*
> - *"value": 検査値など、 身体や検体を測定し得られる数値, 例: 7.5, 20, 1, 5, 165.0.*
> - *"unit": 数値との組で表される単位, 例: mg/日, 行, cm, kg/m2.*
> - *"time": 時間軸上における特定位置の時点や区間を示す表現, 例: 約10年前, その後直後.*
> - *"time span": 時間軸上の位置を問わず時間幅を示す表現, 例: 1日, 長時間, 2カ月間.*
>
> *Return your answer in the following format. DO NOT GIVE ANY EXPLANATION:*
> *entity: ..., type: ...;*
> *...*
> *entity: ..., type: ...;*
> *The optional list for "type" is ["age", "sex", "smoking", "drinking", "state", "body", "tissue", "item", "clinical test", "PN", "judge", "quantity evaluation", "quantity progress", "quality evaluation", "quality progress", "value", "unit", "time", "time span"].*
> **Input:** *[Clinical report of patient in Japanese]*
> **Output:** *entity: [clinical entity], type: [age / sex / smoking / drinking / state / body / tissue / item / clinical test / PN / judge / quantity evaluation / quantity progress / quality evaluation / quality progress / value / unit / time / time span]*

### 6.58 icliniq-10k

The icliniq-10k dataset[78] contains real English conversations between patients and doctors collected from online medical consultation platforms. It includes 7,231 records, each consisting of a patient's healthcare inquiry and the corresponding doctor's response.

- **Language:** English
- **Clinical Stage:** Triage and Referral
- **Sourced Clinical Document Type:** Consultation Record
- **Clinical Specialty:** General
- **Application Method:** Link of icliniq-10k Dataset

#### 6.58.1 Task: icliniq-10k

This task is to generate the doctor's response based on the provided dialogue history of a medical consultation.

> **Task type:** *Question Answering*
> **Instruction:** *Given the following question from a patient, generate the doctor's response based on the dialogue context. Return your answer in the following format. DO NOT GIVE ANY EXPLANATION:*
> *doctor: ...*
> **Input:** *[Clinical query from patient ]*
> **Output:** *doctor: [generated response from doctor's perspective]*

### 6.59 HealthCareMagic-100k

The HealthCareMagic-100k dataset[78] consists of real English conversations between patients and doctors collected from an online medical consultation platform. It contains 111,996 records, each comprising a healthcare inquiry from the patient and the doctor's corresponding reply.

- **Language:** English
- **Clinical Stage:** Triage and Referral
- **Sourced Clinical Document Type:** Consultation Record
- **Clinical Specialty:** General
- **Application Method:** Link of HealthCareMagic-100k Dataset

#### 6.59.1 Task: HealthCareMagic-100k

This task is to generate the doctor's response based on the provided dialogue history of a medical consultation.

> **Task type:** *Question Answering*
> **Instruction:** *Given the following question from a patient, generate the doctor's response based on the dialogue context. Return your answer in the following format. DO NOT GIVE ANY EXPLANATION:*
> *doctor: ...*
> **Input:** *[Clinical query from patient ]*
> **Output:** *doctor: [generated response from doctor's perspective]*